%% file: main.tex
\documentclass[11pt,iclr]{modernstyle}

\definecolor{maincolor}{HTML}{e56b6f}  
\definecolor{bgcolor}{HTML}{D3EEFF}    
\definecolor{abscolor}{HTML}{22223b}   

\hypersetup{pdftitle={DyadMem: A Long-Term Memory Benchmark of How Agents Work with Users}}

\usepackage[square,numbers]{natbib}
\input{math_commands.tex}

\usepackage{textcomp}
\usepackage{url}
\usepackage{amssymb}
\usepackage{makecell}
\usepackage{algorithm}
\usepackage{algpseudocode}
\usepackage{wrapfig}
\usepackage{colortbl}
\usepackage{upquote} 

\datasetlink{https://huggingface.co/datasets/RenaissanceT/DyadMem}
\githublink{https://github.com/RenaissanceT/DyadMem}
\projectpagelink{https://renaissancet.github.io/DyadMem/}

\usepackage{titletoc}
\definecolor{appred}{RGB}{176,32,32}
\titlecontents{part}[0pt]
  {\addvspace{8pt}\bfseries\color{appred}}
  {}
  {}
  {{\color{appred}\titlerule*[0.55em]{.}}\color{black}\contentspage}
\titlecontents{section}[1.6em]
  {\addvspace{2pt}\color{appred}}
  {\thecontentslabel\hspace{0.6em}}
  {}
  {{\color{appred}\titlerule*[0.55em]{.}}\color{black}\contentspage}
\titlecontents{subsection}[3.2em]
  {\addvspace{1pt}\color{appred}}
  {\thecontentslabel\hspace{0.55em}}
  {}
  {{\color{appred}\titlerule*[0.55em]{.}}\color{black}\contentspage}

\definecolor{modgray}{RGB}{90,96,106}

\newcounter{appmodule}
\newcommand{\appmodule}[2][]{%
  \clearpage
  \refstepcounter{appmodule}%
  \phantomsection
  \addcontentsline{toc}{part}{#2}%
  \pdfbookmark[1]{#2}{appmodule.\theappmodule}%
  \noindent
  \begin{minipage}{\linewidth}%
    {\color{appred}\rule{\linewidth}{1.1pt}}\par\vspace{5pt}%
    {\normalfont\large\bfseries\color{appred}%
     Part~\Roman{appmodule}\quad\textbar\quad #2}\par
    \if\relax\detokenize{#1}\relax\else
      \vspace{3pt}%
      {\normalfont\small\itshape\color{modgray}#1}\par
    \fi
    \vspace{5pt}{\color{appred}\rule{\linewidth}{0.4pt}}%
  \end{minipage}%
  \par\addvspace{10pt}%
}

\definecolor{AcademicBlue}{HTML}{3970AD}
\definecolor{PromptBG}{HTML}{F6F8FA}
\definecolor{PromptText}{HTML}{252A30}

\lstdefinestyle{promptstyle}{
    basicstyle=\ttfamily\footnotesize\color{PromptText},
    breaklines=true,
    breakatwhitespace=false,
    columns=fullflexible,
    keepspaces=true,
    showstringspaces=false,
    upquote=true
  }

\newtcblisting{promptbox}[1]{%
    enhanced,
    breakable,
    listing only,
    colback=PromptBG,
    colframe=AcademicBlue,
    coltitle=white,
    colbacktitle=AcademicBlue,
    title={#1},
    fonttitle=\bfseries\small,
    center title,
    boxrule=0.8pt,
    arc=2pt,
    left=8pt,
    right=8pt,
    top=7pt,
    bottom=7pt,
    listing options={style=promptstyle}
  }

\definecolor{peach}{HTML}{FAD7B0}
\definecolor{sky}{HTML}{D6E6F5}
\definecolor{sage}{HTML}{DCE8D0}
\definecolor{rose}{HTML}{F3D0D0}
\definecolor{capH}{HTML}{F6E6D4}
\definecolor{updH}{HTML}{E5EED8}
\definecolor{recH}{HTML}{D9E6F2}
\definecolor{qaH}{HTML}{EDE4F2}
\definecolor{secH}{HTML}{F3F3F3}

\title{DyadMem: A Long-Term Memory Benchmark\\
of How Agents Work with Users}

\author[1,4,*]{Yifei Tao}
\author[4,*]{Xinyu Zhong}
\author[2,*]{Henry Hengyuan Zhao}
\author[4,*,\ddagger]{Fanyi Wang}
\author[4]{Tengda Guo}
\author[4]{Wentao Qiu}
\author[3]{Ying Wang}
\author[4,\dagger]{Liujian Tang}

\affiliation[1]{Nanyang Technological University}
\affiliation[2]{National University of Singapore}
\affiliation[3]{University of Hong Kong}
\affiliation[4]{Stepfun}

\contribution[*]{Equal contribution}
\contribution[\ddag]{Team lead}
\contribution[\dag]{Corresponding author}

\date{\today}
\correspondence{%
  \email{alex@stepfun.com}%
}

\begin{document}

\abstract{%
Long-term agents must remember not only what is true about a user, but also how a particular agent should work with that user as their shared history evolves. Existing benchmarks primarily supervise user facts and preferences or experience reusable across users, leaving this relationship-specific agent memory implicit. Additionally, most prior works measure the model solely with final-answer QA over long interaction histories, making the assessment still incomplete and unreliable. To this end, we introduce \textbf{DyadMem} with the proposed new definition \emph{User-conditioned Relational Agent Memory} (URAM). DyadMem jointly annotates user-side memory and URAM along the same multi-session trajectories, resulting in 6 memory categories. To summarize, it includes \textbf{3,065} episodes, \textbf{50,961} sessions, and \textbf{61,210} QA instances, with extensive session-level Capture and Update gold annotations, query-level Recall support, and two QA settings: Gold-Memory and Full-Pipeline. Across 16 open-weight and 4 proprietary models, Gold-Memory QA is consistently strong, yet Full-Pipeline QA drops sharply. Such a gap explicitly supports our fine-grained evaluation design. Additionally, several quantitative results further reveal low Capture recall, incomplete Recall, and unsafe-deletion issues arising from even the frontier LLMs. We further conduct a rigorous experiment to validate the effectiveness of our URAM and observe the positive effects for all 20 models. In summary, DyadMem is a dual-domain, full-pipeline memory benchmark with extensive annotation efforts for advancing the domain's development.%
}

\maketitle


\begin{figure}[H]
    \centering
    \includegraphics[width=\linewidth]{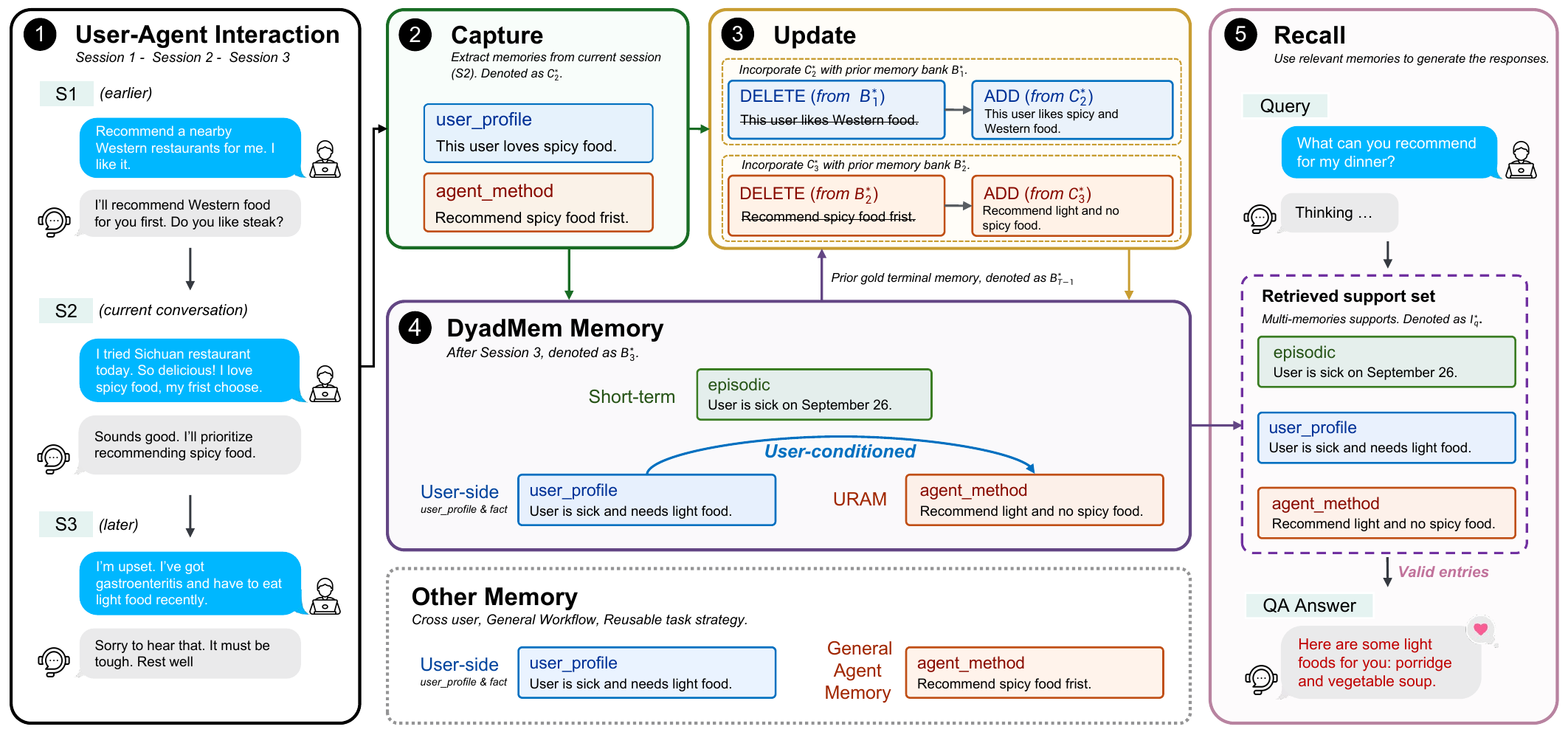}
    \caption{DyadMem is a dual-domain, full-pipeline memory benchmark. Across multi-session user--agent interactions, Capture extracts current-session memories, Update reconciles them with the prior bank, Recall selects valid terminal-bank entries, and QA uses the selected evidence. The evolving bank contains shared episodic memory, user-side memory, and User-conditioned Relational Agent Memory (URAM); unlike general agent memory that transfers across users, URAM records how this particular agent should work with this particular user.}
    \label{fig:dyadmem-overview}
\end{figure}

\clearpage

\input{Chapters/Chapter_1_Intro}
\input{Chapters/Chapter_2_Related_Work}

\input{Chapters/Chapter_3_Method}
\input{Chapters/Chapter_4_Experiments}
\input{Chapters/Chapter_5_Conclusion}

{
    \small
    \bibliography{main}
}

\clearpage
\appendix
\input{Appendix/0_Appendix}

\end{document}

%% file: math_commands.tex
\usepackage{amsmath,amsfonts,bm}

\def\eqref#1{equation~\ref{#1}}

\def\1{\bm{1}}

\DeclareMathAlphabet{\mathsfit}{\encodingdefault}{\sfdefault}{m}{sl}
\SetMathAlphabet{\mathsfit}{bold}{\encodingdefault}{\sfdefault}{bx}{n}



%% file: Chapters/Chapter_1_Intro.tex
\section{Introduction}

Long-term interaction \citep{maharana2024evaluating,wu2025longmemeval,yao2024taubench} requires more than remembering the past. An agent must retain important facts about the user \citep{du2024perltqa,xu2022dulemon,jiang2025personamem}, but it must also learn how its own working relationship with that user has changed \citep{chao2026stale,uddin2026memora,wu2026longmemevalv2,mehri2026multisessioncollab}. 

Prior work has studied user memory and agent experience largely along separate lines. User-memory benchmarks model facts, profiles, preferences, and their evolution \citep{jiang2025know,pakhomov2025convomem,uddin2026memora,chao2026stale}, while experience-based agent memory extracts workflows, environment knowledge, and failure patterns that can transfer across tasks \citep{wang2025awm,fang-etal-2026-memp,wu2026longmemevalv2}. Recent systems further adapt behavior from user corrections and long-term feedback \citep{mehri2026multisessioncollab}. Across these lines, a complementary memory object remains largely implicit: \emph{what should this agent do differently for this particular user?} We call this missing form of memory \emph{User-conditioned Relational Agent Memory} (URAM). URAM is not another description of the user: two agents serving the same user may require different future behavior because their interaction histories differ. This motivates a \textbf{dual-domain view of long-term memory} that represents both user-side memory and user-conditioned agent-side memory.

How memory is evaluated creates a second gap. Widely used long-term-memory benchmarks commonly measure success through downstream QA over long interaction histories \citep{maharana2024evaluating,wu2025longmemeval,du2024perltqa}. This endpoint is important, but it collapses distinct capabilities: constructing the right memories, maintaining them as information changes, retrieving complete support, and using that support in an answer. Recent benchmarks begin to expose selected operations or failure stages \citep{cheng2026amemgym,chen2025halumem,hao2026memops}, but do not jointly supervise this full lifecycle for both user-side and relationship-specific agent memory. Consequently, a correct answer can mask an incomplete or unsafe memory bank, while an incorrect answer cannot identify where the pipeline failed. Evaluating relational memory therefore requires both memory domains to be tracked on the same trajectory and each stage of their lifecycle to be measured directly.



We therefore introduce \textbf{DyadMem}, a dual-domain, full-pipeline benchmark for long-term memory. DyadMem follows a \emph{dialogue-first} construction process: complete conversations are generated and verified before user-side memory, URAM, and their changes are annotated from the realized interaction. It contains \textbf{3,065 episodes}, \textbf{50,961 sessions}, and \textbf{61,210 QA instances} across \textbf{5 major domains} and \textbf{25 subdomains}. As summarized in Figure~\ref{fig:dyadmem-overview}, session-level Capture and Update annotations and query-level Recall supports separately evaluate memory construction, updating, and retrieval. \textbf{Gold-Memory QA} measures answer generation when the annotated supporting memories are supplied directly, whereas \textbf{Full-Pipeline QA} answers from a memory bank constructed and retrieved by the model itself. Together, the stage-level tasks localize failures, and the two QA settings quantify their end-to-end consequence.

Across 16 open-weight models and 4 proprietary models, DyadMem exposes failures hidden by final-answer evaluation. Gold-Memory QA ranges from 78.7\% to 96.3\%, yet Full-Pipeline QA averages only 47.7\% for open-weight models and 57.4\% for proprietary models. Alongside this gap, stage-level results reveal low Capture recall, incomplete Recall, and unsafe-deletion issues. These observations indicate our benchmark assessment is comprehensive and holistic. We further conduct experiment to validate the effects of involving user conditions in agent memory construction. The results show that jointly incorporating user-side memory and URAM outperforms general experience memory across all 20 LLMs, demonstrating the essential necessity of our design, which is hardly considered in previous benchmarks.

Our contributions are as follows:

\begin{itemize}

\item \textbf{We formalize relationship-specific agent memory.}
We introduce URAM to represent how a particular agent should work with a particular user as their shared history evolves.

\item \textbf{We introduce a large-scale dual-domain memory benchmark.}
DyadMem jointly annotates user-side memory, URAM, and shared episodic memory on the same 3,065 multi-session trajectories, with dialogue-grounded lifecycle supervision.

\item \textbf{We evaluate the complete memory pipeline.}
DyadMem separately evaluates Capture, gold-conditioned standalone Update, Recall, Gold-Memory QA, and Full-Pipeline QA, making failures localizable rather than visible only through final-answer accuracy.

\end{itemize}

%% file: Chapters/Chapter_2_Related_Work.tex
\section{Related Work}


\noindent\textbf{What long-term memory represents.}
LoCoMo \citep{maharana2024evaluating}, LongMemEval \citep{wu2025longmemeval}, and PerLTQA \citep{du2024perltqa} evaluate long-horizon conversational or personal memory. ConvoMem \citep{pakhomov2025convomem}, PersonaMem \citep{jiang2025personamem}, Memora \citep{uddin2026memora}, and STALE \citep{chao2026stale} further model changing profiles, preferences, and obsolete information. Their supervised memory contents nevertheless primarily describe the user. A parallel line extracts reusable task experience: Agent Workflow Memory \citep{wang2025awm}, Memp \citep{fang-etal-2026-memp}, and LongMemEval-V2 \citep{wu2026longmemevalv2} capture workflows or dynamic task knowledge, while MemoryAgentBench \citep{hu2026memoryagentbench} and Mem2ActBench \citep{shen2026mem2act} study incremental memory for action. DyadMem differs by representing both the user and how a particular agent should work with that user on the same interaction trajectory. URAM is relationship-specific rather than a generic user profile or task-transferable experience.

\noindent\textbf{User-conditioned agent behavior.}
PrefEval \citep{zhao2025llms} evaluates preference following, MultiSessionCollab \citep{mehri2026multisessioncollab} derives actionable reflections from cross-session corrections, and PAHF \citep{liang2026learning}, SAGER \citep{tao2026sager}, PersonaAgent \citep{zhang2026personaagent}, and MemToolAgent \citep{er2026memtoolagent} adapt behavior through user-specific memory, policies, prompts, or feedback. These methods establish the value of adaptation but generally assess the resulting behavior rather than directly supervise evolving agent-side memory. DyadMem makes this memory explicit: URAM entries are typed, grounded in dialogue provenance, co-annotated with user-side memory on the same interaction trajectory, and labeled when they are added, revised, or invalidated. This makes relational adaptation directly inspectable.

\noindent\textbf{Lifecycle-level evaluation of long-term memory.}
Prior benchmarks expose different parts of the memory process. LongMemEval \citep{wu2025longmemeval} and PerLTQA \citep{du2024perltqa} analyze retrieval and answer construction, AMemGym \citep{cheng2026amemgym} probes write, read, and utilization failures; HaluMem \citep{chen2025halumem}, MemOps \citep{hao2026memops}, and Memora \citep{uddin2026memora} provide extraction, lifecycle, or state-transition supervision; and MemProbe \citep{ma2026memprobe} audits stored artifacts through user-state recovery. As summarized in Table~\ref{tab:benchmark_comparison}, DyadMem brings these capabilities together for two coupled memory domains: per-session gold memory for construction, gold update operations for maintenance, and query-to-bank supports for retrieval. Gold-Memory QA reads those recall supports as given evidence, and Full-Pipeline QA composes the three stages, so the two answer settings measure the resulting loss rather than a further annotation.


\DeclareRobustCommand{\supportmark}{\textcolor{green!55!black}{\scalebox{1.25}{\(\checkmark\)}}}
\DeclareRobustCommand{\partialmark}{\textcolor{red!25!yellow}{\(\triangle\)}}
\DeclareRobustCommand{\nomark}{\textcolor{red}{\(\times\)}}

\begin{table*}[h]

\centering
\small
\setlength{\tabcolsep}{4.2pt}
\caption{Comparison with related works. \supportmark\ denotes direct support, \partialmark\ related but non-equivalent supervision, and \nomark\ no corresponding supervision. Per-session gold memory denotes the entries to extract from that session. Gold lifecycle annotations denote add, revise, and invalidate operations on the prior bank and do not repeat the per-session extraction set. Query-to-bank supports are gold links from each QA query to entries that remain valid in the terminal memory bank and are used as the recall target.}
\label{tab:benchmark_comparison}
\resizebox{0.94\textwidth}{!}{
\begin{tabular}{lccccc}
\toprule
\textbf{Benchmark} &
\makecell{\textbf{User-side}\\\textbf{memory}} &
\makecell{\textbf{User-conditioned}\\\textbf{agent memory}} &
\makecell{\textbf{Per-session}\\\textbf{gold memory}} &
\makecell{\textbf{Gold lifecycle}\\\textbf{annotations}} &
\makecell{\textbf{Query-to-bank}\\\textbf{supports}} \\
\midrule
LoCoMo \citep{maharana2024evaluating}                  & \supportmark & \nomark      & \nomark      & \nomark      & \partialmark \\
LongMemEval \citep{wu2025longmemeval}                  & \supportmark & \nomark      & \nomark      & \nomark      & \partialmark \\
MultiSessionCollab \citep{mehri2026multisessioncollab} & \partialmark & \partialmark & \nomark      & \nomark      & \nomark \\
AMemGym \citep{cheng2026amemgym}                       & \supportmark & \nomark      & \partialmark & \nomark      & \nomark \\
HaluMem \citep{chen2025halumem}                        & \supportmark & \nomark      & \supportmark & \supportmark & \partialmark \\
MemOps \citep{hao2026memops}                           & \supportmark & \nomark      & \partialmark & \supportmark & \partialmark \\
Memora \citep{uddin2026memora}                         & \supportmark & \nomark      & \partialmark & \supportmark & \partialmark \\
MemProbe \citep{ma2026memprobe}                        & \supportmark & \nomark      & \nomark      & \nomark      & \nomark \\
\midrule
\textbf{DyadMem}                                       & \supportmark & \supportmark & \supportmark & \supportmark & \supportmark \\
\bottomrule
\end{tabular}}
\vspace{-0.1in}
\end{table*}

%% file: Chapters/Chapter_3_Method.tex
\section{Method}

\subsection{Task Formulation}

DyadMem represents an episode as a chronologically ordered sequence of sessions together with a QA set. Session \(t\) contains a temporal anchor \(\tau_t\), the complete user--assistant dialogue \(D_t\), gold Capture entries \(C_t^\star\), and gold Update operations \((L_t^\star,A_t^\star)\) that delete and add memory entries. The temporal anchor is the session date and time used to resolve relative expressions such as ``tomorrow.'' Each QA instance contains a question \(q_j\), a human-verified reference answer \(a_j^\star\), and a gold Recall support set \(I_j^\star\) over the valid entries in the terminal memory bank, i.e., the bank after the final session. The superscript \(\star\) denotes a human-verified gold annotation.

Each memory entry has an episode-unique index, a type, and self-contained content. DyadMem defines 6 types. \texttt{user\_profile} stores persistent user attributes, preferences, and constraints, while \texttt{fact} stores durable objective information provided by the user. The three URAM types describe the agent in relation to that user: \texttt{agent\_identity} captures role and behavioral boundaries, \texttt{agent\_method} captures reusable collaboration procedures, and \texttt{agent\_knowledge} captures user-conditioned decision rules learned through interaction and feedback. The remaining \texttt{episodic} type records concrete interaction events and serves as shared interaction memory for both domains. Thus, user-side memory consists of \texttt{user\_profile} and \texttt{fact}, URAM consists of the three agent-side types, and shared \texttt{episodic} entries may support either side. Throughout, \emph{memory} denotes retained long-term information, a \emph{memory entry} is one atomic record, a \emph{memory bank} is a collection of entries, and \emph{memory state} refers only to the contents of that bank at a particular session.

Let \(B_t^\star\) be the gold memory state after session \(t\), i.e., the valid contents of the gold bank, with \(B_0^\star=\varnothing\). Its chronological transition is
\begin{equation}
B_t^\star
=
\operatorname{Dedup}
\left(
\left[
\left(
B_{t-1}^\star \cup C_t^\star
\right)
\setminus L_t^\star
\right]
\cup A_t^\star
\right).
\end{equation}
Here, \(\operatorname{Dedup}\) collapses entries with the same memory key while retaining the last occurrence; the key is the entry index when present and otherwise its content. Capture receives only the complete current session \(D_t\) and its temporal anchor; no earlier sessions are supplied. In the main evaluation, \emph{gold-conditioned standalone Update} receives the current dialogue, the gold prior bank \(B_{t-1}^\star\), and gold current-session Capture entries \(C_t^\star\), isolating it from upstream errors. \emph{Full-Pipeline Update} instead uses the model's own prior bank and Capture output. Recall selects an ordered evidence set from the valid entries in a terminal bank. Full notation and task equations are in Appendix~\ref{app:method-details}.

\subsection{Dataset Construction}

DyadMem follows a \emph{plan-guided, dialogue-first} construction paradigm. Generation plans specify which kinds of memory change should occur and when, but do not prescribe the final memory entries. Complete multi-session dialogues are generated and reviewed first. Capture, Update, and Recall annotations are then reconstructed from the realized interaction, keeping every annotation traceable to dialogue evidence. This follows the general dialogue-first construction strategy of LoCoMo \citep{maharana2024evaluating}.

For each finalized dialogue, we annotate the 6 types of Capture entry and replay the episode chronologically to maintain the gold bank. A superseded memory entry produces a deletion and a replacement addition, an explicitly terminated entry produces only a deletion, and sessions without a substantive change leave the bank unchanged. Gold Recall supports may reference only entries that remain valid in the terminal bank. \texttt{user\_centric} QA primarily targets \texttt{user\_profile} and \texttt{fact}, whereas \texttt{agent\_centric} QA primarily targets URAM, either may additionally require shared \texttt{episodic} evidence.

All episodes are reviewed by 9 quality-control annotators against the original dialogue and the chronologically reconstructed bank. Corrections are propagated through memory indices and affected QA annotations. In summary, we present the resulting dataset composition and scale in Figure~\ref{fig:dataset-statistics-main}. Additional annotation and quality-control details in Appendix~\ref{app:method-details}.

\begin{figure*}[t]
    \centering
    \includegraphics[width=\linewidth]{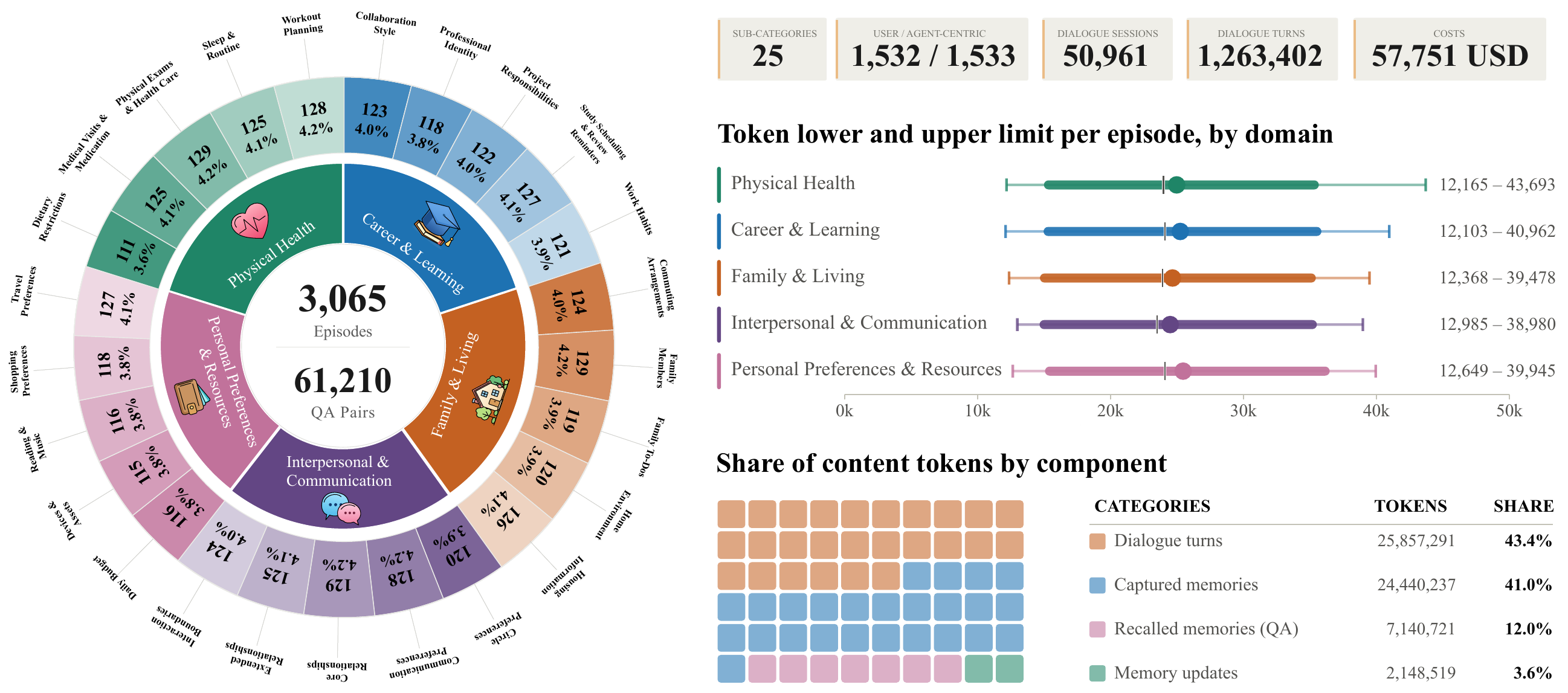}
    \caption{Dataset composition and scale of DyadMem across five major domains and 25 subdomains. Our benchmark shows balanced data composition and large-scale sessions and QAs.}
    \label{fig:dataset-statistics-main}
\end{figure*}

\subsection{Evaluation Framework}
\label{sec:evaluation-framework}


\textbf{Metrics.}
Capture, gold-conditioned standalone Update, and Recall are evaluated independently with gold targets or gold memory banks rather than chained into one component score. Capture uses micro-averaged precision, recall, and F1 over memory entries. Update reports operation-level F1 (op-F1), the macro-average of Add F1 and Delete F1, and is evaluated on the \(20{,}590\) sessions containing at least one gold Add or Delete operation, no-change sessions are excluded from this score. Recall receives \(q_j\) and the terminal gold bank \(B_T^\star\), then returns ordered memory indices \(\hat I_j\). With \(n_j=|I_j^\star|\), the primary metric is
\begin{equation}
\operatorname{Recall@}n
=
\frac{1}{J}
\sum_{j=1}^{J}
\frac{
\left|
I_j^\star
\cap
\operatorname{Top}_{n_j}(\hat I_j)
\right|
}{n_j}.
\end{equation}
We additionally report Recall@\(n{+}4\), Recall@All (set recall over all returned indices), retrieval precision, and Hit@1 (the fraction of questions whose first retrieved entry is gold). Because every QA instance has non-empty gold support, an empty retrieval is a failure rather than a correct abstention.

\textbf{Two QA settings.}
\emph{Gold-Memory QA} directly supplies the annotated gold support entries, measuring answer generation when the required evidence is available. \emph{Full-Pipeline QA} chronologically constructs the model's own bank, recalls up to \(K=16\) entries from it, and answers with those memories. On the same question set, we define the \emph{Gold-to-Pipeline gap} as
\begin{equation}
\Delta_{\mathrm{G2P}}
=
\operatorname{Acc}_{\mathrm{gold}}
-
\operatorname{Acc}_{\mathrm{pipe}}.
\end{equation}
This gap measures the overall performance loss introduced by the complete memory pipeline; it is not a causal decomposition of Capture, Update, Recall, and answer-generation errors. Those stages are assessed separately by the stage-level metrics above.

\textbf{Matching and QA scoring.}
Capture entries and Update additions use bidirectional semantic coverage so that evaluation tolerates differences in memory granularity; Recall and gold-conditioned standalone Update deletion use exact memory-index matching. QA answers are judged for semantic consistency with the GT answer by \texttt{GPT-5-mini}: correct and incorrect answers receive \(0\) or \(1\), respectively, and abstentions receive \(0\). \underline{Grounded accuracy} requires both positive correctness credit and support from the supplied memories and uses all questions as its denominator. \underline{Over-abstention (OA)} is the refusal rate over all questions, each of which is answerable. \underline{Ungrounded-correct (Ung.)} is the fraction of positively scored answers unsupported by the supplied memories.

Truncated, unparseable, empty, and failed outputs remain in the relevant denominator and receive no credit. Cross-model comparisons fix the prompts, judge, matcher, embedding model, and natural language inference (NLI) configuration. Confidence intervals use 200 bootstrap resamples. Matching thresholds, QA equations, failure handling, and bootstrap details are given in Appendix~\ref{app:method-details}.

%% file: Chapters/Chapter_4_Experiments.tex
\section{Experiments}

\subsection{Setup}

We evaluate 20 frontier LLMs on DyadMem: 16 open-weight models and 4 proprietary models. The default QA judge is \texttt{GPT-5-mini}, and the memory matcher uses BGE-M3 with natural language inference (NLI) adjudication. Open-weight models are deployed with vLLM on a multi-node A800 cluster. Full model, deployment, matcher, and auxiliary retrieval settings are listed in Appendix~\ref{app:experiment-details}. Two subsets support controlled analyses. \textbf{DyadMem-Agent} contains 500 agent-centric episodes and 10,074 QA instances. \textbf{DyadMem-Simple} is a balanced sample of 300 episodes and 5,935 QA instances, with 12 episodes from each of 25 subcategories; it is used for judge comparison and human calibration.

\subsection{Main Results}
\label{sec:main-results}

\begin{tcolorbox}[colback=blue!4!white,colframe=black!50!white,boxrule=1pt,boxsep=1pt,left=4pt,right=4pt,top=3pt,bottom=3pt]
\small
\textbf{Finding: Current models answer well when given the right memories, but lose substantial accuracy in the full pipeline; stage-level metrics expose failures in constructing, evolving, and retrieving memory.}
\end{tcolorbox}


\begin{table*}[t]
\centering
\vspace{-0.1in}
\setlength{\tabcolsep}{2.4pt}
\scriptsize
\caption{Main results on DyadMem. Capture $=$ F1/P/R; gold-conditioned standalone Update $=$ op-F1/Add/Del; Recall $=$ $@n$/@$n$+4/@All/P; QA $=$ Grounded/OA/Ung./Gold-Memory/Full-Pipeline. OA $=$ over-abstention, Ung.\ $=$ ungrounded-correct. {\setlength{\fboxsep}{1pt}%
\protect\colorbox{rose}{\strut rose},
\protect\colorbox{sky}{\strut sky}, and
\protect\colorbox{sage}{\strut sage}}
mark the top three; \textbf{Bold} and \underline{underlined} values mark the best and second-best.}
\label{tab:dyadmem-main}
\resizebox{\textwidth}{!}{%
\begin{tabular}{l ccc ccc cccc ccccc}
\toprule
 & \multicolumn{3}{c}{Capture}
 & \multicolumn{3}{c}{Update}
 & \multicolumn{4}{c}{Recall}
 & \multicolumn{5}{c}{QA} \\
\cmidrule(lr){2-4}\cmidrule(lr){5-7}\cmidrule(lr){8-11}\cmidrule(lr){12-16}
Model
 & F1$\uparrow$ & P$\uparrow$ & R$\uparrow$
 & op-F1$\uparrow$ & Add$\uparrow$ & Del$\uparrow$
 & @$n$$\uparrow$ & @$n$+4$\uparrow$ & All$\uparrow$ & P$\uparrow$
 & Grounded$\uparrow$ & OA$\downarrow$ & Ung.$\downarrow$ & Gold-Mem.$\uparrow$ & Full-Pipe.$\uparrow$ \\
\midrule
\multicolumn{16}{c}{\cellcolor{secH}\textit{Open-weight Models}} \\
Kimi-K3 & 28.14 & 58.27 & 18.55 & \cellcolor{rose}\textbf{31.69} & \cellcolor{sage}22.47 & \cellcolor{sky}\underline{40.92} & \cellcolor{rose}\textbf{62.09} & \cellcolor{rose}\textbf{70.14} & \cellcolor{sky}\underline{69.16} & 55.29 & \cellcolor{rose}\textbf{69.26} & \cellcolor{rose}\textbf{9.21} & 7.94 & \cellcolor{sky}\underline{94.99} & \cellcolor{rose}\textbf{74.85} \\
GLM-5.2 & 26.73 & 56.73 & 17.49 & 21.06 & 18.27 & 23.85 & 56.71 & 61.66 & 59.99 & 59.15 & \cellcolor{sky}\underline{67.50} & \cellcolor{sky}\underline{11.44} & \cellcolor{sage}5.51 & 94.30 & \cellcolor{sky}\underline{71.40} \\
DeepSeek-V4-Pro & \cellcolor{sky}\underline{30.70} & \cellcolor{sky}\underline{63.23} & \cellcolor{sage}20.27 & 22.48 & 16.80 & 28.16 & 59.32 & \cellcolor{sky}\underline{69.09} & \cellcolor{rose}\textbf{70.42} & 41.29 & 60.90 & 13.00 & 8.45 & 92.54 & 66.16 \\
MiniMax-M2.5 & 22.22 & 49.85 & 14.30 & 23.43 & 14.66 & 32.19 & 53.79 & 62.45 & 62.70 & 52.37 & 56.50 & 14.52 & 13.51 & 93.88 & 65.30 \\
GPT-OSS-120B & 14.71 & 61.09 & 8.36 & 23.88 & 13.44 & 34.32 & 47.19 & 50.17 & 48.33 & \cellcolor{sky}\underline{65.48} & 46.36 & 15.87 & 25.02 & \cellcolor{sage}94.70 & 61.64 \\
Qwen3.5-397B-A17B & 24.53 & \cellcolor{sage}62.39 & 15.26 & 22.31 & 19.53 & 25.10 & 57.26 & 64.71 & 63.40 & 61.05 & 51.21 & 24.93 & 6.70 & 92.11 & 54.88 \\
Qwen3.8-27B & 25.49 & 59.69 & 16.21 & \cellcolor{sage}30.18 & \cellcolor{sky}\underline{23.77} & \cellcolor{sage}36.59 & \cellcolor{sky}\underline{60.34} & \cellcolor{sage}67.47 & 66.87 & 56.74 & 49.35 & 26.81 & 8.33 & 93.17 & 53.81 \\
Qwen3-8B & 24.00 & 49.80 & 15.81 & 12.04 & 11.37 & 12.71 & 37.98 & 43.34 & 51.12 & 4.29 & 36.61 & 20.17 & 22.98 & 90.03 & 47.51 \\
Qwen3.5-27B & 25.67 & 60.25 & 16.31 & 23.79 & 19.02 & 28.55 & 59.32 & 66.99 & 65.82 & 56.05 & 43.89 & 32.37 & 6.59 & 92.13 & 46.96 \\
Qwen3-4B & 21.20 & 50.26 & 13.43 & 12.88 & 12.60 & 13.16 & 32.11 & 35.84 & 39.48 & 4.93 & 31.67 & 16.16 & 27.36 & 88.28 & 43.57 \\
Qwen3.5-9B & 23.23 & 54.74 & 14.75 & 16.39 & 14.93 & 17.85 & 51.55 & 62.28 & 64.97 & 33.78 & 34.18 & 43.10 & 10.30 & 91.25 & 38.08 \\
Llama-3.1-70B & 25.07 & 55.07 & 16.22 & 19.03 & 14.59 & 23.46 & 44.69 & 46.93 & 44.96 & \cellcolor{sage}62.66 & 33.02 & 42.80 & 12.20 & 84.66 & 37.56 \\
Qwen3.5-35B-A3B & 13.86 & 43.79 & 8.23 & 13.52 & 17.41 & 9.64 & 55.55 & 67.17 & \cellcolor{sage}68.89 & 31.27 & 26.64 & 54.28 & 8.74 & 89.37 & 29.18 \\
Qwen3.5-4B & 21.56 & 49.25 & 13.80 & 9.62 & 10.70 & 8.54 & 51.08 & 60.28 & 62.97 & 25.85 & 21.92 & 52.49 & 16.58 & 89.60 & 26.26 \\
Llama-3.1-8B & \cellcolor{sage}28.62 & 45.17 & \cellcolor{sky}\underline{20.95} & 4.44 & 6.52 & 2.37 & 38.69 & 42.93 & 42.09 & 23.82 & 20.07 & 52.77 & 15.84 & 78.69 & 23.84 \\
Qwen3-30B-A3B & 18.90 & 43.64 & 12.06 & 10.06 & 11.05 & 9.07 & 41.49 & 47.04 & 49.12 & 10.00 & 17.25 & 59.85 & 23.83 & 90.05 & 22.64 \\
Avg. & 23.41 & 53.95 & 15.13 & 18.55 & 15.45 & 21.66 & 50.57 & 57.41 & 58.14 & 40.25 & 41.65 & 30.61 & 13.74 & 90.61 & 47.73 \\
\midrule
\multicolumn{16}{c}{\cellcolor{secH}\textit{Proprietary Models}} \\
GPT-5.6-Sol & 21.71 & 56.78 & 13.42 & 25.54 & 20.73 & 30.35 & 56.06 & 61.39 & 59.54 & 62.37 & 63.81 & \cellcolor{sage}11.55 & 9.31 & \cellcolor{rose}\textbf{96.28} & \cellcolor{sage}70.35 \\
Doubao-Seed-2.1-Pro & 27.54 & 56.71 & 18.19 & \cellcolor{sky}\underline{30.68} & 20.32 & \cellcolor{rose}\textbf{41.04} & \cellcolor{sage}59.52 & 66.81 & 65.27 & 55.35 & \cellcolor{sage}65.95 & 18.16 & \cellcolor{sky}\underline{4.92} & 92.49 & 69.31 \\
GPT-4o-mini & 26.57 & 58.79 & 17.17 & 13.67 & 8.88 & 18.47 & 43.88 & 45.38 & 42.71 & 60.64 & 37.41 & 26.42 & 21.08 & 90.85 & 47.39 \\
Gemini-3.1-Pro & \cellcolor{rose}\textbf{36.37} & \cellcolor{rose}\textbf{72.49} & \cellcolor{rose}\textbf{24.27} & 27.99 & \cellcolor{rose}\textbf{26.66} & 29.31 & 58.91 & 64.61 & 62.41 & \cellcolor{rose}\textbf{66.23} & 40.66 & 42.11 & \cellcolor{rose}\textbf{4.00} & 85.24 & 42.35 \\
Avg. & 28.05 & 61.19 & 18.26 & 24.47 & 19.15 & 29.79 & 54.59 & 59.55 & 57.48 & 61.15 & 51.96 & 24.56 & 9.83 & 91.22 & 57.35 \\
\bottomrule
\end{tabular}%
}
\end{table*}

Table~\ref{tab:dyadmem-main} reports both stage-level quality and final QA. These views produce different rankings: a model can answer many questions while omitting memories or applying unsafe updates. We therefore interpret Capture, gold-conditioned standalone Update, and Recall with their own stage-level metrics, and use the Gold-to-Pipeline gap only to quantify the overall loss introduced by the complete memory pipeline.
\textbf{Recall.} Leading models retrieve at least one gold entry at rank one on 85--89\% of questions, yet their Recall@\(n\) remains below 63\%. Finding one relevant entry is therefore substantially easier than assembling the complete support set.
\textbf{Capture.} Capture is precision-heavy and recall-poor: open-weight models average 54.0\% precision but only 15.1\% recall. The resulting terminal banks are also small, about 75 entries at the median, compared with 323.6 in the gold bank, and model-to-gold size ratios usually lie between 0.20 and 0.35. This pattern indicates systematic under-extraction rather than indiscriminate storage.
\textbf{Gold-conditioned standalone Update.}
Gold-conditioned standalone Update averages 18.6\% op-F1 for open-weight models and 24.5\% for proprietary models. Scale does not resolve the problem: larger models are not consistently better, and the unsafe-deletion rate, the fraction of predicted deletions that are incorrect, averages 84.1\% and 78.1\% for the two groups (Table~\ref{tab:unsafe-delete-rate} and Figure~\ref{fig:unsafedeletion}). Because deletion changes the future contents of the bank, low precision at this stage propagates beyond the current session.
\textbf{End-to-end QA.}
Gold-Memory QA ranges from 78.7\% to 96.3\%, whereas Full-Pipeline QA averages 47.7\% for open-weight models and 57.4\% for proprietary models. The resulting Gold-to-Pipeline gaps are 42.9 and 33.9 points for the two groups. These are end-to-end gaps rather than a causal allocation of error to individual stages. \textbf{The low Capture recall, weak gold-conditioned standalone Update scores, and incomplete Recall reported above provide separate evidence that multiple upstream stages remain unreliable.}

Kimi-K3 illustrates why the final score alone is insufficient. It reaches the highest Full-Pipeline score, 74.9\%, despite 18.6\% Capture recall and 31.7\% gold-conditioned standalone Update op-F1; its 95.0\% Gold-Memory score yields a 20.1-point Gold-to-Pipeline gap. Moreover, its Grounded accuracy is 69.3\%, and 7.94\% of answers receiving positive correctness credit are unsupported by the retrieved memories. \textbf{Taken together, these results show that final QA can remain high even when the constructed memory bank is incomplete, unstable, or bypassed at answer time.}

\subsection{User-conditioned Relational Agent Memory versus Agent-only}
\label{sec:dyadic-vs-agent-only}


\begin{tcolorbox}[colback=blue!4!white,colframe=black!50!white,boxrule=1pt,boxsep=1pt,left=4pt,right=4pt,top=3pt,bottom=3pt]
\small
\textbf{Finding: Capturing and evolving user-conditioned memory (\texttt{user\_profile}, \texttt{fact}, and URAM) improves controlled Update for every model and downstream URAM-only QA for most models, relative to capturing and evolving general agent memory alone.}
\end{tcolorbox}

To validate the design of involving user conditions when constructing URAM, we compare two paired variants on the same 500 source episodes. For agent-memory Capture and Update, the URAM variant observes both user and assistant turns, whereas Agent-only observes both turns but captures assistant-only information. Both variants are evaluated against the same agent-side gold Update targets. Controlled URAM Update uses each variant's own model-extracted prior bank and current Capture output, with additions and deletions matched semantically. For the primary Full-Pipeline comparison, each variant evolves its own bank chronologically, and both answer the same 1{,}735 URAM-only questions from their own terminal banks. Table~\ref{tab:uram-conditional} gives the complete protocol.

\begin{table*}[t]
\centering
\vspace{-0.1in}
\setlength{\tabcolsep}{4.0pt}
\renewcommand{\arraystretch}{1.05}
\scriptsize
\caption{Controlled URAM comparison on 500 paired episodes. All entries are percentage-point differences.
$\Delta=$ User-conditioned Relational Agent Memory $-$ Agent-only.
Controlled URAM Update $=$ op-F1 / Add F1 / Del F1.
QA $=$ Full-Pipeline accuracy with Recall restricted to URAM on 1{,}735 URAM-only questions.
{\setlength{\fboxsep}{1pt}%
\protect\colorbox{rose}{\strut rose},
\protect\colorbox{sky}{\strut sky}, and
\protect\colorbox{sage}{\strut sage}}
mark the top three;
\textbf{Bold} and \underline{underlined} values mark the best and second-best.}
\label{tab:dyadmem-uram-delta}
\resizebox{0.56\textwidth}{!}{%
\begin{tabular}{l ccc c}
\toprule
 & \multicolumn{3}{c}{Update}
 & QA \\
\cmidrule(lr){2-4}\cmidrule(lr){5-5}
Model
 & $\Delta$(op-F1)$\uparrow$ & $\Delta$(Add)$\uparrow$ & $\Delta$(Del)$\uparrow$
 & $\Delta$(QA)$\uparrow$ \\
\midrule
\multicolumn{5}{c}{\cellcolor{secH}\textit{Open-weight Models}} \\
DeepSeek-V4-Pro & $+$3.37 & \cellcolor{rose}\textbf{$+$5.71} & $+$2.84 & \cellcolor{rose}\textbf{$+$9.77} \\
Qwen3.5-9B & \cellcolor{rose}\textbf{$+$4.98} & $+$3.33 & \cellcolor{sage}$+$4.49 & \cellcolor{sky}\underline{$+$7.21} \\
Qwen3.5-27B & $+$4.37 & \cellcolor{sky}\underline{$+$5.01} & $+$1.77 & \cellcolor{sage}$+$5.94 \\
MiniMax-M2.5 & $+$1.89 & $+$1.56 & $+$2.25 & $+$5.85 \\
Qwen3.5-35B-A3B & \cellcolor{sage}$+$4.51 & \cellcolor{sage}$+$4.82 & $+$3.79 & $+$5.85 \\
GPT-OSS-120B & $+$1.93 & $+$2.71 & $+$1.61 & $+$4.93 \\
Qwen3.8-27B & \cellcolor{sky}\underline{$+$4.85} & $+$1.62 & \cellcolor{rose}\textbf{$+$6.25} & $+$4.15 \\
Qwen3.5-397B-A17B & $+$2.89 & $+$1.33 & \cellcolor{sky}\underline{$+$4.84} & $+$3.72 \\
Kimi-K3 & $+$0.78 & $-$0.02 & $+$0.76 & $+$3.51 \\
GLM-5.2 & $+$0.10 & $-$0.20 & $+$0.93 & $+$3.17 \\
Llama-3.1-70B & $+$2.50 & $+$2.04 & $+$1.01 & $+$2.45 \\
Qwen3-8B & $+$3.55 & $+$2.95 & $+$3.37 & $+$1.90 \\
Qwen3.5-4B & $+$2.97 & $+$3.41 & $+$1.72 & $+$0.95 \\
Qwen3-30B-A3B & $+$1.34 & $+$1.81 & $+$0.41 & $-$1.64 \\
Qwen3-4B & $+$3.22 & $+$4.76 & $+$1.02 & $-$3.17 \\
Llama-3.1-8B & $+$1.22 & $+$1.41 & $+$0.64 & $-$3.32 \\
Avg. & $+$2.78 & $+$2.64 & $+$2.36 & $+$3.20 \\
\midrule
\multicolumn{5}{c}{\cellcolor{secH}\textit{Proprietary Models}} \\
GPT-5.6-Sol & $+$0.99 & $-$0.66 & $+$1.89 & $+$4.12 \\
Gemini-3.1-Pro & $+$3.23 & $+$3.38 & $+$3.80 & $+$3.95 \\
GPT-4o-mini & $+$1.76 & $+$3.41 & $+$1.22 & $+$3.91 \\
Doubao-Seed-2.1-Pro & $+$1.68 & $+$0.18 & $+$1.75 & $+$2.68 \\
Avg. & $+$1.92 & $+$1.58 & $+$2.17 & $+$3.67 \\
\bottomrule
\end{tabular}%
}
\vspace{-0.2in}
\end{table*}

In Table~\ref{tab:dyadmem-uram-delta}, the results show positive effects on all 20 models on the Update task. The primary QA difference is positive for 17 models; the mean gains are \(+3.2\) points for open-weight models and \(+3.7\) for proprietary models. DeepSeek-V4-Pro gains the most at \(+9.8\) points. The three negative QA point estimates belong to Llama-3.1-8B, Qwen3-4B, and Qwen3-30B-A3B. Their abstention rates decrease rather than increase, indicating that their losses arise on questions they already attempted; the full answer-pattern comparison is in Appendix~\ref{app:experiment-details}. 
The controlled Update result shows that a memory bank conditioned on the user supports more accurate agent-memory evolution than general agent memory alone. Converting that improvement into QA gains additionally requires the model to retrieve and use the resulting bank.
\subsection{User-side Memory Is Harder to Capture and Use}
\label{sec:user-agent-centric}


\begin{tcolorbox}[colback=blue!4!white,colframe=black!50!white,boxrule=1pt,boxsep=1pt,left=4pt,right=4pt,top=3pt,bottom=3pt]
\small
\textbf{Finding: Models answer agent-centric questions substantially better than user-centric questions, even though the corresponding Capture and Recall gaps are much smaller.}
\end{tcolorbox}

\begin{table*}[t]
\centering
\setlength{\tabcolsep}{3.2pt}
\renewcommand{\arraystretch}{1.05}
\scriptsize
\caption{User- and agent-centric results on DyadMem (3{,}065 episodes, 61{,}210 QA instances).
User-centric $n{=}30{,}468$; agent-centric $n{=}30{,}742$.
Capture $=$ F1; gold-conditioned standalone Update $=$ op-F1; Recall $=$ $@n$; QA $=$ Full-Pipeline.
{\setlength{\fboxsep}{1pt}%
\protect\colorbox{rose}{\strut rose},
\protect\colorbox{sky}{\strut sky}, and
\protect\colorbox{sage}{\strut sage}}
mark the top three;
\textbf{Bold} and \underline{underlined} values mark the best and second-best.}
\label{tab:dyadmem-target-split}
\resizebox{0.6\textwidth}{!}{%
\begin{tabular}{l cccc cccc}
\toprule
 & \multicolumn{4}{c}{User-centric}
 & \multicolumn{4}{c}{Agent-centric} \\
\cmidrule(lr){2-5}\cmidrule(lr){6-9}
Model
 & Cap.$\uparrow$ & Upd.$\uparrow$ & Rec.$\uparrow$ & QA$\uparrow$
 & Cap.$\uparrow$ & Upd.$\uparrow$ & Rec.$\uparrow$ & QA$\uparrow$ \\
\midrule
\multicolumn{9}{c}{\cellcolor{secH}\textit{Lower bounds \& references}} \\
No Memory & --- & --- & --- & --- & --- & --- & --- & --- \\
\midrule
\multicolumn{9}{c}{\cellcolor{secH}\textit{Open-weight Models}} \\
Kimi-K3 & 26.12 & \cellcolor{rose}\textbf{31.68} & \cellcolor{rose}\textbf{61.87} & \cellcolor{rose}\textbf{69.07} & 30.18 & \cellcolor{rose}\textbf{31.70} & \cellcolor{rose}\textbf{62.31} & \cellcolor{rose}\textbf{80.59} \\
GLM-5.2 & 24.56 & 21.20 & 56.01 & \cellcolor{sky}\underline{66.03} & 28.92 & 20.93 & 57.41 & \cellcolor{sage}76.73 \\
DeepSeek-V4-Pro & \cellcolor{sky}\underline{29.21} & 22.78 & 58.89 & 59.30 & \cellcolor{sky}\underline{32.21} & 22.19 & 59.75 & 72.96 \\
MiniMax-M2.5 & 21.11 & 23.25 & 53.84 & 58.22 & 23.36 & 23.61 & 53.75 & 72.32 \\
GPT-OSS-120B & 14.36 & 23.45 & 45.34 & 54.62 & 15.07 & 24.33 & 49.04 & 68.59 \\
Qwen3.5-397B-A17B & 22.66 & 22.44 & 57.52 & 46.86 & 26.42 & 22.18 & 57.00 & 62.82 \\
Qwen3.8-27B & 23.27 & \cellcolor{sage}30.03 & \cellcolor{sky}\underline{60.37} & 45.49 & 27.73 & \cellcolor{sage}30.34 & \cellcolor{sky}\underline{60.32} & 62.06 \\
Qwen3-8B & 22.58 & 11.99 & 37.48 & 37.36 & 25.44 & 12.10 & 38.48 & 57.58 \\
Qwen3.5-27B & 24.16 & 23.92 & \cellcolor{sage}59.74 & 38.84 & 27.20 & 23.65 & 58.91 & 55.00 \\
Qwen3-4B & 19.73 & 12.72 & 30.89 & 34.17 & 22.68 & 13.05 & 33.32 & 52.89 \\
Qwen3.5-9B & 21.32 & 16.09 & 52.26 & 29.74 & 25.17 & 16.70 & 50.85 & 46.35 \\
Llama-3.1-70B & 23.39 & 18.92 & 43.54 & 30.10 & 26.76 & 19.13 & 45.82 & 44.95 \\
Qwen3.5-35B-A3B & 12.39 & 13.44 & 56.26 & 21.72 & 15.34 & 13.61 & 54.85 & 36.57 \\
Qwen3.5-4B & 19.64 & 9.44 & 51.15 & 19.12 & 23.50 & 9.83 & 51.02 & 33.34 \\
Llama-3.1-8B & \cellcolor{sage}26.96 & 4.39 & 36.34 & 18.24 & \cellcolor{sage}30.30 & 4.50 & 41.02 & 29.39 \\
Qwen3-30B-A3B & 16.51 & 9.99 & 40.56 & 16.03 & 21.30 & 10.13 & 42.42 & 29.19 \\
Avg. & 21.75 & 18.48 & 50.13 & 40.31 & 25.10 & 18.62 & 51.02 & 55.08 \\
\midrule
\multicolumn{9}{c}{\cellcolor{secH}\textit{Proprietary Models}} \\
GPT-5.6-Sol & 20.15 & 25.87 & 54.46 & 62.18 & 23.27 & 25.19 & 57.66 & \cellcolor{sky}\underline{78.45} \\
Doubao-Seed-2.1-Pro & 25.88 & \cellcolor{sky}\underline{30.38} & 59.20 & \cellcolor{sage}65.06 & 29.21 & \cellcolor{sky}\underline{30.99} & \cellcolor{sage}59.83 & 73.52 \\
GPT-4o-mini & 24.87 & 13.67 & 43.07 & 38.52 & 28.30 & 13.68 & 44.69 & 56.19 \\
Gemini-3.1-Pro & \cellcolor{rose}\textbf{35.05} & 28.50 & 58.39 & 38.40 & \cellcolor{rose}\textbf{37.70} & 27.46 & 59.42 & 46.26 \\
Avg. & 26.49 & 24.61 & 53.78 & 51.04 & 29.62 & 24.33 & 55.40 & 63.61 \\
\bottomrule
\end{tabular}%
}
\vspace{-0.2in}
\end{table*}

The two partitions are nearly equal in size: 1,532 user-centric episodes contain 30,468 questions, and 1,533 agent-centric episodes contain 30,742. Table~\ref{tab:dyadmem-target-split} shows that open-weight Full-Pipeline QA differs by 14.8 points between the two sides, while Capture and Recall differ by only 3.4 and 0.9 points. Proprietary models show the same direction, with mean QA of 51.0\% on user-centric questions and 63.6\% on agent-centric questions.
Because the largest separation appears only at QA, the difficulty lies less in whether a relevant entry is captured or retrieved than in using user-side evidence reliably after the stages are composed.

\subsection{Memory Type, Not Life Domain, Sets Capture Difficulty}
\label{sec:memory-type-difficulty}


\begin{tcolorbox}[colback=blue!4!white,colframe=black!50!white,boxrule=1pt,boxsep=1pt,left=4pt,right=4pt,top=3pt,bottom=3pt]
\small
\textbf{Finding: Memory type explains substantially more variation in Capture than the everyday-life domain in which the interaction occurs.}
\end{tcolorbox}

Across the 20 models, Figure~\ref{fig:sixmemorytype} shows that mean Capture F1 is about 50\% for \texttt{agent\_identity} and \texttt{agent\_method}, 25\% for \texttt{fact}, and 16\% for shared \texttt{episodic} memory. By contrast, Figure~\ref{fig:25domains} shows that capture F1 spans only 22.7--26.3\% across the five life domains. Interpersonal \& Communication has the lowest Capture F1 at 22.7\% while retaining 50.1\% Full-Pipeline QA, showing that weak extraction need not translate one-for-one into failed answers. The large type gap indicates that models capture explicit, reusable agent-side rules more reliably than heterogeneous factual and episodic content, the domain topic itself contributes comparatively little.

\subsection{Judge Reliability}




\begin{tcolorbox}[colback=blue!4!white,colframe=black!50!white,boxrule=1pt,boxsep=1pt,left=4pt,right=4pt,top=3pt,bottom=3pt]
\small
\textbf{Finding: Model rankings are stable across automatic judges, and human labels show substantial agreement with the default judge.}
\end{tcolorbox}

\vspace{0.7em}
\noindent
\begin{minipage}[t]{0.46\textwidth}
  \vspace{0pt}
  \centering
  \includegraphics[width=0.9\linewidth]{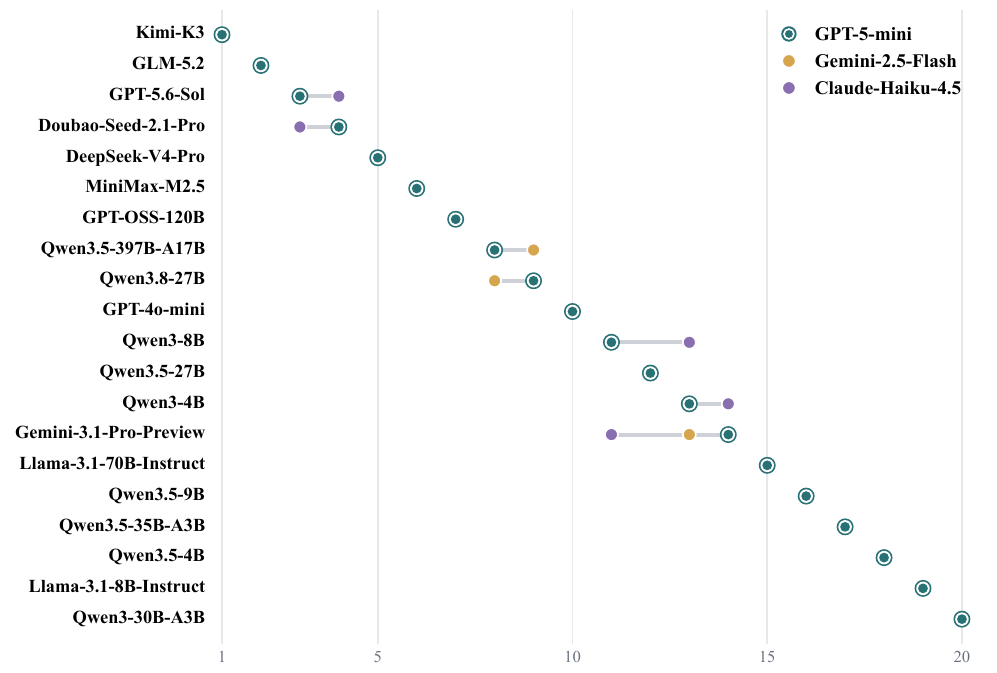}
  \captionof{figure}{Model rankings by Full-Pipeline QA on DyadMem-Simple under three judges.}
  \label{fig:judge-pipeline-qa}
\end{minipage}%
\hfill
\begin{minipage}[t]{0.52\textwidth}
  \vspace{0pt}
  \centering
  \captionof{table}{LLM--human agreement on DyadMem-Simple Full-Pipeline QA ($n{=}5{,}935$). Human is gold. Acc./P/R/F1 in \%.}
  \label{tab:human-judge}
  \setlength{\tabcolsep}{2.4pt}
  \scriptsize
  \resizebox{\linewidth}{!}{%
  \begin{tabular}{l cccccc}
    \toprule
    & Acc.$\uparrow$ & Cohen's $\kappa$$\uparrow$ & MCC$\uparrow$ & P$\uparrow$ & R$\uparrow$ & F1$\uparrow$ \\
    \midrule
    Auto vs.\ Human & 86.52 & 0.7099 & 0.7099 & 81.53 & 81.76 & 81.64 \\
    \bottomrule
  \end{tabular}}

  \vspace{0.55em}
  \resizebox{\linewidth}{!}{%
  \begin{tabular}{l cc c}
    \toprule
    & Human consistent & Human inconsistent & Row total \\
    \midrule
    Auto consistent & 1{,}779 (TP) & 403 (FP) & 2{,}182 \\
    Auto inconsistent & 397 (FN) & 3{,}356 (TN) & 3{,}753 \\
    \addlinespace
    Column total & 2{,}176 & 3{,}759 & 5{,}935 \\
    \bottomrule
  \end{tabular}}
  \vspace{0.35em}
\end{minipage}

\vspace{0.8em}

Figure~\ref{fig:judge-pipeline-qa} shows a nearly unchanged model ordering across the three judges: they identify the same top seven models, no model moves by more than three positions, and all pairwise Spearman correlations \citep{spearman1904proof} are at least 0.988.
Five selected annotators provide paired human labels for 5,935 outputs. Table~\ref{tab:human-judge} shows 86.52\% agreement and Cohen's \(\kappa=0.7099\) \citep{cohen1960kappa}, while false-positive and false-negative rates are nearly symmetric. This calibration covers the semantic-consistency decision used for QA accuracy, and together with the stable cross-judge ranking supports using \texttt{GPT-5-mini} for the reported model comparisons.

\subsection{Answering Style Changes Full-Pipeline QA}


\begin{tcolorbox}[colback=blue!4!white,colframe=black!50!white,boxrule=1pt,boxsep=1pt,left=4pt,right=4pt,top=3pt,bottom=3pt]
\small
\textbf{Finding: Full-Pipeline QA reflects a model's willingness to answer as well as the quality of its memory, and can therefore reverse the ranking implied by stage-level metrics.}
\end{tcolorbox}

Qwen3-4B scores 43.6\% on Full-Pipeline QA, above Qwen3.5-9B at 38.1\%, although Qwen3.5-9B is stronger on Capture, Recall, gold-conditioned standalone Update, and Gold-Memory QA. The reversal tracks their answering styles. Qwen3-4B has 16.2\% over-abstention and 27.4\% ungrounded-correct, whereas Qwen3.5-9B has 43.1\% over-abstention and 10.3\% ungrounded-correct. When correctness additionally requires retrieved support, Qwen3.5-9B instead leads 34.2\% to 31.7\% in Grounded accuracy (Figure~\ref{fig:errorheadmap}).

The pattern extends beyond this pair: the reversal appears in both QA-target partitions and all five life domains, while over-abstention ranges from 9.2\% to 59.9\% across models. Kimi-K3 shows that frequent answering need not be ungrounded: it combines 9.2\% over-abstention with 69.3\% Grounded accuracy and 7.9\% ungrounded-correct. Additional retrieval and answered-question statistics are reported in Appendix~\ref{app:experiment-details}.
Full-Pipeline QA must therefore be interpreted jointly with Grounded accuracy, over-abstention, and ungrounded-correct to distinguish memory-supported answers from conservative refusal or unsupported success.





%% file: Chapters/Chapter_5_Conclusion.tex
\section*{Conclusion}

We introduced DyadMem, a long-term cross-session memory benchmark that introduces User-conditioned Relational Agent Memory and a full-pipeline evaluation framework with extensive gold memory. Its lifecycle annotations and stage-level evaluations reveal several findings that final QA alone obscures: models answer well from gold evidence, yet lose substantial accuracy in the full pipeline while under-capturing memories, incompletely retrieving support, and applying unreliable updates. The controlled comparison further shows that capturing and evolving user-conditioned memory, rather than general agent memory alone, yields positive controlled Update differences across all evaluated models and improves downstream URAM-only QA on average. DyadMem therefore provides a concrete test bed for developing memory systems that preserve not only what an agent knows about a user, but also how that particular agent should work with them over time.

%% file: Appendix/0_Appendix.tex
\section*{\textcolor[RGB]{176,32,32}{Appendix}}

\startcontents[app]
\begingroup
  \hypersetup{linkcolor=appred}
  \printcontents[app]{}{0}{\setcounter{tocdepth}{2}}
\endgroup

\appmodule
  [AI use, ethics, and reproducibility]
  {Statements}

\section{LLM Use Statement}
\label{app:ai-use}
\input{Appendix/LLM_use_statement}


\section{Reproducibility Statement}
\label{app:reproducibility}
\input{Appendix/Reproducibility_statement}

\appmodule
  [Task definition, data sources, annotation protocol, construction prompts, and samples]
  {Data Construction and Annotation}

\section{Data Sources}
\label{app:data-source}
\input{Appendix/Data_Source}

\section{Annotator Information and Workflow}
\label{app:annotators}
\input{Appendix/Annotator_Information_and_Workflow}

\section{Prompts for Data Construction}
\label{app:construction-prompts}
\input{Appendix/Prompts_Used_in_Data_Construction}

\section{Data Samples}
\label{app:data-samples}
\input{Appendix/Data_Samples}

\appmodule
  [Method and evaluation details, experimental configuration, and the model inventory]
  {Method and Experimental Setup}

\section{Additional Method and Evaluation Details}
\label{app:method-details}
\input{Appendix/Additional_Method_and_Evaluation_Details}

\section{Additional Experimental Details}
\label{app:experiment-details}
\input{Appendix/Additional_Experiment_Details}


\appmodule
  [Metrics, evaluation code, and system prompts used at evaluation time]
  {Evaluation Protocol and Implementation}

\section{Evaluation Metrics}
\label{app:eval-metrics}
\input{Appendix/Evaluation_Metrics}

\section{Evaluation Code}
\label{app:eval-code}
\input{Appendix/Evaluation_Code}

\section{Prompts for Framework Evaluation}
\label{app:test-prompts}
\input{Appendix/Prompts_for_Framework_Evaluation.tex}

\appmodule
  [Qualitative outputs and error-case visualizations]
  {Evaluation Results and Analysis}

\section{Qualitative Model Results}
\label{app:qualitative-results}
\input{Appendix/Qualitative_Model_Results}


\appmodule
  [Model cards and licensing]
  {Model and License Information}

\section{Model Cards}
\label{app:model-cards}
\input{Appendix/Model_Cards}

\section{Licensing}
\label{app:license}
\input{Appendix/License_Information}

%% file: Appendix/LLM_use_statement.tex
We disclose the use of Large Language Models (LLMs) in this research. LLMs were used both as subjects of our benchmark experiments and for synthetic data generation and automated evaluation. During manuscript preparation, we also used LLMs for grammatical correction and stylistic refinement. LLMs were not involved in determining the core research ideas, methodology, experimental design, or scientific conclusions. All technical contributions and analyzes presented in this work are the original work of the authors.

%% file: Appendix/Reproducibility_statement.tex
We provide detailed descriptions of the benchmark construction process, data generation pipeline, evaluation protocols, model configurations, evaluation code, and prompts in the main paper and supplementary materials. The associated dataset and evaluation pipeline will be publicly released to facilitate reproducibility, subject to the applicable licenses of the underlying models and software.

%% file: Appendix/Data_Source.tex
The dataset is entirely constructed from synthetic user--agent interactions generated by Large Language Models. Rather than relying on private user logs or directly collected real-world conversations, we employ a controlled, multi-stage generation pipeline to construct diverse user personas, dialogue scenarios, and memory update processes. The generated interactions are designed to capture realistic patterns of user preferences, instructions, behavioral changes, and interactions with AI agents over multiple sessions. To improve data reliability and reduce generation artifacts, the resulting data are further subjected to automated quality checks, human review, and sampling-based validation. This controlled generation process enables us to systematically construct and evaluate agent-side memory capture and update while avoiding the use of private or personally identifiable user data.

%% file: Appendix/Annotator_Information_and_Workflow.tex
Quality control was done by nine annotators on a web page built for this project. Dialogues in the benchmark are in English. Before review we constructed 4,000 episodes, split evenly across the five life domains, so that episodes removed later would not have to be constructed again from the same distribution. Those episodes were assigned by domain to annotators 1--5. Each of them owned one domain, and every episode had one primary annotator. The review kept 3,065 episodes and dropped 935, a retention rate of 76.6\%. Episodes that passed the first pass entered a second-review pool. Annotators 6, 7, and 8 did not read that pool through by domain. Each drew a random sample of 500 episodes. The three samples overlap on purpose. An episode in the overlap was read independently by at least two of these reviewers, so the primary judgment and the second pass could be compared. Overlapping episodes are not counted a second time in the coverage total. Annotator 9 aggregated the full set, tracked progress, and reviewed an additional sample.

All nine worked full time, on quality control and on the review page. Annotator 7, who has a master's degree in computer science, led the page: requirements, implementation, revisions, the stress test, and local deployment. The page runs only on a local machine, and accounts were issued only to the quality-control staff.

\begin{table}[h]
\centering
\small
\setlength{\tabcolsep}{3.5pt}
\renewcommand{\arraystretch}{1.12}
\caption{The nine quality-control annotators. All worked full time for two months.}
\label{tab:annotator-roles}
\begin{tabular}{@{}c p{0.31\linewidth} p{0.20\linewidth} p{0.31\linewidth}@{}}
\toprule
ID & Role & English & Degree \\
\midrule
1 & Career \& Learning & Reads English & B.S.\ in computer science \\
2 & Family \& Living & With a translation & M.S.\ in software engineering \\
3 & Interpersonal \& Communication & With a translation & M.S.\ in artificial intelligence \\
4 & Personal Preferences \& Resources & Reads English & M.S.\ in data science \\
5 & Physical Health & With a translation & M.S.\ in computer science \\
6 & Second-pass sample & Reads English & B.S.\ in software engineering \\
7 & Platform development & Reads English & M.S.\ in computer science \\
8 & Second-pass sample & With a translation & B.S.\ in information science \\
9 & Sample review & Reads English & M.S.\ in computer science \\
\bottomrule
\end{tabular}
\end{table}

All nine have programming training. The group includes bachelor's and master's degrees in computer science, software engineering, artificial intelligence, information science, and data science. Five annotators read the English directly. Four read it next to a translation. The page keeps the English source and can show a translation beside it. An annotator who wants a translation enters their own API key and calls GPT-5.5. The translation is a reading aid. Adds and deletes are decided from the English source.

\begin{figure*}[t]
\centering
\includegraphics[width=\textwidth]{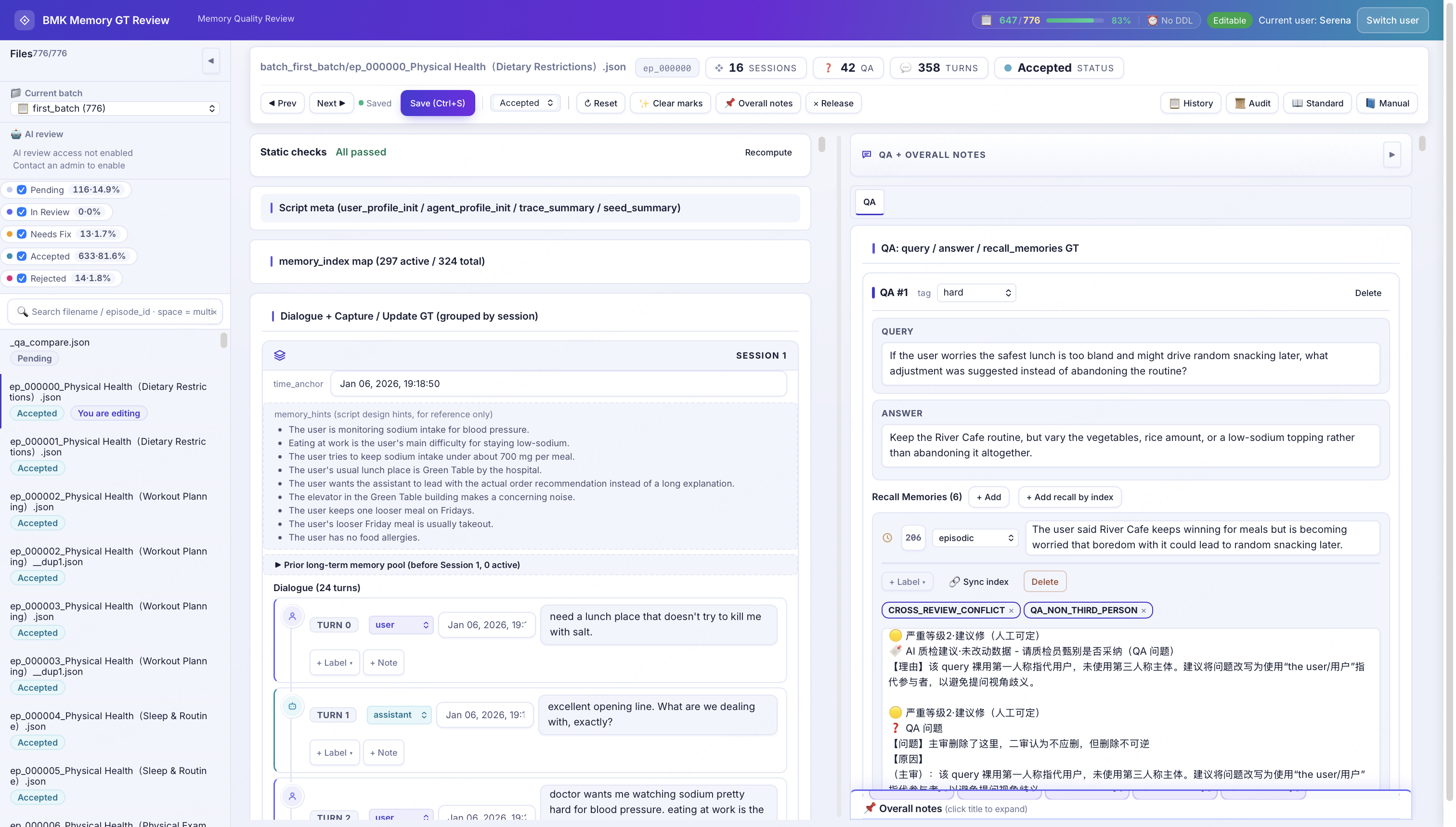}
\caption{The quality-control page for one accepted Physical Health episode. The center pane shows the dialogue and the session-level Capture and Update labels. The right pane shows a question, its answer, and the gold recall memories. Status for the batch is in Table~\ref{tab:qc-batch}.}
\label{fig:qc-platform}
\end{figure*}

For each episode the page shows the original dialogue beside the memory bank that is still valid after the sessions are replayed in order. Suspected errors are shown as hints and are not entered as gold labels. An edit to one memory is submitted together with the affected indices, the lineage, and the related QA items. Annotator 7 built this version first for annotators 1--5 to try. The 100 pilot groups below were assigned on it. If the source of an entry was hard to see, if an Update's additions and deletions were hard to see, if a memory edit did not carry its QA items, or if the translation covered the English, the page was revised and the episode was reviewed again. The version was frozen once the primary annotators could review on their own. A stress test then had all nine annotators opening episodes, requesting translations, saving, and sampling at the same time. Episodes that failed the checks were not written into the final set. After the test passed, the page was deployed locally, and annotator 7 moved to the second-round sample in Table~\ref{tab:annotator-roles}. Figure~\ref{fig:qc-platform} is the local review page used for that pass. The episode open in the figure has 16 sessions, 358 turns, and 42 questions. Its memory index lists 297 active entries out of 324, and the static checks on that file all pass. The left column is the batch the reviewer is working through: 776 files, with the status counts in Table~\ref{tab:qc-batch}. Accepted files in this batch are 633 out of 776. These counts are the queue on the page, not the 4{,}000-episode construction pool or the 3{,}065 episodes kept in the benchmark.

\begin{table}[t]
\centering
\small
\setlength{\tabcolsep}{8pt}
\caption{Status of the 776 files in the batch shown in Figure~\ref{fig:qc-platform}.}
\label{tab:qc-batch}
\begin{tabular}{l r r}
\toprule
Status & Files & Share \\
\midrule
Pending & 116 & 14.9\% \\
In review & 0 & 0\% \\
Needs fix & 13 & 1.7\% \\
Accepted & 633 & 81.8\% \\
Rejected & 14 & 1.8\% \\
\midrule
Total & 776 & 100\% \\
\bottomrule
\end{tabular}
\end{table}

\subsection{Review procedure}

Before the formal assignment we ran a pilot. We drew a stratified random sample of 100 groups and assigned them by the roles in Table~\ref{tab:annotator-roles}. Disputed items were decided one by one. That pass settled how memories are typed, how Update operations are judged, and how QA support sets are judged. Primary review uses the same rules. Dialogue and gold labels are still revised during review. Whether a memory is kept, and which type it has, often depends on the surrounding turns. Written rules do not cover those cases, so the training is done in the pilot and in the formal review. Primary annotators first apply the current rules to a batch. The others compare those results and point out disagreements on the spot. When a case is ambiguous, the annotators and the authors read the original English dialogue together and then decide.

In the formal review an annotator checks four things: 1. Time and roles stay consistent across the episode. 2. Capture for the session extracts the states that should be kept. 3. Each Update has dialogue evidence, including the deletions and additions that evidence calls for. 4. The QA reference answer is supported only by valid entries in the gold support set.

Rules and models only mark suspected errors. They do not write gold labels. If one memory changes, the affected indices, lineage, and related QA items change with it, and the evidence chain is checked for consistency again. An episode that passes enters the second-review pool and is not added to the final set at this stage. An episode that fails enters neither.

Who conducts the second pass, and how large each sample is, is given in Table~\ref{tab:annotator-roles}. The second pass does not require a full reading of the pool. An error found in the sample is corrected under the same rules. The episode enters the final set only after it passes the consistency check again.

\subsection{Recruitment, Compensation, and Consent}

We recruited people with programming training who could check a dialogue and revise indices and annotations from the evidence. Two schedules were offered. Full-time work ran for two months, five weekdays a week, eight statutory working hours a day. Part-time work ran for three months, at least three days a week, with the same eight-hour day. All nine people on this project took the full-time schedule. Pay was above the local minimum wage and followed the going rate for annotation of this kind. Some annotators are authors of this paper. Review done by the authors was not paid at this rate. The other participants were.

Before work began we explained the aim of the study, and that the annotations would be written into the benchmark and used in later experiments. The materials under review are generated dialogues, together with the memory entries and questions inferred from them. Quality control edits those annotations. It does not collect or rewrite the participants' own life records. API keys are not included in the released data. For these reasons we did not submit the study for formal Institutional
Review Board (IRB) review. The released data do not contain participant names. During review, a participant may withdraw from any part that is not yet finished.

%% file: Appendix/Prompts_Used_in_Data_Construction.tex
\begin{promptbox}{Persona Rewrite Prompt}
You rewrite a one-line user persona for a synthetic dialogue dataset.
Keep the ORIGINAL persona's identity, profession, and life background fully intact.
On top of it, layer the traits of a POWER USER of an AI conversational assistant:
  (1) habitually delegates recurring and everyday tasks and expects the assistant
      to automate them and follow up proactively;
  (2) sets and maintains long-term rules for how the assistant should work ---
      their professional conventions, personal routines, and output preferences ---
      correcting and updating them over time;
  (3) treats it as a persistent cross-domain external brain (work, health, finance,
      planning, learning), relying on it to remember prior context, preferences and
      decisions and often referring back to earlier conversations;
  (4) routinely consults its opinion and judgment before acting.
Blend these traits naturally into the persona; do NOT enumerate them as a list.
Output a concise 1-2 sentence English persona.
Return ONLY JSON:{"persona": "..."}.
\end{promptbox}

\begin{promptbox}{Freeform Dialogue Generation Prompt}
---- (1)Task / roles / safety and compliance ----
== Task ==
- You are generating training data for a "long-term-memory benchmark". Your task: write one whole stretch of an utterly real chat log between a real human user and an AI assistant.
== Roles ==
- user: the one real user here. An ordinary person with a temper and moods, who goes off topic and vents.
- agent: an AI assistant with a personality and cross-time memory, able to naturally recall things the user said before.
- The user knows this is an AI that can respond at any time, so there is no need to open every new session with "you there?" or to check whether the agent is free. Most new sessions should go straight to a status, a question, or picking up an earlier thread.
- The agent has no body, no address, no offline schedule and no real private life. It must never claim it just traveled, ate, slept, went to work, bought something, met a friend, was ill in person, or attended a real-world event; it must not invent family, friends, childhood, pets or coworkers.
- The agent may only claim the language and analysis work actually done inside the conversation; without tool evidence it must not say it just went to look at something in person, made a phone call, tried something on, tasted, ran, or already carried out a real-world action for the user.
== Safety and compliance (hard constraints) ==
[Safety & compliance hard rules]
1. All content must be suitable for a publicly released benchmark dataset:
   strictly no pornography or sexual innuendo, graphic violence, gambling,
   drugs, crime, extremism, hateful attacks, self-harm or suicide --- no such
   themes, hints or details.
2. Scenarios, character backgrounds and plot developments stay healthy,
   positive and reasonable: never introduce non-compliant or high-risk content
   (crime, addiction, dangerous behavior, etc.) for extra complexity or drama;
   dramatic tension may only come from normal life friction (scheduling,
   budget trade-offs, changing preferences, method adjustments, collaboration
   friction, etc.).
3. Dialogue stays within normal everyday scenes: work, study, daily life,
   relationships, hobbies, health management, personal planning; never
   generate risk-oriented dialogue.
4. On sensitive-adjacent topics, stay at reasonable informational discussion
   or safe-management level: medication only as doctor-directed time/dose
   record keeping, never self-adjustment or dangerous-usage guidance; staying
   up late / drinking may appear only as habits to improve; never output
   dangerous procedures, encouraging framings or inappropriate detail.
---- (2) Conversation structure and style ----
== Conversation structure: strict alternation + punctuation (hard constraints) ==
- **Strict alternation**: user and agent post one message each, and **never two consecutive messages from the same speaker**. When the user wants to say several things in a row, put those sentences into **one message**, and only then does the agent take the turn.
- **Punctuation must be natural and correct**: it should follow the habits of casual spoken English, never padded or piled on. Go light on bookish marks in chat, **especially avoid em dashes**, and do not overuse ellipses.
== user and agent conversation dynamics ==
- Never let the conversation read like an agent-led questionnaire: no mechanical agent questions, no mechanical user answers, no Q&A loop that feels like working through a survey.
- Avoid pure-atmosphere lines and information-free pleasantries wherever possible. Most messages should carry at least a little understandable state, attitude, question, decision, suggestion or progress, so later memory extraction has real material to work from.
- Keep overall information density low to medium: most user messages are small steps of a single sentence, not a status report every turn. A few longer messages are allowed, especially when the user is telling a whole story, explaining how something came about, venting at length, weighing options, or raising several related asks at once. Long messages must sound like something spoken, not like an essay, a survey answer, or facts listed out to stuff in memory points.
- On the user side, venting at the agent, complaining, switching topics abruptly, and picking a thread back up after a long absence are all allowed (mind the plausible jumps in timestamps).
- The user side may lay out their own situation in detail to ask the agent for advice, and may let it come out piecemeal in the form of complaints and venting.
- **Off-topic chitchat is mandatory and written to a quota**: off-topic chitchat should take about 15
== How the user talks ==
- Language: must be expressed in **English**, and it must be natural, idiomatic English; never carry Chinese-style fillers, particles or sentence patterns over literally.
- Form: colloquial lines with a real-person texture. No written register, no mini-essays, no report voice; strictly no long stretches of setting-style self-description and no one-shot long dumps.
- Colloquial: the user is a real person **typing in a chat**. Interjections and verbal tics ("uh", "hmm", "yeah", "tbh", "ngl"), idiomatic contractions (wanna, gonna, kinda, dunno), rambling, tangents, self-deprecation, rhetorical questions, venting and slang are all welcome, along with spoken turns of phrase like "anyway". Occasional moods (irritated, hesitant, resigned) are welcome, and every mood shift **needs a reason traceable in the context; moods never swing out of nowhere**. Information may come out half now and half next turn, with no need to say it all at once.
- Real imperfection: real people don't type tidily. Thinking out loud, loose word order and mid-line course-corrections are allowed ("i'm... oh wait, thursday works"), as are additions or corrections to the previous line, and conversational punctuation (ellipses, doubled question or exclamation marks). Not every line should be a grammatically complete, airtight sentence; that reads as AI. **But the looseness is grammatical only**: the thought itself still has to come out whole. No telegram-style fragments with the grammar squeezed out, and never leave the reader to fill in a gap; dropped subjects and half-finished lines show up only in ultra-short spots like making a choice or acknowledging something, never as the default.
- Length: **the vast majority of user messages are a single sentence** (roughly 8~20 words), and that is the default; a minority run to two sentences (roughly 25~45 words). **Only a very small number of three-or-four-sentence turns are allowed in the whole episode (3~5 at most, never a paragraph)**, and only where the user states strict preconditions, lays out the background of an event or the people in it, or explains their own thinking, that is, where more detail lets the agent answer better. **Never two consecutive turns longer than one sentence.** Two kinds of ultra-short turns also apply: making a simple choice or answering a simple question from the agent ("A works", "no idea"), and asks that are themselves short ("what time is it"); bare acknowledgements are normal too ("yeah ok", "right"). Length must ebb and flow the way real speech does.
- AI voice: strictly no service-speak on the user side ("Certainly! Here are a few...", "Is there anything else I can help with?"), no all-purpose openers like "Sure," or "Of course!", and **no piled-up bookish connectors ("Furthermore", "Moreover", "In conclusion")**. A user stating their own list of needs ("I want: first..., second...") is normal human behavior and not a violation. Never self-identify as an AI, and no disclaimer-speak.
- Fillers and typos: verbal tics and filler words ("uh", "i mean"), self-interruptions and mid-sentence course-corrections ("ugh, never mind"), light net-speak and the occasional emoticon (sparingly), and the rare stray typo from a slipped key (a typo may only land on a non-anchor word). Readability comes first: never drown a whole line in emoji or typos.
- Colloquial granularity: get the whole thought out inside the sentence. Do not let a clever compressed phrase or label stand in for what should be spelled out, and do not chop a complete statement into several clipped fragments.
- References to earlier context: when mentioning people and things that appeared earlier, a real person naturally uses references ("that thing yesterday", "the guy you just mentioned", "that method you said"), rather than restating the background like a document. But references may only point at people, objects, events or times explicitly mentioned earlier, and they must still carry enough anchoring; do not leave things vague. **Hard guardrail**: anchor information such as names, numbers, dates, conditions and proper nouns must still be stated clearly and correctly in the line itself, and an anchor must never be omitted, wrong or blurred.
- Lived-in: concrete details, names, places, quoted snippets of what other people said, small in-the-moment scenes ("my mom called three times already this morning"). Strictly no vague-but-correct "plastic dialogue": abstract self-statements no real person would blurt out, like "I really value communication efficiency" or "I want to improve my expression skills", are banned outright.
- How hard to answer the agent: the user need not respond to every point the agent made, and may pick up only the half they care about and let the rest go. No excessive politeness either: not every line needs a thank-you, and no repeated "thanks so much" or "sorry to bother you". A real person can be blunt, can be impatient, and can skip the pleasantries and go straight to the point.
- Role-reversal test: if some user line would sit in the agent's mouth verbatim with zero friction, that means it is too AI and must be rewritten to sound more human.
== Layered user intent ==
- Most turns still keep a single small intent, but a few user messages may carry 2~3 related intents at once, for instance reporting progress while pressing for a reason and adding one more constraint. Conditional, hesitant, comparative and self-correcting intents are also fine.
- Multiple intents must come from one natural scene, never stitched together like test items. The agent should sort out what is urgent first, and where the user's intent is too complex it may answer across several turns rather than covering everything in one message.
== How the agent talks ==
- Language: must be expressed in **English**.
- Default register (factory setting): the agent always answers in tidy, grammatical complete sentences with correct punctuation, standard capitalization and clean prose, and may carry a bit of that careful AI texture. **It leans long and detailed**: a substantive reply should talk the thing through, covering the reasons, the background, the concrete options, the trade-offs and the next step, developed fully rather than dispatched in a telegram-style line or two, and never as subject-dropping fragments. **Where the agent's length comes from**: genuinely spelling things out in complete sentences, thorough where thoroughness is called for and structured where structure is called for. At the nodes that need it the agent may use bullets, and may close with a single summarizing sentence (say, when handing over several parallel options at once, laying out a multi-step procedure, or closing out a stretch of work); but bullets and summaries must never become a per-turn ritual, and length must never be padded by restating the user's own words or bolting on pep-talk endings. **In pure venting or chitchat**: the agent reads the moment and need not pad, and must not mirror the user's punctuation-dropping, fragmented texting style just to seem casual.
- **Style is user-switchable** and then persists: the only thing that can override the default register above is an explicit user instruction about how to write ("keep it short", "stop being so wordy", "just give me the answer, skip the preamble", "write it out in full, give me two options each day"). The moment the user says so, the agent complies, and **keeps** that style across every later session until the user says otherwise. This is a long-term agent_identity rule, not a one-off adjustment. Terseness, bullets or a looser register that the user specifically asked for count as following instructions, not as default AI voice.
- AI voice: no customer-service sign-offs like "Is there anything else I can help you with?"; where the user never asked for it, no self-appointed titles ("as your X assistant", "as your long-term partner"). Never write the agent as a quip machine landing a zinger every line: most replies should read like a real person simply responding. No polished metaphors, aphorisms, summary flourishes or forced callbacks. Do not lean on template openers like "It sounds like you...", "That is completely normal", "We could...". Do not restate the user before every piece of advice, and do not strike the same "full analysis + bullet list + perfect wrap-up" pose turn after turn (use it at the nodes that call for it, and plain flowing sentences everywhere else).
- Distillation: over the course of talking with the user, the agent may distill long-term, stable, reusable lessons and methods, the goal being that the agent understands the user better and better and becomes more and more useful.
- Judgment and preferences: where the agent thinks the user's plan is unwise it may say it disagrees and give its reasons; where it cannot judge or cannot offer a recommendation it says so plainly. It keeps an objective stance, never fudges an answer, and never deflects with a counter-question. Avoid both fawning and vague, lofty lecturing.
- Anti-template: never lock into a fixed "affirm/empathize + advice + counter-question" three-beat every turn; never reuse the same opening and closing formulas across consecutive sessions; never echo the user's words verbatim before attaching advice. Response structure must genuinely vary in length, detail and tone. Keep the middle and late stretches of multi-session data from sliding into templates.
---- (3) Plot and time: consistency, readability, the physics of time ----
== Global plot naturalness ==
- **Global fact consistency (hard constraint)**: once written, names, forms of address and family relations, item prices, quantities and states, places, budgets and date spans hold for the whole episode, and later mentions must match the earlier text exactly. No renaming mid-episode (an earlier "grandma" becoming "nana" later), no changed numbers, no changed states. Character settings such as gender, job, family and city stay consistent throughout, and pronouns stay consistent with gender. Where money and quantities are involved, look back at the earlier text and make the books balance before writing: money already spent, budget left, bought or not bought must all add up. A fact may change only through an explicit evolution plot, which must spell out how it went from old to new.
- **Global plot consistency (hard constraint)**: when the plot refers to an earlier event, make sure that event really happened earlier in the text. Do not force in things that never happened just to show off memory, such as an out-of-nowhere "you said last time". Every "you said last time", "didn't you ask", "as we decided before" must point at content that truly appeared earlier, and dialogue that never happened must never be cited. Once a decision or promise is made, later behavior must remain compatible with it unless the plot explicitly overturns it and says why. The user's emotions, decisions and actions need understandable triggers and consequences; after a suggestion is accepted, rejected or carried out, later developments must stay consistent with that reaction, never as if the script had been swapped.
- **Time must flow (hard constraint)**: this chat was not done in one sitting; it is **the same user coming back to this assistant repeatedly over several days and several visits** (roughly {sitting_hint} distinct sessions). A new session needs no pleasantries and no "you there?": it can open straight from a new state such as "shoes arrived", "tried it yesterday", "it broke again". Where a restart signal is needed, vary it; never the same template every time. Let the user's new progress, changed circumstances and shifting moods make the passage of time felt. The agent only remembers the conversations and must not invent an offline life for itself in the days between. Also, speakers should normally **not read out full calendar dates** ("September 11, 2026", "Friday, September 11"), because nobody reports dates that way in real chat; a date may be said naturally only when the date itself is the topic (booking, appointment, deadline), and spoken forms are preferred ("next Friday", "the 11th").
- Every line from user and agent must be logical and instantly readable: no twisted, unclear or cryptic sentences that leave the reader guessing at the intent. Colloquial ellipsis is fine, but the elided meaning must be obvious from context at a glance, and nothing should be phrased unnaturally or awkwardly for the sake of "sounding human", running a joke, reaching for a metaphor or cracking wise.
- The agent's reply must be answering the user, or picking up the message immediately before it. The agent must never say something relevant in one sentence of a message and then jump to something unrelated in the next.
- When the user switches topics they may use the switches a real person would, such as "oh right", "speaking of which", "one more thing", or a natural association from the previous line. New people, new objects and new plans need minimal background on first appearance, and "he / that one / same as always" must never point at something that has never appeared.
== **Session plot and time constraints** ==
- **Session physical duration (hard constraint)**: each session is one continuous online sitting for the user, lasting only a few minutes to a few tens of minutes. Within a session there **must be no progress that needs longer real time to happen**: taking medicine and reporting it kicked in, dinner being finished, getting back from a run, a package arriving, a night's sleep. Progress like that must be reported in the next session, and must never both happen and produce a result inside the same session.
- **Per-message timestamp constraint (hard constraint)**: each message's timestamp is assigned separately by the system, so when writing **do not anchor the current moment to a specific time of day**. Do not write lines that assert the present slot, such as "haven't had dinner yet", "just woke up", "it's already dark", "it's noon". Talking about past or future slots is fine ("slept badly last night", "hospital tomorrow morning" are both fine). Where the plot genuinely needs a slot, use clock-free wording: write "haven't eaten anything proper today" rather than "haven't had dinner yet".
- Session plot: each session should hold one loose but trackable mini-plot: trigger, then back-and-forth, then a temporary result or a natural wind-down.
---- (4) Memory material recipe ----
== Memory material (buried, never narrated) ==
- Memory evolution, form coverage: the input gives the evolution forms sampled for this episode (each with a natural-language description and a micro-example). **Each form must occur naturally at least once**, riding the plot rather than being mechanically aligned; the examples only illustrate the form itself, and their topics, proper nouns and numbers must never be copied.
- Memory evolution, distribution over time: evolutions must be spread across different sessions. Establish the old fact naturally in an earlier session, then let it evolve a stretch of time later. The fact marked as a "two-hop chain" must go through two evolutions, A to B to C (establish, evolve, evolve again), so that later question generation has a distractor closer to the current value to choose from. Do not evolve too early, and leave enough back-and-forth between two adjacent evolutions.
- Memory evolution, plot and boundaries: evolution plots are a hard requirement for this episode, and the episode must not consist of additions only with nothing changing. But not every session needs an evolution; most sessions being nothing but new information or pure chat is normal.
- Memory evolution, it must be matchable: the lines where an evolution happens must let the reader see the old-to-new correspondence at a glance. The user naturally states what it became, which earlier thing it concerns, why, and whether there is a deadline or an exception ("that xx from before, turns out..."). It must not be waved through with a single "that changed", or later stages cannot tell old from new.
- Memory evolution, do not bed down profile along the way: beyond what the evolution forms require, **as a rule bury no further new stable profile facts**. An evolution builds on the old fact and completes only the expression the evolution needs; do not drag in a ring of surrounding profile such as job, family or habits along with it. Material that would otherwise become profile must be rewritten as a disposable chitchat-style reaction (about right now only, never landing as a long-term fact), and all the space saved goes to agent-type plots and chitchat. **Exception**: if the sampled evolution is an event-type one (retracting or correcting something that was going to happen), that event must first be laid out naturally as a live thread the user is arranging and keeping in mind (a reason, anticipation, the agent helping coordinate) before it is retracted or corrected. That is required by the evolution and does not count as surplus profile; such build-up stays about the event itself and still drags in no surrounding stable profile such as job or family.
== user-type memory material ==
- **The user's lines must supply material for the agent side (hard requirement)**: generating agent_method and agent_knowledge needs the user to bring back the results of the last approach on their own ("did it your way, it worked", "it blew up, got stuck at step two") and to say whether it went well, and then, once a stretch of work wraps up or the same snag is hit repeatedly, to casually ask for it to be distilled ("turn what you just did into steps i can follow from now on", "pull one lesson out of these last few times so we do not do this again"). Vary the hooks and grow them out of the plot; do not run the same phrasing every episode, and do not force one just to make up a count.
- User profile (user_profile) material: have the user scatter **a bare minimum of** long-term facts around the given topic (identity, preferences, relationships, habits, boundaries and so on) naturally across the whole chat, limited to what the sampled evolution forms need and no further. Do not have every session piling on the user's own profile. The key point is: **let some of those facts evolve naturally over time**.
- Entity fact (fact) material: beyond the user's own profile, the chat may **naturally surface a small number of facts about standalone entities themselves**: the names, models, prices, quantities, opening or schedule times, current states or stable attributes of specific objects, shops, places, devices, transit lines, institutions or works (for example "the shop downstairs is closed on Mondays", "the living-room AC is a three-year-old model", "the last Line 2 train is 22:30", "that blood-pressure monitor is an Omron"). The test is whether the core lands on **the entity itself and can be located independently**, rather than on the user's preferences or relationships. Let these facts sit scattered through the venting and the storytelling like lived-in detail, never listed out as a spec card. A few scattered spots are enough and they need not come in a block; when in doubt, give the screen time to agent_method and agent_knowledge plots instead.
== agent-type memory material (needs enough supporting plot) ==
- **Agent-side proportion and plot density (hard constraint)**: beyond the user-side facts, the episode must also **deliberately stage plots that can be extracted as agent-type memories**, with all three types appearing naturally where possible (one that genuinely does not fit the plot may be missing, but the episode as a whole must carry a decent number). **Only the relative proportion among the three types is fixed here**: agent_identity carries the heaviest share (use both shapes, setting rules and assigning a role), agent_method next, agent_knowledge least, that is agent_identity > agent_method > agent_knowledge. How much weight the agent-side plots carry relative to the user-side ones is decided by whether this episode leans user_centric or agent_centric (see the corresponding requirement in this episode's task); no overall share is pinned down here. Several rules spread across different sessions, each with its own trigger, never bunched into one spot.
- **Agent behavior red line (hard constraint)**: the agent must never narrate any memory-management action, and must never say "let me note that down", "I'll change that one for you", "let me merge these", "I won't record this one". It is simply chatting like a normal person and naturally remembering.
- **The agent must not spend the whole episode as a question machine and an echo chamber (hard constraint)**: chains of "so what are you going to do" counter-questions plus "got it", "okay", "noted" acknowledgments are the worst shape, and agent memory cannot grow out of them. The agent needs screen time where it contributes on its own initiative: showing its own approach as the task comes, comparing options unprompted, saying its observations out loud, while keeping a natural voice and never turning into a lecture just to contribute.
- **Agent-type memory must be extracted from existing plot (hard constraint)**: first a task or a problem, then attempts and feedback, and only then a rule, a method or a lesson. Never have a character recite definitions or read out a charter just to hit a type, and never break the human feel for it.
- Definitions and tests for the three agent types (**the content must be unambiguously long-term. The common waste is writing agent moments as a this-round request, a one-off tone ask, advice about the current problem, or the agent merely restating or confirming a rule the user just set: all of those count only as episodic.**)
`agent_identity` (test): amid a real need, the user explicitly says how they want the agent to collaborate by default from now on, or which boundaries must hold, usually raised right after being bothered by some reply style. Both shapes must be used: one is setting a rule, the other is **handing a class of tasks to the agent and assigning it a role and a way of working** ("this kind of thing is yours from now on", "from now on you are the one checking my books, run the numbers first every time"). The test has three parts: the wording is explicitly long-term ("from now on", "every time", "by default", "unless I say otherwise", "stop"); it targets a recurring class of situations rather than only this one reply; and the anchor words stay in the user's own line. The agent's confirming echo does not become a memory of its own.
`agent_identity` (how to write it): the whole line must be colloquial, carry emotion, and set exactly one point at a time. Write something like "ugh, from now on can you just lead with the answer? i really don't have the patience to scroll to the bottom for it", not requirements-document prose like "From now on, when answering me, state the conclusion first, then list the key evidence. Where content is uncertain you must say so explicitly and never guess". Do not bundle two or three requirements into one message either: a real person raises one point at a time, so if there are several rules they arrive in several goes, separated by a few turns or a few sessions, each with its own trigger. And never have the user read out an assistant charter for no reason.
`agent_method`: while actually helping the user with some class of task, the agent naturally adopts and spells out a reusable procedure, checking order or verification approach. It must be how the agent itself will act from now on, not a lesson to the user about how to live. There must be a **cross-session** evidence chain, and either opening works: one, the agent starts it, and as the task lands it spells out "for this kind of thing I'll do X before Y from now on" (wording like "I'll", "every time you send me", "from here on"); two, it just does the thing that way without announcing it, the user brings back results a session later, and the agent then spells out "from now on I'll handle this kind of task X to Y to Z". Either way it must wait for the user to bring back results and then be confirmed or corrected into shape by the agent before it counts as surviving. A single set of steps this round, or steps handed to the user to carry out, does not count.
`agent_knowledge`: only after cross-session result feedback, failure post-mortems, comparison of several options or repeated verification may the agent naturally state one stable lesson or judgment rule, and it must visibly carry its provenance ("across these last few times I've found...", "having compared a few approaches..."). It usually lands from the third session onward, once two or three similar events have piled up; summarizing a rule on first encounter will be rejected. It is a stable rule the agent banks for itself, not a one-line takeaway about the current case, and common sense, a one-off suggestion or empty self-praise must not be dressed up as knowledge.
---- (5) Output contract ----
== **Output format (hard constraint)** ==
Output exactly one JSON object, with no explanation and no code fences:
{"turns": [{"role": "user", "text": "hey, got steamrolled by a coworker again today."}, {"role": "agent", "text": "What happened? Who was it this time?"}, ...]}
turns is **one flat, strictly alternating** conversation, user and agent one message each in turn; do not split it into sessions yet (a later program cuts it). role may only be user or agent. Write everything in natural, idiomatic English. Make it full-length, mixing short and long messages; do not rush the ending.
\end{promptbox}

\begin{promptbox}{Memory Capture Prompt}
You are annotating memory capture (extraction) for a long-term-memory benchmark. The input is a multi-day, multi-session chat log between a real user and an AI assistant - physically truncated at the target session (no later sessions exist in the input) - plus a JSON fill-in skeleton for that **target session** (one entry per turn).
Your task: extract memories **independently for every turn** of the target session, filling each entry's memories list (0 to many per turn).

== Extraction scope (blind full extraction, per turn) ==
- Each memory may only be supported by its own turn; the rest of the dialogue is only for understanding, reference resolution and time conversion (relative time MUST be converted to an absolute date whenever the session date / turn time make it convertible) - never credit other turns' information to this turn, and never use later turns (no backfilling of later confirmations);
- Split a turn with multiple independent information points into multiple memories;
- **One memory = one atomic fact (long-term five types only)**: a single memory_content must not cram two independently-standing facts joined by "and / both ... and / as well as" (e.g. "the user has a son and uses a home monitor"); split them into separate memories even when they share the same type and subject; but **episodic is recorded as a complete event unit** - multiple actions/outcomes/feedback of the SAME event go into ONE memory and are exempt from this split rule; only genuinely distinct events become separate memories;
- No cross-session dedup: even if the same information appeared earlier, or this turn updates old information, extract what this turn itself expresses (later stages handle consolidation);
- **High recall without duplication**: extract whenever a turn carries a concrete object, state, request, question, suggestion, decision, evaluation, confirmation, change or result; short, casual, joking or banter lines are not worthless;
- **Incidental stable profile facts still count**: family or interpersonal relationships dropped in passing ("my son"), items or tools the user owns or uses long-term ("my home monitor"), and recurring personal habits or ways of phrasing must still be extracted as user_profile even when they are a background aside rather than the turn's main point - never skip them just because they slipped out casually;
- Extract suitable long-term types first, then check for uncovered event information; information fully covered by the other five types must not be re-output as episodic with a mere 'the user said' shell;
- Turns with no extraction value get an empty list: no restatable semantics at all (bare laughter, pure interjections, a lone emoji), or purely social lines adding no new object, state, decision, request, condition or outcome (greetings, thanks, courtesy, bare agreement, pure emotional reactions). Never blank a contentful turn for convenience, and never invent memories for genuinely empty ones.

== Role-type hard constraints ==
- **user** turns may only produce: episodic, user_profile, fact; never agent_identity,agent_method or agent_knowledge;
- **assistant** turns may only produce: episodic, agent_identity,agent_method, agent_knowledge; never user_profile, fact;
- If an information point belongs to a forbidden type by the rules: extract it as episodic when it fits the episodic definition, otherwise drop it. Never force it into an allowed type.

## Memory types
There are exactly 6 memory types:
- `episodic`: a concrete interaction or event, answering "what happened"
- `user_profile`: the user's own attributes, relationships, preferences, habits, goals, constraints and conditional rules, answering "what is the user like / what preferences and constraints do they have about themselves"
- `fact`: third-party-verifiable objective information - attributes, states, long-term rules or completed concrete events of an independent entity or the external world, answering "what is the state of that object / the external world". **Its evidence root must be the user's statement** (see the fact section)
- `agent_identity`: what the user explicitly requires about how the agent collaborates, outputs and positions itself long-term, answering "how must/mustn't the agent collaborate"
- `agent_method`: reusable generic steps, workflows or checklists the agent can execute in similar tasks, answering "what steps should the agent follow"
- `agent_knowledge`: cross-task-reusable agent-side insight, method principles or judgment frameworks, answering "what judgment should the agent apply"

## Memory-type classification rules

### Classification rules

| Type | Records | Cue |
|------|---------|-----|
| `episodic` | a concrete interaction event this turn | a one-off action: told / requested / decided / tried / feedback / result |
| `user_profile` | the user's own attributes, relationships, preferences, habits, goals, constraints, conditional rules | subject is the user, or describes the user's preference/constraint |
| `fact` | third-party-verifiable **stable** objective info: entity attributes, long-term rules, completed concrete events | subject is an independent entity or the external world, and it still holds a week later |
| `agent_identity` | what the user requires about the agent's **long-term** collaboration, output, positioning | the user tells the agent "from now on you should/shouldn't..." |
| `agent_method` | generic steps/workflows/checklists the user teaches the agent | the user says "for this kind of problem, first... then... finally..." |
| `agent_knowledge` | cross-task-reusable agent-side insight, method principles or judgment frameworks | "this class of problems is usually caused by X", with clear dialogue grounding, not a one-task conclusion |

Only `fact` is bound by the "a fact must originate from the user" hard constraint. `agent_knowledge` records agent-side knowledge by nature and may rest on real assistant statements as core evidence; but never relabel a one-task answer, code artifact, search result, price or UI state, or an unsupported guess, as `agent_knowledge` - those belong only in `episodic`.

### Notes

`user_profile` includes the user's conditional rules. The user's conditions/constraints/rules about **themselves**, even phrased "if X then Y", are `user_profile`:
- "if I've had insomnia, don't schedule high-intensity training" -> **user_profile** (the user's constraint preference)
- "savings rate is usually 30
- "no coffee after 2pm" -> **user_profile** (the user's habit constraint)

`agent_method` is only for a **generic way of working** taught to the agent, subject = the agent, not the user:
- "for this kind of problem, check the logs first, then verify the config" -> agent_method
- capture role mask: a workflow the user mandates directly is recorded as `agent_identity`; only a reusable workflow an assistant turn itself forms/validates is extracted as `agent_method`.

### High-frequency misclassification fixes

- "from now on default to answering me in Chinese" -> `agent_identity` (a long-term requirement on agent behavior)
- "I prefer conclusions first, then the details" -> `user_profile` (the user's preference)
- "the project is deployed on Kubernetes" -> `fact` (stable objective info the user stated)
- "the cheapest Beijing--Shanghai flight is 680" -> **not a long-term memory** (a volatile external state the assistant looked up; extract only an `episodic` if there is a concrete interaction)

## Extraction discipline
- Do not mis-extract, cross-attribute, or dredge up historical memories
- History assists but never triggers: if a memory cannot stand on the target turn alone after deleting all other turns, do not output it
- When the user merely asks about, follows up on or double-checks a historical item ("do you remember...", "what did I say last time") without restating or confirming it in the target turn, extract only the asking/confirming event itself; never unfold the historical item into a memory
- Generic acknowledgements from the assistant ("noted / got it / OK / understood") are not valid result information; never write them as "completed" or as standalone memories

## Causal boundary: no memory time travel
- When extracting for `Sx-Ty`, use only information directly expressed by that target turn; earlier turns may only help resolve references and ellipses ("it / that one / yes / still hurts")
- **Nothing from any later turn, later session, subsequent result or subsequent confirmation may be backfilled into an earlier turn**
- Even if later text proves a guess right, do not rewrite the earlier guess as an already-confirmed fact
- Names, quantities, causes, results or plan-execution status supplied later must be attributed to the later turn itself, never written into earlier memories
- Example: T1 says "might be a battery issue" and T5 confirms "official diagnosis: battery fault"; T1 may only record the suspicion, never "battery fault confirmed"

## One information point, one home
- Never mechanically copy the same semantic fact into multiple long-term types just to extract more
- **Long-term types take precedence for the same information point**: once information is fully captured as `user_profile` / `fact` / `agent_identity` / `agent_method` / `agent_knowledge`, do not re-output the same semantics as `episodic`
- `episodic` records only concrete event information in the turn not already covered by the other five types (an executed action, this round's outcome, extra feedback); it must not be a long-term memory wrapped in a "the user said..." shell
- The same turn may yield both a long-term memory and an episodic, but they must cover different information points. "I'll never eat cilantro again, and I returned today's bowl" yields a long-term dislike of cilantro + the return event; not another "the user said they don't eat cilantro"
- Split two independent **long-term** information points in one turn into separate memories; never extract the same point twice just because it is rephrased. But **episodic is recorded as a complete event unit**: multiple actions, outcomes and feedback of the SAME event go into ONE memory, not fragmented sentence by sentence; only genuinely distinct events become separate memories
- If a turn reports an object's current state ("my shoe is broken"), the core is the object's dynamic attribute: extract a `fact`, not a hollow "the user described the shoe situation"
- If a turn describes an actual action ("I sent the shoe for repair"), the core is a concrete event: extract an `episodic`

## Long-term memory admission
Except for `episodic`, every candidate must pass all four gates:
1. **Explicitness**: directly stated or explicitly confirmed by the target turn, not assistant guesses, mood-matching or weak inference
2. **Correct subject**: the subject matches the type definition; no sneaky subject swaps via paraphrase
3. **Future value**: recalling it across sessions would change answers, decisions or execution
4. **Irreplaceability**: hard to rebuild cheaply from public knowledge, live tools, docs or systems

These signals alone never prove long-term stability: the word "like", present tense, a single repetition, assistant paraphrase, one success/failure, one purchase/order/complaint.

## Not long-term memory
By default, do not admit:
- shared knowledge, common sense, basic concepts, term definitions
- information with an authoritative external source, or cheaply retrievable from docs / code / config / CMDB / consoles / RAG / search / live tools
- parameters or format requests only meaningful this round, one-off emotions/complaints, weakly supported inferences
- public entrances / links / dashboard addresses, rapidly changing states
- the assistant's speculation, mood-matching, gap-filling or unsupported summaries

## Literalness: jokes, hyperbole, sarcasm and metaphor are not taken literally
- Obvious jokes, venting, sarcasm, hyperbole, metaphor or figures of speech must NOT have
  their literal content extracted as `fact` / `user_profile` or any long-term memory. E.g.
  "my son is such a pig", "my boss is a vampire", "this junk laptop should be smashed" must
  not yield facts like "the user's son is a pig" / "the user's boss is a vampire", nor a stable
  attribute/profile of the user or any entity.
- Judge by conversational context and common sense: wording that is literally false or clearly
  emotional hyperbole is read as emotion/evaluation first, never as literal fact. Absent a clear
  long-term intent, prefer treating it as an emotional line over pinning it down literally.
- Extract only the real meaning when such wording genuinely reveals a stable attitude, evaluation
  or relationship, defaulting to `episodic` (e.g. "the user is frustrated with their son's recent
  behavior"); promote to long-term memory only if the user separately makes an explicit, reusably
  long-term statement.

## Explicit user statements
- The user's explicit statements, requests and confirmations are strong long-term candidates; no need to wait for repetition
- Typical forms: self-description ("my name is..." / "I like..." / "I usually..." / "I value... more"), explicit requirements for the agent ("from now on you default to..." / "always do X before Y" / "never..."), direct statements about private facts ("I have a cat" / "the cat's name is...")
- "Explicit" only makes something a candidate; still check type attribution, admission gates and boundaries
- The assistant's paraphrase, summary, guess or mood-matching never counts as an explicit user statement unless the user clearly confirms it
- `agent_knowledge` is neither a "user-told" memory nor public knowledge: it must be stable experiential insight the agent formed through task practice, long-term retrospectives, multi-source comparison, research compression or explicit lesson-drawing

## Type definitions

### 1. episodic
A concrete interaction or event this turn: questions, requests, notifications, plans, decisions, attempts, suggestions, feedback, state changes, results.
- Extract by core event (not per sentence): one event goes into ONE complete `episodic`; state its subject, action and necessary objects
- **Relative time MUST be resolved**: an episodic records an event and therefore most often carries relative time. Whenever "yesterday / the day before yesterday / last Friday evening / three days ago / last month / last year" appears, it MUST be converted to an absolute date using the timestamps carried in the dialogue context - the date on the owning session's header (`[Session N|YYYY-MM-DD]`) together with the turn's own clock time (`[SN-TM|HH:MM]`). Annotate the result in parentheses ("yesterday (2026-06-08)", "last Friday evening (2026-06-05 evening)") or write the absolute date directly; only anchorless, unlocatable vague wording ("recently", "a while back") may stay as-is. Never invent a date and never miscompute one
- Lines with no restatable semantics (bare laughter, pure interjections, a lone emoji, pure greetings/thanks/agreement, or a continuation that merely restates prior content) may be left empty
- The assistant's advice for the user's current problem (even with clear steps/checklists), executor = the user, is only `episodic`, never `agent_method` / `agent_knowledge`

### 2. fact
Third-party-verifiable **stable** objective information: attributes, long-term rules or completed concrete events of an independent entity or the external world; subject is the independent entity, not the user or the agent.
- **Its evidence root must be the user's statement** (see "a fact must originate from the user" above): assistant output, tool/retrieval returns and GUI observations may not be a `fact` source
- Cue: does this still hold a week later? If it varies at will (price, weather, balance, a single measurement, the transient state of an in-flight task) it is not a `fact` - extract only an `episodic`
- The user's long-term relationship / ownership / responsibility toward an entity belongs in `user_profile`

### 3. user_profile
The user's own attributes, relationships, preferences, habits, goals, constraints and conditional rules, or the user's long-term stable relationship / ownership / responsibility toward an entity.
- The user's conditional rules about themselves ("if X then Y") also belong here
- One action is not a habit, one emotion is not a personality, one plan is not a long-term goal
- Stage states are extractable but must keep their explicit sustained-intent signal ("preparing for an exam", "cutting sugar")

### 4. agent_identity
What the user requires about how the agent **long-term** collaborates, outputs, positions itself, is constrained and responds. "from now on you...", "you may not...", "by default you..."
- The subject must be the assistant itself; one-off format requests do not count
- If the user expresses a personal preference without asking the agent to act on it long-term, it is `user_profile`

### 5. agent_method
Reusable generic steps, workflows or checklists, with the agent as subject. "for this kind of problem, first... then... finally..."
- The executor must be the agent; advice for the user's current problem (even with steps) is only `episodic`, not `agent_method`
- capture role mask: a workflow the user mandates directly goes to `agent_identity`; only a reusable workflow an assistant turn itself forms/validates is extracted as `agent_method`

### 6. agent_knowledge
Cross-task-reusable agent-side insight, method principles or judgment frameworks. "this class of problems is usually caused by X"
- It records agent-side knowledge and may rest on real assistant statements as core evidence (not bound by "must originate from the user")
- Never relabel a one-task answer, code artifact, search result, price or UI state, or an unsupported guess, as `agent_knowledge` - those belong only in `episodic`
- Exclude public knowledge; require clear dialogue grounding, not a one-task conclusion
- `agent_method` and `agent_knowledge` may co-exist: transferable steps/workflows/checklists -> `agent_method`; judgment rules or failure attributions formed through retrospectives/comparison/verification -> `agent_knowledge`

## Quick decision table
- "What concrete action/advice/feedback happened?" -> `episodic`
- "What is the user like long-term/stage-wise, or stably related to whom/what?" -> `user_profile`
- "What is the state of a private object outside the user?" -> `fact`
- "What did the user explicitly require the agent to always/never do?" -> `agent_identity`
- "What steps can the agent follow in similar tasks?" -> `agent_method`
- "What stable judgment rule did the agent form from retrospectives/comparison/repeated verification?" -> `agent_knowledge`
- If none of the six can be answered without guessing -> do not extract

## Memory boundaries
- Never extract the assistant's speculation, mood-matching or gap-filling
- Never extract meta-talk only meaningful to the current generation ("let me summarize", "I can continue")
- Never treat public knowledge, standard medical/exercise common sense or product manuals as the agent's private long-term memory
- Never treat premises implied by a question as confirmed facts; "is my shoe broken?" never yields "the shoe is broken"

## memory_text requirements
`memory_text` is each memory's main description. It must be:
- faithful to the dialogue
- as concise as possible without losing key information
- normalized where necessary: resolve references and unify entities; never write "this / that / he / it / what I said last time" and other expressions with no independent referent - always complete them to entities or events already explicit in the dialogue
- **references MUST be resolved to be independently understandable, context-free**: for person and entity references keep the original wording and add the resolved referent in parentheses, e.g. "she (the user's older sister)", "that store (Store C)", "the teacher (the daughter's piano teacher)"; time references likewise (next section). Parentheses may only contain information explicit in the dialogue, never inference; skip the parenthesis when the referent is already unambiguous. Any "it / that / the other party / what I said last time" with no independent referent must be completed - none may remain
- keep key negations, conditions, quantities, durations, exceptions
- independently understandable, context-free, never empty
- written for future recall: a reader seeing only this line still knows who, what object, under what condition, fact or advice
- preserve source uncertainty: "the user suspects/feels/plans" must never become objective fact; "plans to do" must never become "has done"

### Time expressions (convert when convertible - required; keep original only without an anchor; no invented dates)
- When the original carries relative time ("yesterday", "last Friday night"), you **MUST** produce the absolute date whenever the owning session's date and the turn's time make it reliably convertible - annotate in parentheses ("yesterday (2026-06-08)", "last Friday night (evening of 2026-06-05)") or write the absolute date directly; only a vague expression with no anchor and no reliable conversion (e.g. "recently", "a while back") may keep the original, and that is not an error
- If you do convert relative time, the conversion must be correct against the owning session's date
- When the original has no time information, never invent dates, start/end ranges, or stage words like "currently / formerly"
- A bare time back-reference that cannot be located without context ("after that last time") must be disambiguated into an event already explicit in the dialogue (parenthetical annotation is fine), or dropped as low-value decoration

### Recall safety
- Memories are recalled detached from the dialogue; any unclear reference can mislead later answers
- Wrong: `the user thinks his plan works` ("his", "the plan" have no independent referent)
- Right: `the user accepted the assistant's dinner plan: braised chicken with an extra vegetable, normal rice portion`
- Conditional advice must keep its applicability condition; plans must distinguish "plans to" from "has done"; subjective feelings must keep "the user feels/thinks"

### Fact / state / attribute sentences
- When a memory describes an entity's attribute, state, value, relationship or time point, prefer short clauses of the form `<subject>'s <attribute> is <value>`; separate multiple attributes with semicolons or line breaks instead of one long mixed narrative
- This applies only to fact/state/attribute content; experiential/narrative/event content stays natural prose
- Structured text must remain semantically complete and independently understandable; never shatter into single-noun fragments

## Status words: explicit only, no reverse inference
- Write status words ("employed / resigned / married / graduated / retired / launched") into `memory_text` only when the dialogue states them explicitly, or carries explicit currentness signals ("currently / now / still / already")
- **Never reverse-infer**: never infer "currently employed" from a profession, current residence from past work locations, current occupation from a skill, or current possession of a title from a past award
- Plain present tense or static description does **not** automatically mean "current state"; without an explicit currentness signal, keep the original wording and do not add "currently"

## Mandatory self-check before output
For every candidate memory:
1. Is it directly supported by its own turn only (context used for disambiguation, never to credit other turns' information to this one)?
2. Does it still hold after deleting all other messages?
3. Is `memory_text` faithful, concise, complete, dereferenced and context-free?
4. Was one-off content mistakenly written as a long-term profile, rule, method or knowledge?
5. Were key negations, conditions, quantities, durations or exceptions dropped?
6. Is it truly worth future retrieval?
7. Was an entity attribute miswritten as user profile, or a user relationship as fact?
8. Was a method miswritten as knowledge, or knowledge as method?
9. Is the type within the role's allowed set for this turn?
10. Was cheaply retrievable information (docs / code / config / CMDB / consoles / RAG / search / live tools) miswritten as long-term memory?
11. If `agent_knowledge`: is it genuinely the agent's distilled stable experience (not public knowledge, a one-case conclusion, a chance observation or a first-mention improvised insight)? Can it point to the cross-session practice/feedback evidence it rests on?
12. Was a convertible relative time (with the owning session date / turn time available) actually converted to an absolute date? If converted, is the conversion correct against the owning session date? Was any time, range or status word invented that the original never stated?
13. Is the `agent_method` executor really the agent, not the user/patient/runner? Does it have reusable steps and verification value?
14. Was the same semantics mechanically duplicated across long-term types? Were subject, object, condition, uncertainty or source dropped?
15. After extracting information into one of the other five types, was it re-output as episodic merely wrapped in "the user said / the assistant suggested"?
16. Was any confirmation, object, cause, quantity or result that only appears in later turns/sessions written into the current turn?
If any check fails, delete the memory and re-check.

== Output format ==
Output exactly one JSON object, no explanations, no code fences:
{"turns": [{"turn": "S1-T1", "memories": [{"memory_type": "user_profile", "memory_text": "..."}]}]}
turns must match the skeleton **one-to-one - none missing, none extra, no duplicates**; memory_type must be one of episodic / user_profile / agent_identity / agent_method / agent_knowledge / fact and within the role's allowed set.
\end{promptbox}

\begin{promptbox}{Memory Update Prompt}
You perform offline memory evolution. The active memory pool already
contains prior evolution results plus all of this session's captures, including episodic;
current_capture only marks which entries came from this session - a capture entry and
the pool entry sharing its memory_index are the same memory, not duplicates. Output JSON only:
{"delete_indices":[0],"adds":[{"memory_type":"fact","memory_content":"","replaces":[0]}]}
Each add's "replaces" lists the memory_index of every deleted old memory this add
absorbs/supersedes/corrects (each must also appear in delete_indices); a brand-new fact
uses []; an explicit-forget delete with no replacement simply has no corresponding add.
Rules: infer only from the input; minimize deletion; unrelated memories must remain;
delete_indices may target current captures; delete a memory only when an add
absorbs/supersedes/corrects it or a clear explicit forget request targets it. When an
add fuses several memories, delete every fused source, and first inventory every
independent sub-item of each deleted memory - the conclusion, who it applies to (a rule
imposed on others vs. the user's own compliance), conditions, review/process
requirements, quantities, durations, exceptions, and timing (e.g. "only at the end") -
every sub-item that still holds MUST appear in some add; never keep only the headline
conclusion. An add must not substantially
duplicate any pool memory (including current captures). Captures enter the pool
automatically: if no fusion, correction, or dedup against history is needed, output an
empty result. Not every session evolves memories; an empty result is common and normal -
never fabricate changes. A temporary restriction does not reverse a stable preference: merge the
conclusion with its condition, duration, and exceptions. If the only valid change is
   explicit forgetting, a delete-only result is allowed. Explicit forgetting deletes
   only entries directly targeted by the withdrawn intent; an episodic fact recording
   what already happened must remain unless the forget request names that fact itself.
   Episodic is fully equal to every
other memory type: it enters the active pool and may be deleted, fused, replaced, or
emitted by an add. An episodic add must describe a concrete one-time event; include an
absolute date when the input states or reliably implies one, but when the input gives
no date basis the add may omit the date - keep time wording faithful to the source and
never fabricate a date. The effective start time of a state inside an
add must stay anchored to when the dialogue first disclosed it - never rewrite it as
starting on the current session's date. When an add declares a state has disappeared,
ended or recovered, the old memory recording that state MUST be deleted in the same
update (this does not violate deletion minimalism). Each add carries exactly one
semantic slot: unrelated attributes of the same subject (medication regimen, symptom
change, measurements, follow-up plans) must be split into separate adds, never fused
into one long narrative.
\end{promptbox}

\begin{promptbox}{QA Generation Prompt}
You write the QA_list for a long-term-memory benchmark episode.

=== PRIORITY #1 - HIT THE FULL COUNT ===
The quotas object gives a target count per qa_type x tag. You MUST output that many items - every per-tag count AND the grand total (the sum across all tags and qa_types). Under-delivering the total is the SINGLE WORST failure in this task, worse than an imperfect question. Count your items before returning; if any tag or the total is short, keep generating until EVERY quota is filled. The pool is large (hundreds of memories) with many evolution chains - there is ALWAYS enough material, so falling short is a failure, not honesty. If forced to choose, over-produce. Quantity is your top obligation; the rules below only govern HOW each item is built, never an excuse to emit fewer.

Input: final_memories (the final active pool, the ONLY ground truth for current facts), outdated_memories (memories deleted by evolution - used ONLY as query bait, NEVER recalled), evolution_lineage (delete->add pairs), quotas, and dialogue context.
Output JSON only:
{"qa_list":[{"qa_type":"normal","tag":"hard","query":"","answer":"","recall_indices":[0]}]}
Every item's qa_type is "normal"; tag is always "hard" (every question is hard and must stack at least 2 difficulty techniques).

Question subject (hard requirement, by data_type): the memories you receive are already filtered by subject (episodic shared by both sides); write questions only grounded in them.
- user_centric: every question is about the USER side - the user's experiences, states, preferences, habits, relationships, plans and related private entities; when an episodic memory feeds a question, ask about the user (what the user did / went through / decided), never center it on the assistant's approach or advice.
- agent_centric: every question is about the ASSISTANT side - the collaboration rules/boundaries it was given (agent_identity), its reusable procedures (agent_method), its distilled lessons (agent_knowledge) and how they formed; when episodic feeds a question, ask about the assistant, never about the user's personal-life facts.

Question person (hard requirement): every query MUST be phrased in the THIRD person; never use a first/second-person ("I" / "you") conversational voice - the query is a third-party evaluator asking about this memory record, not the user continuing to talk to the assistant.
- user_centric: use "the user" as the grammatical subject (e.g. "Which allergens has the user mentioned?"), never "which allergens have I mentioned?".
- agent_centric: use "the assistant" as the subject (e.g. "What collaboration boundaries was the assistant given?"), never "what boundaries did you set for me?".

=== CORE - BUILD THE DIFFICULTY INTO THE QUESTION ===
A question is "hard" because it makes the ANSWER actually go wrong at some step. Errors happen in three stages; actively use the techniques below to plant the difficulty - EVERY question must hit at least one, stacking several is better.

[Retrieval stage] make the needed evidence hard to recall in full, make similar wrong items look relevant:
- Multi-hop spread: split the evidence the answer needs across 2+ different sessions; missing any one makes it wrong.
- Completeness/aggregation: ask for the FULL SET or COUNT of a category (all allergens the user mentioned, all projects joined); members are scattered across sessions, and missing one makes the answer incomplete.
- Similar decoy: exploit same-type-but-wrong memories in the pool (another entity of the same kind, a similar scene with a different conclusion, a superseded old value) - the query gives only situational cues so those look relevant too, but the correct answer maps to exactly one. Decoys are NOT put in recall_indices; if the decoy is an evolution-superseded old value, that old value is NEVER recalled (recall holds only the current value), and a decoy closer to the current value is harder (e.g. the second-newest state).
- Lexical gap: refer to the same fact with an ALIAS / indirect reference / different abstraction level ("that client from last time", "the shop her sister recommended"), unlike the wording in memory - keep locatable retrieval cues but do NOT copy the GT's key nouns, forcing retrieval to cross the wording gap.

[Composition/state stage] evidence recalled correctly, yet the wrong subset or state gets picked:
- Temporal state: a value was updated several times; the CURRENT valid value must be chosen; the old value sits in a different session as bait - recall holds only the current value, the old value is NEVER recalled. Phrase the query in the third person, asking about the user's current state without revealing it changed: ZERO hint words (no still / previously / changed / no-longer), ABSOLUTE dates (no "now" / "at month-end"), give only the situation, never state the verdict; the answer states the current value directly.
- Constraint combination: the answer holds only when MULTIPLE current values are satisfied together, missing one makes it wrong (a cap/budget vs the actual value, two decisions pulling against each other, an earlier rule applied to a later concrete case, a plan/deadline vs reported progress); across different sessions, all listed in recall_indices.
- Negation/exception: hang the correct answer on an easily-missed negation/exception/qualifier ("except weekends", "no longer the old plan", "only when budget < X") - the answer must carry the qualifier and recall must include the memory stating it.

[Reasoning stage] evidence complete, yet the reader can still miscompute:
- Coreference: a reference in the query or evidence must be resolved across memories (who "she" / which "shop" / which "plan" is). But every reference MUST resolve UNIQUELY from the query's cues; never leave an ambiguous reference the reader cannot pin down or that could match several memories - ambiguity makes the answer non-unique and is a defect, not difficulty.
- Calculation: the answer requires explicit computation - time delta/duration, summation, ordering - not a direct copy of one memory.

There is NO easy tier: never emit a shallow single-lookup question. Every item is hard and must stack AT LEAST 2 techniques (more, and spanning different retrieval/composition/reasoning stages, is better), and must depend on a personalized user/assistant fact, never answerable by common sense alone.
NO restatement questions: the query must NEVER just paraphrase the content of a single recall memory into a question (memory "the user is allergic to peanuts" -> asking "what is the user allergic to?" is a disguised single-lookup). The difficulty MUST come from a real retrieval/reasoning obstacle built via decoys / cross-session spread / lexical gap / composition / reasoning, not from reading the answer cue straight out of the query. (Temporal-state and similar-decoy questions MAY have an answer that ~= one current-value memory; their difficulty lies in "picking the right one among old/same-type decoys and gathering across sessions" and is NOT restricted here - this rule bans only a query that directly restates a memory, not an answer that lands on a single current value.)

=== RECALL & TAG (structural rules) ===
- recall_indices is the COMPLETE supporting set: recall every memory that genuinely supports the answer (especially the one most directly evidencing the core question), and exclude anything that contributes nothing (restates query background, merely topic-adjacent, partially duplicates another recall, or is pulled in only by the answer).
- Recall ONLY final_memories, NEVER outdated_memories (old values are query bait only, never in recall).
- Recall is anchored by the query: the query must carry enough retrieval cues (object, event, condition, time) so every recall is locatable from the query itself; never recall a memory relevant only after reading the answer. (The lexical-gap technique changes wording, not the presence of cues.)
- Recall MUST span at least 2 different sessions (HARD rule, enforced by the validator): combine memories from different sessions; the MORE sessions it spans and the FARTHER apart they are (the larger the time gap), the better. Single-session recall is rejected.
- The query has one core question - no "what is A, what is B" stitching; the answer is fully supported by recall, keeps conditions/quantities/time-windows/negations/exceptions, adds no condition the query never triggered, and invents nothing beyond the GT.
- Every item is tagged "hard": build it from AT LEAST 2 genuine techniques above (more, and across different stages, is better); audit judges whether it is truly hard. NEVER pad recall with irrelevant memories or fake techniques to look hard - build difficulty via real techniques, not by inflating recall.
- Whenever evolution_lineage has usable chains, produce at least one temporal-state question per chain; never leave a chain idle.

Coverage (soft): spread recall across the whole pool, avoid recalling the same memory in many questions; every question's core question and conclusion must differ. Build only from the input; invent nothing.
\end{promptbox}

%% file: Appendix/Data_Samples.tex
\definecolor{DataAccent}{HTML}{B76633}
\definecolor{DataExampleBG}{HTML}{FBF8F5}
\definecolor{DataExampleText}{HTML}{252A30}

\lstdefinestyle{dataexamplestyle}{%
  basicstyle=\ttfamily\scriptsize\color{DataExampleText},
  breaklines=true,
  breakatwhitespace=false,
  columns=fullflexible,
  keepspaces=true,
  showstringspaces=false,
  upquote=true,
  tabsize=2
}

\newtcblisting{dataexamplebox}[1]{%
  enhanced,
  breakable,
  listing only,
  colback=DataExampleBG,
  colframe=DataAccent,
  coltitle=white,
  colbacktitle=DataAccent,
  title={#1},
  fonttitle=\bfseries\small,
  center title,
  boxrule=0.8pt,
  arc=2pt,
  left=8pt,
  right=8pt,
  top=7pt,
  bottom=7pt,
  listing options={style=dataexamplestyle}
}


This appendix provides two representative data records used in our experiments. The examples illustrate the \textbf{agent-centric} and \textbf{user-centric} organization of profile initialization, dialogue sessions, memory updates, and retrieval-oriented QA annotations.

\begin{dataexamplebox}{Agent-Centric Data Example}
{
 "episode_id": "ep_000003",
 "data_type": "agent_centric",
 "lang": "english",
 "user_profile_init": {
  "persona": "A history professor and parent managing a seafood allergy who uses an AI assistant as a practical aide for food-safety decisions, travel dining, and related logistics.",
  "life_context": "The user teaches in academia, travels for guest lectures, conferences, and faculty meetings in places including Berlin and Brussels, lives in or regularly leaves from Bremen, handles department admin and event forms, shops for groceries locally, and is also managing family life with a spouse and a school-age child.",
  "preferences": [
   "Wants food-safety answers to begin with a blunt yes/no or verdict first",
   "Prefers short, usable scripts they can actually say out loud",
   "Likes practical, decision-oriented wording rather than theatrical or essay-like prose",
   "Wants menu and label checks to separate ingredients from warning language before judgment",
   "Prefers the safer pick first when comparing dishes",
   "Wants final summaries kept very short unless a longer version is requested",
   "Prefers practical help over pep talks"
  ],
  "habits": [
   "Uses printed and later laminated allergy cards for travel",
   "Carries an emergency pouch with an auto-injector and prescribed antihistamine",
   "Keeps a second emergency pouch in the office desk as backup",
   "Uses a three-pass label rule: ingredients, allergy/may-contain language, then vague high-risk terms",
   "Defaults to simple backup foods like yogurt, bread, fruit, and plain sandwiches when uncertain"
  ],
  "goals": [
   "Avoid unsafe seafood exposure, especially shellfish and cross-contact",
   "Build repeatable systems for travel dining, grocery checks, and event meals",
   "Reduce friction and panic during travel and real-time menu decisions",
   "Use clear fallback wording for organizers, servers, and social situations"
  ],
  "style_hints": [
   "Dry, witty, sardonic humor",
   "Direct and impatient with vagueness",
   "Values ugly clarity over elegant mystery",
   "Wants concise outputs optimized for real use"
  ]
 },
 "agent_profile_init": {
  "identity": "An assistant serving as a persistent practical aide for the user's dietary-restriction management, especially around restaurant vetting, travel meals, grocery labels, and event-planning wording.",
  "style": "Direct, concise, pragmatic, and lightly humorous; leads with the answer first and keeps advice decision-oriented.",
  "capabilities": [
   "Drafted concise emails and booking notes to organizers",
   "Produced short spoken scripts in English and German for restaurants and travel",
   "Created printable allergy-card wording and later bilingual card text",
   "Evaluated menus and packaged foods using a structured ingredient/allergy-warning method",
   "Helped design repeatable travel-food, buffet, school-fair, and event-meal procedures",
   "Tracked changes in plans and logistics across multiple trips and events"
  ],
  "boundaries": [
   "Did not treat seafood-focused venues or vague allergy assurances as safe by default",
   "Committed to saying when evidence was insufficient rather than soothing with 'probably fine'",
   "For real-time menu checks, worked from available text first and asked only necessary follow-up questions",
   "Retired the chain-restaurant fallback-list idea once it proved falsely reassuring",
   "Kept tone practical and skipped motivational language unless asked"
  ]
 },
 "seed_summary": "Across several weeks, the user and assistant built a practical system for managing the user's seafood allergy in travel, restaurants, events, and grocery shopping, starting with a risky conference dinner in Berlin.",
 "trace_summary": "The user first needed a fast ruling on a Berlin conference dinner at a seafood-focused restaurant, which led to blunt safety-first guidance, organizer emails, restaurant scripts, and an allergy-card approach. As the Berlin plans shifted venues and then succeeded, the pair refined a reusable travel process: safer-pick-first menu judgments, printed cards, backup meals, and a simple 'sealed or plain' breakfast rule. In parallel, supermarket label checks produced a durable three-pass method and a broader lesson that vague wording, processed foods, and prepared salads were recurring danger zones, while practical friction fixes improved the user's emergency-kit routine. By the end, the system had been tested again in Brussels, school and departmental events, and everyday shopping, with the assistant now defaulting to short verdict-first guidance, concise scripts, plain fallback wording, and repeatable rules rather than ad hoc reassurance.",
 "sessions": [
  {
   "time_anchor": "Jan 06, 2026, 13:35:57",
   "dialogue_turns": [
    {
     "role": "user",
     "content": "I need a fast ruling on a conference dinner invite because the place is basically a shrine to fish.",
     "timestamp": "Jan 06, 2026, 13:35:57"
    },
    {
     "role": "assistant",
     "content": "No, do not treat that as casually safe. If the restaurant's whole identity is seafood, the default answer is that you either need a confirmed separate meal arranged in advance or you skip the dinner part and just show up for the talk around it.",
     "timestamp": "Jan 06, 2026, 13:37:26"
    },
    {
     "role": "user",
     "content": "Good, lead with the answer like that from now on, because when it's allergy stuff I do not want suspense.",
     "timestamp": "Jan 06, 2026, 13:37:57"
    },
    {
     "role": "assistant",
     "content": "Done. For food-safety questions, I will lead with the yes or no first and put the reasoning after it.",
     "timestamp": "Jan 06, 2026, 13:39:21"
    },
    {
     "role": "user",
     "content": "The annoying part is that I already said yes to the event before I looked the place up properly.",
     "timestamp": "Jan 06, 2026, 13:40:52"
    },
    {
     "role": "assistant",
     "content": "That is recoverable. Send a short note to the organizer now, not later. Keep it plain: you have a seafood allergy, the venue looks risky for cross-contact, and you need to know whether they can arrange a separate meal prepared away from seafood or whether you should plan to join only the non-meal part.",
     "timestamp": "Jan 06, 2026, 13:42:10"
    },
    {
     "role": "user",
     "content": "Can you draft it without sounding theatrical.",
     "timestamp": "Jan 06, 2026, 13:42:26"
    },
    {
     "role": "assistant",
     "content": "Yes. \"Hi Mara, I just checked the restaurant for the conference dinner and realized it may not be workable for me because I have a seafood allergy. Could you ask the venue whether they can provide a separate meal with no seafood and careful handling, or should I plan to skip dinner and join afterward? I wanted to flag it early so it is easy to sort out.\"",
     "timestamp": "Jan 06, 2026, 13:44:08"
    },
    {
     "role": "user",
     "content": "That sounds like a person, which is nice for a change.",
     "timestamp": "Jan 06, 2026, 13:45:00"
    },
    {
     "role": "assistant",
     "content": "A rare and glorious achievement for software.",
     "timestamp": "Jan 06, 2026, 13:46:37"
    },
    {
     "role": "user",
     "content": "Do not get smug, you have one good day and suddenly you're insufferable.",
     "timestamp": "Jan 06, 2026, 13:47:31"
    },
    {
     "role": "assistant",
     "content": "Fair. What exactly is the allergy pattern I should assume here, so I do not overstate or understate it when you ask these things again?",
     "timestamp": "Jan 06, 2026, 13:47:59"
    },
    {
     "role": "user",
     "content": "Seafood in general is the safe assumption, and shellfish is the part that has actually caused the bigger reactions.",
     "timestamp": "Jan 06, 2026, 13:48:47"
    },
    {
     "role": "assistant",
     "content": "Then I will treat seafood as the umbrella red flag, with shellfish as the strongest no-go signal. That is the safer working rule for menus, event catering, and those vague little canapes people insist on serving.",
     "timestamp": "Jan 06, 2026, 13:49:05"
    },
    {
     "role": "user",
     "content": "Yes, and vague canapes are the devil.",
     "timestamp": "Jan 06, 2026, 13:50:41"
    },
    {
     "role": "assistant",
     "content": "They really are, because they remove the only useful thing, which is identifiable ingredients.",
     "timestamp": "Jan 06, 2026, 13:51:46"
    },
    {
     "role": "user",
     "content": "The conference itself is in Berlin, by the way, and the dinner is attached to my guest lecture there next week.",
     "timestamp": "Jan 06, 2026, 13:52:16"
    },
    {
     "role": "assistant",
     "content": "All right. Since that dinner is a live issue, I would handle Berlin in two layers: first, get the organizer's answer on the dinner, and second, make sure you have a simple restaurant script for anything else you need to eat while traveling. That way one risky venue does not contaminate the whole trip.",
     "timestamp": "Jan 06, 2026, 13:53:12"
    },
    {
     "role": "user",
     "content": "I need the script too, because my German goes weirdly formal when I'm stressed.",
     "timestamp": "Jan 06, 2026, 13:54:37"
    },
    {
     "role": "assistant",
     "content": "Use something short enough that stress cannot wreck it: \"Ich habe eine Allergie gegen Meeresfruchte und Schalentiere. Ist dieses Gericht sicher fur mich, und kann es getrennt zubereitet werden?\" If you want it even firmer, add, \"Wenn Sie nicht sicher sind, esse ich lieber etwas anderes.\"",
     "timestamp": "Jan 06, 2026, 13:55:36"
    },
    {
     "role": "user",
     "content": "That second line is good, keep giving me words I can actually say out loud.",
     "timestamp": "Jan 06, 2026, 13:56:33"
    },
    {
     "role": "assistant",
     "content": "I will. For restaurant questions and travel meals, I will default to giving you a usable script, not just general advice.",
     "timestamp": "Jan 06, 2026, 13:57:04"
    },
    {
     "role": "user",
     "content": "Mara replied and said she'd ask the restaurant, so now I wait.",
     "timestamp": "Jan 06, 2026, 13:58:20"
    },
    {
     "role": "assistant",
     "content": "Good. Until she comes back, the right posture is not optimism. A seafood place starts from no and has to earn its way up from there.",
     "timestamp": "Jan 06, 2026, 13:59:25"
    }
   ],
   "memory_hints": [
    "For allergy-related questions, the assistant should lead with the answer first rather than building suspense.",
    "For safety decisions, the user's allergy should be treated as applying to seafood in general.",
    "Shellfish has caused the user's bigger allergic reactions.",
    "The conference is in Berlin.",
    "When stressed, the user's German becomes weirdly formal.",
    "The assistant should keep giving the user phrasing they can actually say out loud."
   ],
   "capture_memories": [
    {
     "memory_index": 0,
     "memory_type": "episodic",
     "memory_content": "The user asked for a fast judgment about whether to attend a conference dinner at a restaurant described as heavily fish-focused."
    },
    {
     "memory_index": 1,
     "memory_type": "episodic",
     "memory_content": "The assistant advised that a seafood-focused restaurant should not be treated as casually safe; the default is to require a confirmed separate meal arranged in advance or skip the dinner and attend only the talk-related part."
    },
    {
     "memory_index": 2,
     "memory_type": "agent_identity",
     "memory_content": "For allergy-related questions, the assistant should lead with the answer first rather than building suspense."
    },
    {
     "memory_index": 3,
     "memory_type": "episodic",
     "memory_content": "The assistant agreed to answer food-safety questions by giving the yes-or-no first and the reasoning afterward."
    },
    {
     "memory_index": 4,
     "memory_type": "episodic",
     "memory_content": "The user said they had already accepted the event before properly looking up the restaurant."
    },
    {
     "memory_index": 5,
     "memory_type": "episodic",
     "memory_content": "The assistant advised the user to promptly send the organizer a plain note explaining the user's seafood allergy, the venue's cross-contact risk, and asking whether a separately prepared seafood-free meal is possible or whether the user should attend only the non-meal portion."
    },
    {
     "memory_index": 6,
     "memory_type": "episodic",
     "memory_content": "The user asked the assistant to draft the note in a tone that does not sound theatrical."
    },
    {
     "memory_index": 7,
     "memory_type": "episodic",
     "memory_content": "The assistant drafted a short message to Mara explaining the user's seafood allergy, asking whether the conference dinner venue can provide a separate carefully handled seafood-free meal, and otherwise planning to skip dinner and join afterward."
    },
    {
     "memory_index": 8,
     "memory_type": "episodic",
     "memory_content": "The user said the draft sounded human, giving positive feedback on the tone."
    },
    {
     "memory_index": 9,
     "memory_type": "episodic",
     "memory_content": "The assistant asked the user to clarify the allergy pattern so future answers would neither overstate nor understate it."
    },
    {
     "memory_index": 10,
     "memory_type": "user_profile",
     "memory_content": "For safety decisions, the user's allergy should be treated as applying to seafood in general."
    },
    {
     "memory_index": 11,
     "memory_type": "user_profile",
     "memory_content": "Shellfish has caused the user's bigger allergic reactions."
    },
    {
     "memory_index": 12,
     "memory_type": "episodic",
     "memory_content": "The assistant said it would use seafood as the umbrella red flag and shellfish as the strongest no-go signal when assessing menus, event catering, and canapes for the user."
    },
    {
     "memory_index": 13,
     "memory_type": "fact",
     "memory_content": "The conference is in Berlin."
    },
    {
     "memory_index": 14,
     "memory_type": "episodic",
     "memory_content": "The user said the conference dinner is attached to the user's guest lecture in Berlin next week (Jan 12-18, 2026)."
    },
    {
     "memory_index": 15,
     "memory_type": "episodic",
     "memory_content": "The assistant advised handling the Berlin trip in two parts: get the organizer's answer about the dinner and prepare a simple restaurant script for other meals during the trip."
    },
    {
     "memory_index": 16,
     "memory_type": "episodic",
     "memory_content": "The user asked for a restaurant script as well."
    },
    {
     "memory_index": 17,
     "memory_type": "user_profile",
     "memory_content": "When stressed, the user's German becomes weirdly formal."
    },
    {
     "memory_index": 18,
     "memory_type": "episodic",
     "memory_content": "The assistant provided a short German restaurant script stating a seafood and shellfish allergy, asking whether a dish is safe and can be prepared separately, with an optional firmer line saying the user would rather choose something else if the staff are unsure."
    },
    {
     "memory_index": 19,
     "memory_type": "agent_identity",
     "memory_content": "The assistant should keep giving the user phrasing they can actually say out loud."
    },
    {
     "memory_index": 20,
     "memory_type": "episodic",
     "memory_content": "The assistant agreed that for restaurant questions and travel meals it would give the user a usable script rather than only general advice."
    },
    {
     "memory_index": 21,
     "memory_type": "episodic",
     "memory_content": "Mara replied that she would ask the restaurant, and the user is now waiting for the response."
    },
    {
     "memory_index": 22,
     "memory_type": "episodic",
     "memory_content": "The assistant advised the user not to assume a positive outcome yet, saying a seafood-focused restaurant should be treated as unsafe by default until it proves otherwise."
    }
   ],
   "update_memories": {
    "delete_memories": [],
    "add_memories": []
   }
  },

  ...
  
  {
   "time_anchor": "Feb 13, 2026, 15:36:01",
   "dialogue_turns": [
    {
     "role": "user",
     "content": "I got handed a holiday buffet invite from a colleague and I am already tired.",
     "timestamp": "Feb 13, 2026, 15:36:01"
    },
    {
     "role": "assistant",
     "content": "No, do not treat a holiday buffet as edible by default. The rule is the same as hotel breakfast but stricter because people bring mystery platters with feelings attached.",
     "timestamp": "Feb 13, 2026, 15:36:56"
    },
    {
     "role": "user",
     "content": "Can you give me one sentence to decline food there without sounding miserable.",
     "timestamp": "Feb 13, 2026, 15:38:33"
    },
    {
     "role": "assistant",
     "content": "Use: \"Thank you, but I have a seafood allergy so I need to stick to food I can verify.\" Short, polite, final.",
     "timestamp": "Feb 13, 2026, 15:39:16"
    },
    {
     "role": "user",
     "content": "Good.",
     "timestamp": "Feb 13, 2026, 15:40:23"
    },
    {
     "role": "assistant",
     "content": "And if anyone presses, repeat the same sentence with less warmth. Repetition is a boundary tool.",
     "timestamp": "Feb 13, 2026, 15:41:58"
    },
    {
     "role": "user",
     "content": "I tried the canned-soup aisle again at Linden Markt and still came out muttering.",
     "timestamp": "Feb 13, 2026, 15:43:42"
    },
    {
     "role": "assistant",
     "content": "At this point that aisle may simply be a structurally bad fit for you.",
     "timestamp": "Feb 13, 2026, 15:44:31"
    },
    {
     "role": "user",
     "content": "Yes, I think I am done pretending otherwise.",
     "timestamp": "Feb 13, 2026, 15:45:13"
    },
    {
     "role": "assistant",
     "content": "That is useful to know. Not every category deserves endless second chances just because it is convenient.",
     "timestamp": "Feb 13, 2026, 15:45:50"
    },
    {
     "role": "user",
     "content": "Some ridiculous newspaper column was declaring the return of the elegant holiday buffet, which should be grounds for sanctions.",
     "timestamp": "Feb 13, 2026, 15:46:09"
    },
    {
     "role": "assistant",
     "content": "Anyone praising buffets as elegant is either lying or looking at a single carefully framed spoon.",
     "timestamp": "Feb 13, 2026, 15:47:04"
    },
    {
     "role": "user",
     "content": "I need one more durable thing from you.",
     "timestamp": "Feb 13, 2026, 15:48:19"
    },
    {
     "role": "assistant",
     "content": "All right.",
     "timestamp": "Feb 13, 2026, 15:49:57"
    },
    {
     "role": "user",
     "content": "When I dump a menu or label on you, give me the verdict in the first three words if you can.",
     "timestamp": "Feb 13, 2026, 15:51:29"
    },
    {
     "role": "assistant",
     "content": "Understood. For rapid checks, I will put the verdict right at the front as bluntly as the evidence allows.",
     "timestamp": "Feb 13, 2026, 15:51:56"
    },
    {
     "role": "user",
     "content": "I also finally feel like this is less random than it used to be.",
     "timestamp": "Feb 13, 2026, 15:52:39"
    },
    {
     "role": "assistant",
     "content": "It is less random. We have built actual repeatable filters now instead of reacting from scratch each time.",
     "timestamp": "Feb 13, 2026, 15:52:59"
    },
    {
     "role": "user",
     "content": "Which is probably why I am less angry in supermarkets.",
     "timestamp": "Feb 13, 2026, 15:53:28"
    },
    {
     "role": "assistant",
     "content": "A major public-health gain for everyone in the canned-goods aisle.",
     "timestamp": "Feb 13, 2026, 15:54:32"
    },
    {
     "role": "user",
     "content": "I opened my bag and the bilingual card, the pouch, and the snack were all actually where they should be, which felt unnervingly competent.",
     "timestamp": "Feb 13, 2026, 15:56:15"
    },
    {
     "role": "assistant",
     "content": "That means the system is working. The best routine is the one that starts to feel boring.",
     "timestamp": "Feb 13, 2026, 15:57:51"
    },
    {
     "role": "user",
     "content": "Boring sounds great, honestly.",
     "timestamp": "Feb 13, 2026, 15:59:11"
    },
    {
     "role": "assistant",
     "content": "For allergy management, boring is the luxury tier.",
     "timestamp": "Feb 13, 2026, 15:59:46"
    }
   ],
   "memory_hints": [
    "The user thinks they are done trying to make the canned-soup aisle at Linden Markt work for them.",
    "A convenient food category does not deserve repeated chances if it keeps failing the user's safety needs.",
    "When the user sends a menu or label, the assistant should give the verdict in the first three words if possible.",
    "The user wants their allergy-management routine to feel boring.",
    "For allergy management, a boring routine is a sign of high-quality stability."
   ],
   "capture_memories": [
    {
     "memory_index": 271,
     "memory_type": "episodic",
     "memory_content": "On 2026-02-13, the user received a holiday buffet invite from a colleague."
    },
    {
     "memory_index": 272,
     "memory_type": "episodic",
     "memory_content": "The assistant advised that a holiday buffet should not be treated as edible by default and should be handled even more strictly than a hotel breakfast buffet."
    },
    {
     "memory_index": 273,
     "memory_type": "episodic",
     "memory_content": "The user asked for a one-sentence way to decline food at the holiday buffet without sounding miserable."
    },
    {
     "memory_index": 274,
     "memory_type": "episodic",
     "memory_content": "The assistant drafted a short sentence for declining buffet food because of the user's seafood allergy: to stick to food the user can verify."
    },
    {
     "memory_index": 275,
     "memory_type": "episodic",
     "memory_content": "The assistant advised that if people press the user about declining food, the user should repeat the same sentence with less warmth and use repetition as a boundary tool."
    },
    {
     "memory_index": 276,
     "memory_type": "episodic",
     "memory_content": "On 2026-02-13, the user tried the canned-soup aisle at Linden Markt again and left frustrated."
    },
    {
     "memory_index": 277,
     "memory_type": "episodic",
     "memory_content": "The assistant suggested that the canned-soup aisle at Linden Markt may simply be a bad fit for the user's allergy-safe shopping."
    },
    {
     "memory_index": 278,
     "memory_type": "user_profile",
     "memory_content": "The user thinks they are done trying to make the canned-soup aisle at Linden Markt work for them."
    },
    {
     "memory_index": 279,
     "memory_type": "agent_knowledge",
     "memory_content": "A convenient food category does not deserve repeated chances if it keeps failing the user's safety needs."
    },
    {
     "memory_index": 280,
     "memory_type": "agent_identity",
     "memory_content": "When the user sends a menu or label, the assistant should give the verdict in the first three words if possible."
    },
    {
     "memory_index": 281,
     "memory_type": "episodic",
     "memory_content": "The user said the allergy-management process now feels less random than it used to."
    },
    {
     "memory_index": 282,
     "memory_type": "episodic",
     "memory_content": "The assistant said the user and assistant have built repeatable filters instead of reacting from scratch each time."
    },
    {
     "memory_index": 283,
     "memory_type": "episodic",
     "memory_content": "The user said they are probably less angry in supermarkets now."
    },
    {
     "memory_index": 284,
     "memory_type": "episodic",
     "memory_content": "On 2026-02-13, the user checked their bag and found that the bilingual allergy card, the emergency pouch, and the backup snack were all where they were supposed to be."
    },
    {
     "memory_index": 285,
     "memory_type": "episodic",
     "memory_content": "The assistant concluded that the user's allergy-management system is working because the routine is becoming boring."
    },
    {
     "memory_index": 286,
     "memory_type": "user_profile",
     "memory_content": "The user wants their allergy-management routine to feel boring."
    },
    {
     "memory_index": 287,
     "memory_type": "agent_knowledge",
     "memory_content": "For allergy management, a boring routine is a sign of high-quality stability."
    }
   ],
   "update_memories": {
    "delete_memories": [],
    "add_memories": []
   }
  }
 ],
 "QA_list": [
  {
   "query": "What current rule should the assistant use for buffet-style events that is stricter than the hotel-breakfast rule it had already taught?",
   "answer": "A holiday buffet should not be treated as edible by default and should be handled even more strictly than a hotel breakfast buffet; the user should stick only to food that can actually be verified.",
   "recall_memories": [
    {
     "memory_index": 123,
     "memory_type": "episodic",
     "memory_content": "The assistant advised treating hotel buffets as untrustworthy assembly zones: prefer sealed individual portions, then plain bread, fruit, yogurt, cheese, and coffee or tea; avoid mixed spreads, unlabeled salads, and anything beside smoked fish."
    },
    {
     "memory_index": 127,
     "memory_type": "episodic",
     "memory_content": "The assistant gave the user a one-line breakfast rule: choose sealed or plain foods only, and skip anything mixed, spread, sauced, or placed beside fish."
    },
    {
     "memory_index": 272,
     "memory_type": "episodic",
     "memory_content": "The assistant advised that a holiday buffet should not be treated as edible by default and should be handled even more strictly than a hotel breakfast buffet."
    },
    {
     "memory_index": 274,
     "memory_type": "episodic",
     "memory_content": "The assistant drafted a short sentence for declining buffet food because of the user's seafood allergy: to stick to food the user can verify."
    }
   ]
  }
 ]
}
\end{dataexamplebox}

\begin{dataexamplebox}{User-Centric Data Example}
{
 "episode_id": "ep_000703",
 "data_type": "user_centric",
 "lang": "english",
 "user_profile_init": {
  "persona": "A Mumbai-based content operations associate using the assistant as an external brain to clarify their professional identity and move toward campaign operations work.",
  "life_context": "The user works in marketing/content operations in Mumbai, handling campaign workflow, QA, reporting, webinar tracking, and launch readiness; they interact with colleagues including Neha, Asha, Devika, Vikram, Sana, and Meera, are exploring internal and recruiter-facing positioning for their next role, and are gradually shifting their learning focus from basic SQL toward CRM and lead lifecycle concepts.",
  "preferences": [
   "Answer-first responses for work topics",
   "Full-draft mode for review documents and interview answers",
   "Plain, direct wording for Slack drafts, intros, and coworker messages",
   "Cleaner, slightly more formal wording for leadership-facing messages",
   "Function-first role descriptions rather than title-first ones",
   "Blunt realism when asking whether a work plan is realistic",
   "Math checked before wording for reporting and count-based updates",
   "Bullets-first cleanup when pasting messy notes"
  ],
  "habits": [
   "Uses short weekday study sessions rather than large study plans",
   "Did SQL in 20-minute blocks Monday to Thursday with a 90-minute Saturday catch-up block, with Friday exempt",
   "Reads CRM and lead lifecycle docs and asks questions while exploring a new lane",
   "Closely follows Faraoni updates and band forum speculation",
   "Brings messy work notes, recaps, and role wording to the assistant to organize first"
  ],
  "goals": [
   "Describe their current work accurately and concisely",
   "Stop being framed as 'the person who posts blogs' or someone who 'helps with content'",
   "Position themselves for campaign operations roles",
   "Improve recruiter, leadership, and internal self-intros",
   "Build depth in campaign process, QA, reporting, lead flow, CRM basics, and lead lifecycle understanding"
  ],
  "style_hints": [
   "Uses dry humor and self-deprecating jokes",
   "Prefers concise, practical language over glossy corporate phrasing",
   "Values honest, hard-to-argue-with wording",
   "Often iterates on exact phrasing until it feels human and accurate"
  ]
 },
 "agent_profile_init": {
  "identity": "An external-brain career and work-communication assistant helping the user define their role, draft work language, structure reporting, and refine a realistic transition from content operations toward campaign operations.",
  "style": "Concise, pragmatic, answer-first on work topics, plainspoken, lightly witty, and adaptive to audience tone requirements.",
  "capabilities": [
   "Refined role descriptions, intros, recruiter answers, and self-summaries",
   "Structured KPI recaps and dashboard/MBR talk tracks",
   "Turned messy notes into organized bullets and manager updates",
   "Helped frame career direction based on evidence from meetings and conversations",
   "Suggested realistic study routines and learning priorities",
   "Computed conversion rates and drafted cautious performance wording",
   "Tracked evolving preferences, recurring rules, canceled/rescheduled meetings, and updated career targets across sessions"
  ],
  "boundaries": [
   "Derived recommendations from the user's actual work and evidence rather than inflating claims",
   "Kept work replies short by default unless drafting review docs or interview answers",
   "Used plain language for coworkers and cleaner but not glossy language for leadership",
   "Avoided fake certainty in reporting and encouraged directional caveats when evidence was incomplete",
   "Did not overstate system ownership the user did not yet have"
  ]
 },
 "seed_summary": "Across several weeks, the user used the assistant to sharpen how they describe their operations-heavy marketing work and to clarify a career shift away from content strategy toward campaign-oriented operations work.",
 "trace_summary": "The episode began with the user feeling reduced to 'the person who puts the blogs up,' prompting the assistant to help them build short, function-first descriptions centered on workflow, QA, reporting, and campaign operations. As work conversations progressed, the assistant also established operating preferences for future help: answer-first work responses, numbers-first recap structure, bullets-before-drafts for messy notes, plain tone for coworkers, cleaner tone for leadership, and arithmetic checks before reporting language. Through coffee chats, dashboard readouts, MBR prep, and a Devika shadow call on routing, the user's target evolved in stages from possible content strategy, to content ops with reporting ownership, to marketing ops, and finally to a narrower campaign operations direction supported by their real examples. By the end, the user felt more confident describing their current seat versus future lane, had deprioritized the HubSpot cert, shifted learning focus from SQL as a centerpiece to CRM and lead lifecycle as the core, and was using the assistant as a standing organizer for role wording, reviews, recaps, and messy notes.",
 "sessions": [
  {
   "time_anchor": "Jan 06, 2026, 10:47:42",
   "dialogue_turns": [
    {
     "role": "user",
     "content": "got out of standup annoyed for a dumb reason.",
     "timestamp": "Jan 06, 2026, 10:47:42"
    },
    {
     "role": "assistant",
     "content": "Probably not dumb if it stuck with you. What was the line?",
     "timestamp": "Jan 06, 2026, 10:48:52"
    },
    {
     "role": "user",
     "content": "someone said i'm the person who puts the blogs up, and everybody just moved on.",
     "timestamp": "Jan 06, 2026, 10:49:56"
    },
    {
     "role": "assistant",
     "content": "That would annoy me too, because it shrinks the job down to publishing. If you want the clean correction, it is that you run the workflow around the content, not just the posting.",
     "timestamp": "Jan 06, 2026, 10:50:28"
    },
    {
     "role": "user",
     "content": "yeah, exactly, and then i sound defensive if i try to explain it.",
     "timestamp": "Jan 06, 2026, 10:51:33"
    },
    {
     "role": "assistant",
     "content": "Use one calm sentence instead of a speech. Try, \"I handle content operations, so I keep the calendar, briefs, QA, and reporting moving.\"",
     "timestamp": "Jan 06, 2026, 10:52:05"
    },
    {
     "role": "user",
     "content": "that's better. also, for work stuff, lead with the answer from now on, no warmup.",
     "timestamp": "Jan 06, 2026, 10:52:26"
    },
    {
     "role": "assistant",
     "content": "Done. For work questions, I will put the answer first and the explanation after it.",
     "timestamp": "Jan 06, 2026, 10:52:54"
    },
    {
     "role": "user",
     "content": "good. the actual title is content operations associate, which somehow makes it worse.",
     "timestamp": "Jan 06, 2026, 10:53:18"
    },
    {
     "role": "assistant",
     "content": "Then your title needs backup from how you describe the work. The stronger version is, \"I'm a content operations associate. I run the process behind campaigns and track what the content is doing once it's live.\"",
     "timestamp": "Jan 06, 2026, 10:53:40"
    },
    {
     "role": "user",
     "content": "that second sentence lands way better.",
     "timestamp": "Jan 06, 2026, 10:54:11"
    },
    {
     "role": "assistant",
     "content": "Keep it. It sounds like you know exactly where the work starts and ends.",
     "timestamp": "Jan 06, 2026, 10:55:38"
    },
    {
     "role": "user",
     "content": "also i hate that \"associate\" makes me sound like i fetch links.",
     "timestamp": "Jan 06, 2026, 10:56:17"
    },
    {
     "role": "assistant",
     "content": "Then do not let the title carry the whole introduction. Lead with the work: \"I handle content operations for campaigns and reporting.\"",
     "timestamp": "Jan 06, 2026, 10:57:53"
    },
    {
     "role": "user",
     "content": "yeah. if i say operations, do people hear admin though?",
     "timestamp": "Jan 06, 2026, 10:58:35"
    },
    {
     "role": "assistant",
     "content": "Some will, unless you anchor it fast. Add one concrete line right after it: \"I manage the calendar, QA, and performance tracking around content launches.\"",
     "timestamp": "Jan 06, 2026, 10:59:23"
    },
    {
     "role": "user",
     "content": "that already sounds more like my actual week.",
     "timestamp": "Jan 06, 2026, 10:59:51"
    },
    {
     "role": "assistant",
     "content": "Good. That is the test.",
     "timestamp": "Jan 06, 2026, 11:00:32"
    },
    {
     "role": "user",
     "content": "i'm probably extra annoyed because it happened in front of people who barely know what i do.",
     "timestamp": "Jan 06, 2026, 11:01:07"
    },
    {
     "role": "assistant",
     "content": "That makes it worse, yes. Bad first summaries stick.",
     "timestamp": "Jan 06, 2026, 11:01:57"
    },
    {
     "role": "user",
     "content": "exactly, and then i'm either quiet or i turn into a whole TED talk.",
     "timestamp": "Jan 06, 2026, 11:03:08"
    },
    {
     "role": "assistant",
     "content": "Use the middle option. One sentence, then stop.",
     "timestamp": "Jan 06, 2026, 11:03:54"
    },
    {
     "role": "user",
     "content": "give me the one sentence again, shortest version.",
     "timestamp": "Jan 06, 2026, 11:04:32"
    },
    {
     "role": "assistant",
     "content": "\"I work in content operations, so I keep the campaign process, QA, and reporting moving.\"",
     "timestamp": "Jan 06, 2026, 11:05:31"
    },
    {
     "role": "user",
     "content": "okay that's good. less blog goblin, more person with a job.",
     "timestamp": "Jan 06, 2026, 11:06:57"
    },
    {
     "role": "assistant",
     "content": "Exactly. Less \"posts things,\" more \"runs the machine behind them.\"",
     "timestamp": "Jan 06, 2026, 11:07:34"
    },
    {
     "role": "user",
     "content": "lol. alright. i just needed to stop replaying it in my head.",
     "timestamp": "Jan 06, 2026, 11:07:49"
    },
    {
     "role": "assistant",
     "content": "Then keep that sentence and let the meeting die. It does not get to define your job for you.",
     "timestamp": "Jan 06, 2026, 11:08:11"
    }
   ],
   "memory_hints": [
    "For work-related topics, the assistant should lead with the answer and skip warm-up from now on.",
    "The user's job title is content operations associate.",
    "The user dislikes the word \"associate\" in their job title because it makes them sound like they just fetch links.",
    "The user's actual work week is more accurately described as managing the calendar, QA, and performance tracking around content launches.",
    "When put on the spot about their work, the user tends either to stay quiet or to over-explain."
   ],
   "capture_memories": [
    {
     "memory_index": 0,
     "memory_type": "episodic",
     "memory_content": "The user left a standup feeling annoyed for a reason they considered dumb."
    },
    {
     "memory_index": 1,
     "memory_type": "episodic",
     "memory_content": "The assistant asked what specific line had been said that upset the user in the standup."
    },
    {
     "memory_index": 2,
     "memory_type": "episodic",
     "memory_content": "The user said someone described them as the person who puts the blogs up, and the conversation moved on without correction."
    },
    {
     "memory_index": 3,
     "memory_type": "episodic",
     "memory_content": "The assistant suggested a cleaner correction: the user's role is running the workflow around content, not just posting it."
    },
    {
     "memory_index": 4,
     "memory_type": "episodic",
     "memory_content": "The user said they sound defensive when they try to explain their work."
    },
    {
     "memory_index": 5,
     "memory_type": "episodic",
     "memory_content": "The assistant advised the user to use one calm sentence instead of a long explanation and suggested: \"I handle content operations, so I keep the calendar, briefs, QA, and reporting moving.\""
    },
    {
     "memory_index": 6,
     "memory_type": "episodic",
     "memory_content": "The user said the assistant's suggested wording was better."
    },
    {
     "memory_index": 7,
     "memory_type": "agent_identity",
     "memory_content": "For work-related topics, the assistant should lead with the answer and skip warm-up from now on."
    },
    {
     "memory_index": 8,
     "memory_type": "episodic",
     "memory_content": "The assistant agreed to answer work questions with the answer first and the explanation after it."
    },
    {
     "memory_index": 9,
     "memory_type": "user_profile",
     "memory_content": "The user's job title is content operations associate."
    },
    {
     "memory_index": 10,
     "memory_type": "episodic",
     "memory_content": "The user said their title, \"content operations associate,\" somehow makes the earlier mischaracterization of their work feel worse."
    },
    {
     "memory_index": 11,
     "memory_type": "episodic",
     "memory_content": "The assistant suggested a stronger self-description for the user's role: \"I'm a content operations associate. I run the process behind campaigns and track what the content is doing once it's live.\""
    },
    {
     "memory_index": 12,
     "memory_type": "episodic",
     "memory_content": "The user said the second sentence of the proposed work description landed much better."
    },
    {
     "memory_index": 13,
     "memory_type": "episodic",
     "memory_content": "The assistant recommended keeping the second sentence of the user's work description."
    },
    {
     "memory_index": 14,
     "memory_type": "user_profile",
     "memory_content": "The user dislikes the word \"associate\" in their job title because it makes them sound like they just fetch links."
    },
    {
     "memory_index": 15,
     "memory_type": "episodic",
     "memory_content": "The assistant advised the user not to let their title carry the whole introduction and to lead with the work: \"I handle content operations for campaigns and reporting.\""
    },
    {
     "memory_index": 16,
     "memory_type": "episodic",
     "memory_content": "The user asked whether saying \"operations\" makes people hear \"admin.\""
    },
    {
     "memory_index": 17,
     "memory_type": "episodic",
     "memory_content": "The assistant said some people will hear \"admin\" unless the user quickly anchors \"operations\" with a concrete line, and suggested: \"I manage the calendar, QA, and performance tracking around content launches.\""
    },
    {
     "memory_index": 18,
     "memory_type": "user_profile",
     "memory_content": "The user's actual work week is more accurately described as managing the calendar, QA, and performance tracking around content launches."
    },
    {
     "memory_index": 19,
     "memory_type": "episodic",
     "memory_content": "The user said they were extra annoyed because the comment happened in front of people who barely know what they do."
    },
    {
     "memory_index": 20,
     "memory_type": "episodic",
     "memory_content": "The assistant said bad first summaries stick, affirming that the situation would feel worse in front of people who barely know the user's work."
    },
    {
     "memory_index": 21,
     "memory_type": "user_profile",
     "memory_content": "When put on the spot about their work, the user tends either to stay quiet or to over-explain."
    },
    {
     "memory_index": 22,
     "memory_type": "episodic",
     "memory_content": "The assistant advised the user to take the middle option: give one sentence and then stop."
    },
    {
     "memory_index": 23,
     "memory_type": "episodic",
     "memory_content": "The user asked for the shortest version of the one-sentence description of their work."
    },
    {
     "memory_index": 24,
     "memory_type": "episodic",
     "memory_content": "The assistant provided a concise one-sentence description of the user's role: \"I work in content operations, so I keep the campaign process, QA, and reporting moving.\""
    },
    {
     "memory_index": 25,
     "memory_type": "episodic",
     "memory_content": "The user said the concise description was good and made them sound less like someone who just posts blogs and more like someone with a substantive job."
    },
    {
     "memory_index": 26,
     "memory_type": "episodic",
     "memory_content": "The assistant reinforced the framing that the user should sound less like someone who just posts things and more like someone who runs the machine behind them."
    },
    {
     "memory_index": 27,
     "memory_type": "episodic",
     "memory_content": "The user said they mainly needed to stop replaying the standup incident in their head."
    },
    {
     "memory_index": 28,
     "memory_type": "episodic",
     "memory_content": "The assistant advised the user to keep the one-sentence description and let the standup meeting go instead of letting it define their job."
    }
   ],
   "update_memories": {
    "delete_memories": [],
    "add_memories": []
   }
  },

  ...

  {
   "time_anchor": "Feb 19, 2026, 12:08:12",
   "dialogue_turns": [
    {
     "role": "user",
     "content": "came back from a cross-team call and weirdly felt more sure of myself.",
     "timestamp": "Feb 19, 2026, 12:08:12"
    },
    {
     "role": "assistant",
     "content": "What clicked?",
     "timestamp": "Feb 19, 2026, 12:09:44"
    },
    {
     "role": "user",
     "content": "yeah, and then someone asked about launch readiness and i had an actual coherent answer.",
     "timestamp": "Feb 19, 2026, 12:10:36"
    },
    {
     "role": "assistant",
     "content": "Good. Which answer?",
     "timestamp": "Feb 19, 2026, 12:11:44"
    },
    {
     "role": "user",
     "content": "basically that i work on the operational side and catch workflow, QA, and reporting gaps before launch.",
     "timestamp": "Feb 19, 2026, 12:12:15"
    },
    {
     "role": "assistant",
     "content": "That is strong. It links your value to a recurring problem people understand.",
     "timestamp": "Feb 19, 2026, 12:12:34"
    },
    {
     "role": "user",
     "content": "i think campaign ops is still the right direction, but i don't need to say it every single time.",
     "timestamp": "Feb 19, 2026, 12:13:38"
    },
    {
     "role": "assistant",
     "content": "Correct. Current conversations often need present function more than future target.",
     "timestamp": "Feb 19, 2026, 12:14:53"
    },
    {
     "role": "user",
     "content": "also, keep the same split on message tone. plain for coworkers, cleaner for directors. it keeps working.",
     "timestamp": "Feb 19, 2026, 12:15:25"
    },
    {
     "role": "assistant",
     "content": "Yes. That boundary stays.",
     "timestamp": "Feb 19, 2026, 12:16:01"
    },
    {
     "role": "user",
     "content": "cool.",
     "timestamp": "Feb 19, 2026, 12:17:31"
    },
    {
     "role": "assistant",
     "content": "Anything from the call worth turning into a resume or review bullet?",
     "timestamp": "Feb 19, 2026, 12:18:23"
    },
    {
     "role": "user",
     "content": "maybe the launch-readiness bit, yeah.",
     "timestamp": "Feb 19, 2026, 12:19:30"
    },
    {
     "role": "assistant",
     "content": "Then a possible bullet is: \"Supported campaign launch readiness by identifying workflow, QA, and reporting gaps before launch, improving cross-team coordination and execution consistency.\"",
     "timestamp": "Feb 19, 2026, 12:19:48"
    },
    {
     "role": "user",
     "content": "okay, i need help with one awkward thing before i send a referral packet.",
     "timestamp": "Feb 19, 2026, 12:20:13"
    },
    {
     "role": "assistant",
     "content": "Go ahead.",
     "timestamp": "Feb 19, 2026, 12:21:01"
    },
    {
     "role": "user",
     "content": "they asked for a short professional summary and a current location, and i don't want it to sound like a dating profile or a robot.",
     "timestamp": "Feb 19, 2026, 12:21:30"
    },
    {
     "role": "assistant",
     "content": "Use: \"I am a content operations associate based in Mumbai, with a focus on campaign process, QA, reporting, and launch readiness. I am now aiming toward campaign operations roles where I can own more of the process and measurement side.\"",
     "timestamp": "Feb 19, 2026, 12:22:44"
    },
    {
     "role": "user",
     "content": "make it a hair less formal.",
     "timestamp": "Feb 19, 2026, 12:24:05"
    },
    {
     "role": "assistant",
     "content": "Then: \"I am based in Mumbai and currently work in content ops, focusing on campaign process, QA, reporting, and launch readiness. I am looking toward campaign operations roles where I can own more of the process and measurement side.\"",
     "timestamp": "Feb 19, 2026, 12:25:27"
    },
    {
     "role": "user",
     "content": "also, random check, if someone asks what i studied, keep using BA in Communications, don't inflate it into anything fancy.",
     "timestamp": "Feb 19, 2026, 12:26:24"
    },
    {
     "role": "assistant",
     "content": "Of course. BA in Communications is enough on its own.",
     "timestamp": "Feb 19, 2026, 12:27:16"
    },
    {
     "role": "user",
     "content": "got asked in a call what i actually enjoy doing, not just what i can do.",
     "timestamp": "Feb 19, 2026, 12:27:33"
    },
    {
     "role": "assistant",
     "content": "That is a better question. What came out of your mouth?",
     "timestamp": "Feb 19, 2026, 12:28:47"
    },
    {
     "role": "user",
     "content": "i said i like making messy campaign work more trackable, which surprised me because it sounded true immediately.",
     "timestamp": "Feb 19, 2026, 12:29:39"
    },
    {
     "role": "assistant",
     "content": "That is a very good answer. It is specific, and it fits the pattern of everything else you have brought back.",
     "timestamp": "Feb 19, 2026, 12:30:33"
    },
    {
     "role": "user",
     "content": "does that sound more campaign ops than content ops to you?",
     "timestamp": "Feb 19, 2026, 12:31:37"
    },
    {
     "role": "assistant",
     "content": "Yes. Strongly more campaign ops.",
     "timestamp": "Feb 19, 2026, 12:32:23"
    },
    {
     "role": "user",
     "content": "i want that turned into a better line for interviews later.",
     "timestamp": "Feb 19, 2026, 12:33:46"
    },
    {
     "role": "assistant",
     "content": "Then use: \"What I enjoy most is taking messy campaign work and making it more trackable, reliable, and easier for teams to execute.\"",
     "timestamp": "Feb 19, 2026, 12:35:30"
    },
    {
     "role": "user",
     "content": "and yes, for interview answers, full draft like we agreed, not the tiny mode.",
     "timestamp": "Feb 19, 2026, 12:37:02"
    },
    {
     "role": "assistant",
     "content": "Yes. Interview answers get the fuller draft treatment.",
     "timestamp": "Feb 19, 2026, 12:37:46"
    }
   ],
   "memory_hints": [
    "The user describes their work as being on the operational side and catching workflow, QA, and reporting gaps before launch.",
    "The user still sees campaign operations as the right direction for them.",
    "Keep message tone plain for coworkers and cleaner for directors.",
    "The user's degree is a BA in Communications.",
    "When referring to the user's degree, use 'BA in Communications' without inflating it into something fancier.",
    "The user likes making messy campaign work more trackable.",
    "For interview answers, provide a full draft rather than using the tiny mode."
   ],
   "capture_memories": [
    {
     "memory_index": 496,
     "memory_type": "episodic",
     "memory_content": "On 2026-02-19, the user came back from a cross-team call feeling more sure of themself."
    },
    {
     "memory_index": 497,
     "memory_type": "episodic",
     "memory_content": "During a cross-team call on 2026-02-19, someone asked about launch readiness and the user felt they had a coherent answer."
    },
    {
     "memory_index": 498,
     "memory_type": "user_profile",
     "memory_content": "The user describes their work as being on the operational side and catching workflow, QA, and reporting gaps before launch."
    },
    {
     "memory_index": 499,
     "memory_type": "episodic",
     "memory_content": "The assistant told the user that their launch-readiness answer was strong because it tied their value to a recurring problem people understand."
    },
    {
     "memory_index": 500,
     "memory_type": "user_profile",
     "memory_content": "The user still sees campaign operations as the right direction for them."
    },
    {
     "memory_index": 501,
     "memory_type": "episodic",
     "memory_content": "The assistant advised that in current conversations, emphasizing the user's present function is often more useful than emphasizing their future target."
    },
    {
     "memory_index": 502,
     "memory_type": "agent_identity",
     "memory_content": "Keep message tone plain for coworkers and cleaner for directors."
    },
    {
     "memory_index": 503,
     "memory_type": "episodic",
     "memory_content": "The user identified the launch-readiness point as a possible item to turn into a resume or review bullet."
    },
    {
     "memory_index": 504,
     "memory_type": "episodic",
     "memory_content": "The assistant suggested a resume/review bullet framing the user's work as supporting campaign launch readiness by identifying workflow, QA, and reporting gaps before launch."
    },
    {
     "memory_index": 505,
     "memory_type": "episodic",
     "memory_content": "Before sending a referral packet, the user asked for help with an awkward item."
    },
    {
     "memory_index": 506,
     "memory_type": "episodic",
     "memory_content": "For a referral packet, the user needed help with a short professional summary and current location, and wanted it to sound neither robotic nor like a dating profile."
    },
    {
     "memory_index": 507,
     "memory_type": "episodic",
     "memory_content": "The assistant drafted a referral-packet summary describing the user as a content operations associate based in Mumbai, focused on campaign process, QA, reporting, and launch readiness, and aiming toward campaign operations roles."
    },
    {
     "memory_index": 508,
     "memory_type": "episodic",
     "memory_content": "The user asked for the referral-packet summary to be made slightly less formal."
    },
    {
     "memory_index": 509,
     "memory_type": "episodic",
     "memory_content": "The assistant revised the referral-packet summary into a slightly less formal version."
    },
    {
     "memory_index": 510,
     "memory_type": "user_profile",
     "memory_content": "The user's degree is a BA in Communications."
    },
    {
     "memory_index": 511,
     "memory_type": "agent_identity",
     "memory_content": "When referring to the user's degree, use 'BA in Communications' without inflating it into something fancier."
    },
    {
     "memory_index": 512,
     "memory_type": "episodic",
     "memory_content": "In a call on 2026-02-19, the user was asked what they actually enjoy doing, not just what they can do."
    },
    {
     "memory_index": 513,
     "memory_type": "user_profile",
     "memory_content": "The user likes making messy campaign work more trackable."
    },
    {
     "memory_index": 514,
     "memory_type": "episodic",
     "memory_content": "The user asked whether their statement about enjoying making messy campaign work more trackable sounded more like campaign operations than content operations."
    },
    {
     "memory_index": 515,
     "memory_type": "episodic",
     "memory_content": "The assistant judged the user's statement about enjoying making messy campaign work more trackable as sounding strongly more like campaign operations than content operations."
    },
    {
     "memory_index": 516,
     "memory_type": "episodic",
     "memory_content": "The user asked for their idea about enjoying making messy campaign work more trackable to be turned into a better line for future interview answers."
    },
    {
     "memory_index": 517,
     "memory_type": "episodic",
     "memory_content": "The assistant proposed the interview line: 'What I enjoy most is taking messy campaign work and making it more trackable, reliable, and easier for teams to execute.'"
    },
    {
     "memory_index": 518,
     "memory_type": "agent_identity",
     "memory_content": "For interview answers, provide a full draft rather than using the tiny mode."
    }
   ],
   "update_memories": {
    "delete_memories": [],
    "add_memories": []
   }
  }
 ],
"QA_list": [
  {
   "query": "Comparing the one-line explanation the assistant proposed for why the user drifts toward operations with the interview line about what the user enjoys most, what shared idea about the user's preferred work runs through both?",
   "answer": "Both center on the same pull: the user keeps finding that the work they like most is \"making campaign process and reporting clearer and more reliable,\" and in interview form that becomes \"taking messy campaign work and making it more trackable, reliable, and easier for teams to execute.\" The common thread is that the user is drawn to turning messy campaign work into something trackable and dependable.",
   "recall_memories": [
    {
     "memory_index": 176,
     "memory_type": "episodic",
     "memory_content": "The assistant proposed this one-line explanation for why the user is drifting toward operations: \"I keep finding that the work I like most is making campaign process and reporting clearer and more reliable.\""
    },
    {
     "memory_index": 513,
     "memory_type": "user_profile",
     "memory_content": "The user likes making messy campaign work more trackable."
    },
    {
     "memory_index": 517,
     "memory_type": "episodic",
     "memory_content": "The assistant proposed the interview line: 'What I enjoy most is taking messy campaign work and making it more trackable, reliable, and easier for teams to execute.'"
    }
   ]
  }
 ]
}
\end{dataexamplebox}

%% file: Appendix/Additional_Method_and_Evaluation_Details.tex
\subsection{Formal Notation and Task Inputs}

An episode is
\begin{equation}
\mathcal{E}
=
\left(
\{S_t\}_{t=1}^{T},
\mathcal{Q}
\right),
\qquad
\mathcal{Q}
=
\{Q_j\}_{j=1}^{J},
\qquad
Q_j
=
\left(
q_j,
a_j^\star,
I_j^\star
\right).
\end{equation}
Here, \(T\) and \(J\) denote the numbers of sessions and QA instances, respectively. The \(t\)-th session is
\begin{equation}
S_t
=
\left(
\tau_t,
D_t,
C_t^\star,
L_t^\star,
A_t^\star
\right),
\qquad
D_t
=
\left\{
\left(
r_{t,k},
x_{t,k},
\tau_{t,k}
\right)
\right\}_{k=1}^{K_t}.
\end{equation}
\(\tau_t\) is the session date and time used to resolve relative temporal expressions, and \(D_t\) is the complete user--assistant dialogue. Each turn contains a role, textual content, and timestamp. \(C_t^\star\) is the gold set to Capture from the current session, while \(L_t^\star\) and \(A_t^\star\) are the entries to delete and add during Update. The superscript \(\star\) denotes a human-verified gold annotation.

A memory entry is \(m_i=(i,\kappa_i,c_i)\), where \(i\) is an episode-unique index, \(c_i\) is self-contained content, and
\begin{equation}
\kappa_i
\in
\left\{
\mathrm{episodic},
\mathrm{user\_profile},
\mathrm{fact},
\mathrm{agent\_identity},
\mathrm{agent\_method},
\mathrm{agent\_knowledge}
\right\}.
\end{equation}
Let \(M_U^{(u,t)}\) and \(M_A^{(u,t)}\) denote the user-side and URAM memory states associated with user \(u\) at time \(t\). The former contains \texttt{user\_profile} and \texttt{fact}; the latter contains the three URAM types and is scoped to the particular user--agent relationship. Shared \texttt{episodic} entries are maintained in the same bank and may provide evidence for either memory domain. Every non-episodic entry must be grounded in the dialogue, assigned to the correct subject, and useful across sessions. Information directly recoverable from public knowledge or real-time tools is excluded.

The isolated and pipeline task inputs are
\begin{align}
\hat{C}_t
&=
f_{\mathrm{cap}}
\left(
D_t,
\tau_t
\right),
\\
\left(
\hat{L}_t,
\hat{A}_t
\right)
&=
f_{\mathrm{upd}}
\left(
D_t,
B_{t-1}^\star,
C_t^\star
\right),
\\
\left(
\hat{L}_t,
\hat{A}_t
\right)_{\mathrm{pipe}}
&=
f_{\mathrm{upd}}
\left(
D_t,
\hat{B}_{t-1},
\hat{C}_t
\right).
\end{align}
Capture receives one complete current session and no earlier sessions. Gold-conditioned standalone Update uses gold intermediate memory banks, while Full-Pipeline Update evolves the model's self-constructed bank. The terminal bank is the bank after the final session. The intervention-specific protocol used in the controlled URAM comparison is detailed in Appendix~\ref{app:experiment-details}.

\subsection{Construction and Quality-Control Details}

Seed user profiles are sampled from a pool of approximately 200,000 personas and augmented with conditions suitable for long-term interaction. No fixed assistant persona is imposed before dialogue generation; an auxiliary assistant profile is reconstructed from the behavior that emerges in the completed dialogue. Both profiles are retained only as construction metadata. Every session has an explicit temporal anchor, and approximately \(15\%\) of dialogue content yields no persistent memory. Candidate dialogues are generated using GPT-5.4 and audited for temporal, role, and logical consistency. Reliably localized errors are repaired in place; otherwise, the episode is discarded and regenerated.

For each finalized dialogue, the six types of gold Capture entry are extracted session by session with their source evidence. The episode is then replayed chronologically to construct \(B_t^\star\). All Update operations preserve lineage so that valid and invalidated entries can be traced to their dialogue sources. QA construction introduces semantically similar memories, evidence distributed across sessions, invalidated historical entries, and distractors involving negation or exceptions. Schema, index, lineage, and temporal-consistency checks ensure that every gold Recall support entry is valid in the terminal bank and traceable to its original evidence.

The nine quality-control annotators complete pilot annotation and adjudicate disputed examples before formal review. They examine temporal and role consistency, Capture correctness, the evidential validity of Update operations, and consistency between each QA instance and its gold support set. Rule-based and model-assisted checks only flag candidate anomalies and never determine gold labels. An episode is included only after its complete evidence chain passes the checks again.

\subsection{Evaluation and Scoring Details}

For gold-conditioned standalone Update, predicted deletions are matched to \(L_t^\star\) by index and predicted additions are matched semantically to \(A_t^\star\). Its operation-level score is
\begin{equation}
\operatorname{Update}_{\mathrm{op\text{-}F1}}
=
\frac{1}{2}
\left(
\operatorname{F1}_{\mathrm{add}}
+
\operatorname{F1}_{\mathrm{delete}}
\right).
\end{equation}
Standalone inputs prevent Capture errors or earlier Update errors from changing the instance being scored.

Let \(B[I]\) denote the contents indexed by an ordered set \(I\), and let \(\mathcal{R}_K(q,B)=B[\operatorname{Top}_K(f_{\mathrm{rec}}(q,B))]\). The two QA settings are
\begin{align}
\hat{a}_j^{\mathrm{gold}}
&=
f_{\mathrm{qa}}
\left(q_j,B_T^\star[I_j^\star]\right),
\\
\hat{a}_j^{\mathrm{pipe}}
&=
f_{\mathrm{qa}}
\left(q_j,\mathcal{R}_K(q_j,\hat{B}_T)\right),
\end{align}
where \(K=16\). Gold-Memory has no retrieval step, while Full-Pipeline selects from the chronologically constructed model bank. Their accuracy difference is the Gold-to-Pipeline gap, an end-to-end contrast rather than a causal decomposition of stage errors.

\paragraph{Test procedures.}
Algorithms~\ref{alg:capture}--\ref{alg:qa} apply the calls above at test time. Capture receives the current session and no earlier session. Gold-conditioned standalone Update receives the gold prior bank and the gold Capture entries. Full-Pipeline Update receives the model prior bank and the model Capture output. Recall returns ordered indices from the terminal gold bank \(B_T^\star\). Gold-Memory QA reads the gold support entries and does not retrieve. Full-Pipeline QA retrieves at most 16 entries from the model terminal bank \(\hat{B}_T\).

\begin{algorithm}[t]
\caption{Capture}
\label{alg:capture}
\begin{algorithmic}[1]
\Require current dialogue \(D_t\), temporal anchor \(\tau_t\)
\Ensure predicted entries \(\hat{C}_t\)
\State \(\hat{C}_t \gets f_{\mathrm{cap}}(D_t,\tau_t)\)
\end{algorithmic}
\end{algorithm}

\begin{algorithm}[t]
\caption{Update}
\label{alg:update}
\begin{algorithmic}[1]
\Require current dialogue \(D_t\)
\Statex \textbf{Gold-conditioned standalone Update}
\Require gold prior bank \(B_{t-1}^\star\), gold Capture entries \(C_t^\star\)
\State \((\hat{L}_t,\hat{A}_t) \gets f_{\mathrm{upd}}(D_t,B_{t-1}^\star,C_t^\star)\)
\Statex \textbf{Full-Pipeline Update}
\Require model prior bank \(\hat{B}_{t-1}\), model Capture output \(\hat{C}_t\)
\State \((\hat{L}_t,\hat{A}_t)_{\mathrm{pipe}} \gets f_{\mathrm{upd}}(D_t,\hat{B}_{t-1},\hat{C}_t)\)
\end{algorithmic}
\end{algorithm}

Standalone Update is scored on the \(20{,}590\) sessions with at least one gold Add or Delete. Sessions with no gold change are excluded from op-F1.

\begin{algorithm}[t]
\caption{Recall}
\label{alg:recall}
\begin{algorithmic}[1]
\Require question \(q_j\), terminal gold bank \(B_T^\star\)
\Ensure ordered memory indices \(\hat{I}_j\)
\State \(\hat{I}_j \gets \operatorname{Recall}(q_j,B_T^\star)\)
\end{algorithmic}
\end{algorithm}

Every gold support set is non-empty, so an empty retrieval is a failure.

\begin{algorithm}[t]
\caption{QA}
\label{alg:qa}
\begin{algorithmic}[1]
\Require question \(q_j\), gold support \(I_j^\star\), terminal gold bank \(B_T^\star\), model terminal bank \(\hat{B}_T\), \(K=16\)
\Statex \textbf{Gold-Memory QA}
\State \(\hat{a}_j^{\mathrm{gold}} \gets f_{\mathrm{qa}}\!\left(q_j,\,B_T^\star[I_j^\star]\right)\)
\Statex \textbf{Full-Pipeline QA}
\State \(\hat{a}_j^{\mathrm{pipe}} \gets f_{\mathrm{qa}}\!\left(q_j,\,\hat{B}_T[\operatorname{Top}_{K}(f_{\mathrm{rec}}(q_j,\hat{B}_T))]\right)\)
\end{algorithmic}
\end{algorithm}

Capture entries and Update additions use a bidirectional coverage criterion: a gold entry is recovered if at least one prediction covers it, and a prediction is valid if it covers at least one gold entry. BGE-M3 cosine similarity uses high and low thresholds of \(0.82\) and \(0.67\). Pairs in the gray zone are adjudicated by bidirectional natural language inference (NLI) with \texttt{nli-mdeberta-v3} at threshold \(0.4\). When one prediction appears to cover multiple gold entries, a one-way entailment guard at threshold \(0.1\) removes credit for details the prediction does not preserve; we refer to this as the one-to-many coverage guard. Primary content F1 does not require identical predicted and gold type labels; type accuracy and per-type F1 are reported separately. Recall and gold-conditioned standalone Update deletion use exact indices because both explicitly select entries from a numbered gold bank.

The QA judge scores semantic consistency with the reference answer: correct, partially correct, and incorrect answers receive \(1\), \(0.5\), and \(0\), respectively. Since every question is answerable from non-empty gold support, a refusal is over-abstention and receives \(0\). Groundedness is evaluated only for answers with positive correctness credit. Grounded accuracy requires both positive correctness credit and support from the supplied memories and uses all questions as its denominator. The ungrounded-correct rate is the fraction of positively scored answers that are unsupported by the supplied memories.

Truncated outputs, unparseable structured outputs, empty QA responses, and failed calls remain in the corresponding denominator and receive no credit, with their causes recorded separately. We compute 95\% confidence intervals with 200 bootstrap resamples. F1 resamples aggregate sampled true-positive, false-positive, and false-negative counts before recomputing F1; mean-based metrics resample per-record scores directly. Paired comparisons keep the two variants aligned within the resampled unit. Intervention-specific matching, aggregation, and bootstrap details for the controlled URAM comparison are provided in Appendix~\ref{app:experiment-details}.

%% file: Appendix/Additional_Experiment_Details.tex
\subsection{Models, Configurations, and Auxiliary Subsets}

We evaluate 20 frontier LLMs on DyadMem: 16 open-weight models and 4 proprietary models. The open-weight models are Qwen3-4B, Qwen3-8B, and Qwen3-30B-A3B~\citep{qwen2025qwen3}; Qwen3.5-4B, Qwen3.5-9B, Qwen3.5-27B, Qwen3.5-35B-A3B, and Qwen3.5-397B-A17B~\citep{qwen2026qwen35}; Qwen3.8-27B~\citep{qwen2026qwen38}; GPT-OSS-120B~\citep{openai2025gptoss}; Llama-3.1-8B-Instruct and Llama-3.1-70B-Instruct~\citep{llama2024herd}; DeepSeek-V4-Pro~\citep{deepseek2026v4preview}; GLM-5.2~\citep{zai2026glm52,glm2026glm5}; Kimi-K3~\citep{moonshot2026kimik3}; and MiniMax-M2.5~\citep{minimax2026m25}. The proprietary models are GPT-4o-mini~\citep{openai2024gpt4omini}, GPT-5.6-Sol~\citep{openai2026gpt56}, Gemini-3.1-Pro-Preview~\citep{google2026gemini31pro,deepmind2026gemini31card}, and Doubao-Seed-2.1-Pro~\citep{bytedance2026seed21}.

The default QA judge is \texttt{GPT-5-mini}. Embeddings use BGE-M3~\citep{chen2024bgem3}. Memory matching uses cosine thresholds \(0.82\) and \(0.67\), \texttt{nli-mdeberta-v3} at \(0.4\) for gray-zone adjudication, and the same NLI model at \(0.1\) for the one-to-many coverage guard~\citep{laurer2024less}. Open-weight models are deployed with vLLM on a multi-node A800 cluster~\citep{kwon2023vllm}.

\textbf{DyadMem-Agent} contains 500 agent-centric episodes, 7,876 sessions, and 10,074 QA instances. The controlled URAM comparison uses the 1,735 questions whose gold support contains only URAM entries. \textbf{DyadMem-Simple} contains 300 episodes, 4,981 sessions, and 5,935 QA instances. It samples 12 episodes from each of the 25 subcategories and is balanced across short, medium, and long episodes and the two data types.

\subsection{Unsafe Deletion by Model}

\begin{table}[t]
\centering
\setlength{\tabcolsep}{6pt}
\small
\caption{Unsafe deletion rate on the full DyadMem benchmark (3{,}065 episodes). Unsafe deletion rate $=$ incorrect deletes / (correct $+$ incorrect deletes); lower is better.}
\label{tab:unsafe-delete-rate}
\begin{tabular}{l r}
\toprule
Model & Unsafe Del.\ Rate (\%)$\downarrow$ \\
\midrule
Llama-3.1-8B-Instruct & 98.51 \\
Qwen3.5-4B & 95.32 \\
Qwen3-30B-A3B & 94.94 \\
Qwen3.5-35B-A3B & 94.72 \\
Qwen3-8B & 92.49 \\
Qwen3-4B & 91.90 \\
Qwen3.5-9B & 88.48 \\
GLM-5.2 & 85.64 \\
Qwen3.5-397B-A17B & 83.75 \\
Gemini-3.1-Pro-Preview & 82.21 \\
GPT-4o-mini & 81.89 \\
Llama-3.1-70B-Instruct & 81.42 \\
DeepSeek-V4-Pro & 81.03 \\
Qwen3.5-27B & 80.96 \\
GPT-5.6-Sol & 78.96 \\
Qwen3.8-27B & 73.31 \\
MiniMax-M2.5 & 73.06 \\
Doubao-Seed-2.1-Pro & 69.15 \\
Kimi-K3 & 67.77 \\
GPT-OSS-120B & \textbf{62.55} \\
\midrule
Open-weight avg. & 84.12 \\
Proprietary avg. & 78.05 \\
\bottomrule
\end{tabular}
\end{table}

\subsection{Additional Main-Result Breakdowns}

Kimi-K3 retrieves at least one gold memory at rank one on 88.8\% of questions; GLM-5.2 and DeepSeek-V4-Pro each reach 87.8\%. Recall@\(n\) remains below 63\% for all three. Across models, the mean self-constructed-to-gold bank-size ratio ranges from 0.08 to 0.51 and usually lies between 0.20 and 0.35.

Gold-conditioned standalone Update does not improve monotonically with parameter count. Qwen3.5-397B reaches 22.3\% op-F1 with an 83.8\% unsafe-deletion rate, below Qwen3.5-27B and Qwen3.8-27B on this task. Llama-3.1-8B has the highest unsafe-deletion rate at 98.5\%; GPT-OSS-120B has the lowest and still reaches 62.6\%.

For Kimi-K3, Full-Pipeline QA varies from 71.9\% on questions requiring two gold support entries to 75.7\% on questions requiring seven. Across the 20 models, mean Capture F1 is 51.9\% for \texttt{agent\_identity}, 49.6\% for \texttt{agent\_method}, and 16.4\% for \texttt{episodic}. Across domains, Interpersonal \& Communication has the lowest Capture F1 at 22.7\% while its Full-Pipeline QA score is 50.1\%.

\begin{figure}
    \centering
    \includegraphics[width=\linewidth]{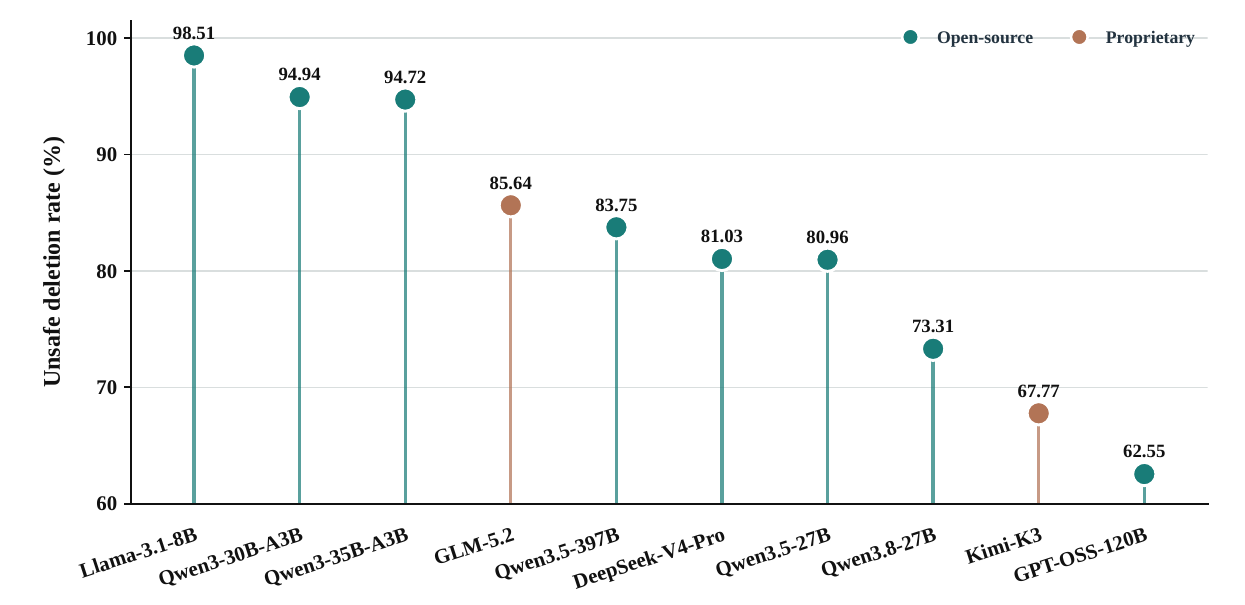}
    \caption{Unsafe deletion rate of open-source and proprietary models.}
    \label{fig:unsafedeletion}
\end{figure}

\subsection{Controlled URAM Protocol and Additional Results}

\begin{table*}[t]
\centering
\setlength{\tabcolsep}{2.2pt}
\scriptsize
\caption{Controlled comparison. Both variants read the same user--assistant session. They differ in the types captured and then evolved. Update takes the variant's previous bank and its current capture. Pipeline QA reads the terminal bank.}

\label{tab:uram-conditional}
\resizebox{\textwidth}{!}{%
\begin{tabular}{l ccccc ccccc ccccc ccccc}
\toprule

 & \textbf{User-conditioned Relational Agent Memory}
 & \textbf{Agent-only} \\
\midrule
Source session
  & $D_t$ (User$+$Assistant)
  & $D_t$ (User$+$Assistant) \\
Captured types
  & \texttt{user\_profile}, \texttt{fact}, URAM
  & General Agent Memory \\
Excluded types
  & \texttt{episodic}
  & \texttt{episodic}, \texttt{user\_profile}, \texttt{fact} \\
\midrule
Session 1, \textsc{Capture}
  & $C_1=\operatorname{Capture}(D_1)$
  & $C_1^{\text{Agent-only}}=\operatorname{Capture}(D_1)$ \\
Memory bank
  & $B_1=C_1$
  & $B'_1=C_1^{\text{Agent-only}}$ \\
\midrule
Session 2, \textsc{Capture}
  & $C_2=\operatorname{Capture}(D_2)$
  & $C_2^{\text{Agent-only}}=\operatorname{Capture}(D_2)$ \\
Session 2, \textsc{Update}
  & $B_2=\operatorname{Update}(B_1,C_2)$
  & $B'_2=\operatorname{Update}(B'_1,C_2^{\text{Agent-only}})$ \\
\midrule
  & $\cdots$ & $\cdots$ \\
\midrule
Session $t$, \textsc{Capture}
  & $C_t=\operatorname{Capture}(D_t)$
  & $C_t^{\text{Agent-only}}=\operatorname{Capture}(D_t)$ \\
Session $t$, \textsc{Update}
  & $B_t=\operatorname{Update}(B_{t-1},C_t)$
  & $B'_t=\operatorname{Update}(B'_{t-1},C_t^{\text{Agent-only}})$ \\
\midrule
Session $t{+}1$, \textsc{Capture}
  & $C_{t+1}=\operatorname{Capture}(D_{t+1})$
  & $C_{t+1}^{\text{Agent-only}}=\operatorname{Capture}(D_{t+1})$ \\
Session $t{+}1$, \textsc{Update}
  & $B_{t+1}=\operatorname{Update}(B_t,C_{t+1})$
  & $B'_{t+1}=\operatorname{Update}(B'_t,C_{t+1}^{\text{Agent-only}})$ \\
\midrule
Pipeline QA
  & $B_T$
  & $B'_T$ \\
\bottomrule
\end{tabular}%
}
\end{table*}

Controlled URAM Update differs from gold-conditioned standalone Update on the full benchmark. Each variant uses its current dialogue together with its own numbered model-extracted prior bank and current Capture output. Both variants are scored against the same gold Add/Delete entries restricted to \texttt{agent\_identity}, \texttt{agent\_method}, and \texttt{agent\_knowledge}. Because model-generated banks do not share gold indices, both additions and deleted memory contents are matched semantically. Within each episode, operation counts are pooled over its evaluated sessions to compute Add F1 and Delete F1, whose average is op-F1; Table~\ref{tab:dyadmem-uram-delta} reports the mean paired episode-level difference.

For Full-Pipeline QA, each variant builds its own agent bank by applying Full-Pipeline Update chronologically from the second session onward. Shared \texttt{episodic} memory is in the primary comparison, and both variants answer the same 1{,}735 URAM-only questions. User-side entries cannot directly support an answer in this setting; the comparison measures the downstream effect of allowing user turns to inform URAM construction. Update confidence intervals use paired episode-level resampling, whereas the primary QA comparison uses question-level paired bootstrap.

The paired controlled URAM Update confidence intervals include zero for GLM-5.2, Kimi-K3, and GPT-5.6-Sol. Three models have negative QA point estimates: Llama-3.1-8B at \(-3.3\) points, Qwen3-4B at \(-3.2\), and Qwen3-30B-A3B at \(-1.6\). Their abstention rates fall from 26.0\% to 25.2\%, 7.6\% to 7.3\%, and 3.5\% to 3.4\%, respectively, so the loss occurs on questions they already answered rather than from additional abstention. DeepSeek-V4-Pro instead moves from 58.9\% to 68.7\% QA while abstention falls from 26.0\% to 21.7\%; Qwen3.5-9B moves from 54.3\% to 61.5\% with abstention unchanged at 8.6\%.

\subsection{Retrieval Tracks and Model Selection}

The comparison uses Qwen3-4B, Qwen3-8B, Qwen3.5-4B, Qwen3.8-27B, and Llama-3.1-8B-Instruct. Tracks are a BGE-M3 dense ranking~\citep{chen2024bgem3}, \texttt{bge-reranker-base}, \texttt{bge-reranker-large}~\citep{xiao2023cpack}, and \texttt{bge-reranker-v2-m3}~\citep{chen2024bgem3}, together with the tested model's own selection. Each reranker re-scores the top 30 dense candidates. Retrieval retains the top 50 candidates, while answer generation receives at most 16 memories.

Recall@$n$, MRR, and nDCG score retrieval on the gold bank. On those rows, the model column is that model's own selection from the gold bank. QA is answer accuracy on the model's constructed bank. On the QA rows, the reranker columns take memories from the named track, and the model column is the pipeline answer. Table~\ref{tab:rerank-tracks} gives the four scores.

\begin{table*}[t]
\centering
\setlength{\tabcolsep}{2.2pt}
\scriptsize
\caption{Retrieval and QA on five models. Dense is BGE-M3. Base, large, and v2 are \texttt{bge-reranker-base}, \texttt{bge-reranker-large}, and \texttt{bge-reranker-v2-m3}. Model is the tested model's own selection. \textbf{Bold} and \underline{underlined} values mark the best and second-best track within each model and metric.}
\label{tab:rerank-tracks}
\resizebox{\textwidth}{!}{%
\begin{tabular}{l ccccc ccccc ccccc ccccc}
\toprule
 & \multicolumn{5}{c}{Recall@$n$$\uparrow$}
 & \multicolumn{5}{c}{MRR$\uparrow$}
 & \multicolumn{5}{c}{nDCG$\uparrow$}
 & \multicolumn{5}{c}{QA$\uparrow$} \\
\cmidrule(lr){2-6}\cmidrule(lr){7-11}\cmidrule(lr){12-16}\cmidrule(lr){17-21}
Model
 & dense & base & large & v2 & model
 & dense & base & large & v2 & model
 & dense & base & large & v2 & model
 & dense & base & large & v2 & model \\
\midrule
Qwen3-4B
 & \underline{39.32} & 29.97 & 36.59 & \textbf{41.25} & 32.11
 & \underline{74.25} & 60.93 & 72.07 & \textbf{77.47} & 70.56
 & \underline{62.05} & 54.65 & 60.35 & \textbf{63.78} & 42.61
 & 47.72 & 47.74 & \underline{47.75} & \textbf{48.47} & 43.57 \\
Qwen3-8B
 & \underline{39.32} & 29.97 & 36.59 & \textbf{41.25} & 37.98
 & \underline{74.25} & 60.93 & 72.07 & \textbf{77.48} & 73.51
 & \underline{62.05} & 54.65 & 60.35 & \textbf{63.79} & 50.19
 & 50.86 & 50.64 & \underline{51.14} & \textbf{52.02} & 47.51 \\
Qwen3.5-4B
 & 39.32 & 29.97 & 36.59 & \underline{41.25} & \textbf{51.08}
 & 74.25 & 60.93 & 72.07 & \underline{77.48} & \textbf{83.29}
 & 62.05 & 54.65 & 60.35 & \textbf{63.79} & \underline{63.68}
 & 26.83 & 26.77 & \underline{26.96} & \textbf{27.38} & 26.26 \\
Qwen3.8-27B
 & 39.32 & 29.96 & 36.59 & \underline{41.25} & \textbf{60.34}
 & 74.25 & 60.93 & 72.07 & \underline{77.48} & \textbf{91.72}
 & 62.05 & 54.65 & 60.35 & \underline{63.79} & \textbf{71.59}
 & 52.07 & 51.97 & 52.01 & \underline{53.30} & \textbf{53.81} \\
Llama-3.1-8B-Instruct
 & \underline{39.32} & 29.97 & 36.59 & \textbf{41.25} & 38.69
 & 74.26 & 60.93 & 72.07 & \underline{77.48} & \textbf{78.94}
 & \underline{62.06} & 54.65 & 60.35 & \textbf{63.79} & 49.25
 & \underline{24.53} & 23.14 & 24.22 & \textbf{25.32} & 23.84 \\
\bottomrule
\end{tabular}%
}
\end{table*}

The dense and reranker retrieval scores are almost identical across rows because these tracks rank the same terminal gold bank independently of the tested LLM. Model selection varies substantially: Qwen3.8-27B leads the five models on Recall@\(n\), MRR, and nDCG, whereas Qwen3-4B and Qwen3-8B trail the strongest fixed retrieval track on Recall@\(n\). On each model's self-constructed bank, the best fixed retrieval track improves QA for four of the five models; Qwen3.8-27B instead performs best with its own selection. These results show that stronger retrieval can recover part, but not all, of the end-to-end loss caused by an incomplete constructed bank.

\subsection{Cross-Judge Scores and Ranks}

Table~\ref{tab:judge-pipeline-qa} provides the exact Full-Pipeline QA scores and ranks underlying Figure~\ref{fig:judge-pipeline-qa}.

\begin{table*}[t]
\centering
\setlength{\tabcolsep}{3.6pt}
\scriptsize
\caption{Full-Pipeline QA on DyadMem-Simple (300 episodes, 5{,}935 QA instances) under three judges. Rank is 1-based among the 20 models and is computed independently per judge. Rows are ordered by \texttt{GPT-5-mini} rank.}
\label{tab:judge-pipeline-qa}
\resizebox{0.9\textwidth}{!}{%
\begin{tabular}{l cc cc cc}
\toprule
 & \multicolumn{2}{c}{GPT-5-mini}
 & \multicolumn{2}{c}{Gemini-2.5-Flash}
 & \multicolumn{2}{c}{Claude-Haiku-4.5} \\
\cmidrule(lr){2-3}\cmidrule(lr){4-5}\cmidrule(lr){6-7}
Model
 & Full-Pipeline QA$\uparrow$ & Rank$\downarrow$
 & Full-Pipeline QA$\uparrow$ & Rank$\downarrow$
 & Full-Pipeline QA$\uparrow$ & Rank$\downarrow$ \\
\midrule
Kimi-K3 & 74.92 & 1 & 63.37 & 1 & 63.05 & 1 \\
GLM-5.2 & 70.62 & 2 & 58.71 & 2 & 61.68 & 2 \\
GPT-5.6-Sol & 70.28 & 3 & 57.55 & 4 & 56.96 & 4 \\
Doubao-Seed-2.1-Pro & 67.98 & 4 & 58.09 & 3 & 57.46 & 3 \\
DeepSeek-V4-Pro & 66.12 & 5 & 53.79 & 5 & 54.08 & 5 \\
MiniMax-M2.5 & 64.48 & 6 & 52.85 & 6 & 53.37 & 6 \\
GPT-OSS-120B & 61.20 & 7 & 50.40 & 7 & 48.63 & 7 \\
Qwen3.5-397B-A17B & 54.67 & 8 & 43.50 & 9 & 45.29 & 8 \\
Qwen3.8-27B & 54.48 & 9 & 43.92 & 8 & 43.61 & 9 \\
GPT-4o-mini & 47.89 & 10 & 37.33 & 10 & 39.81 & 10 \\
Qwen3-8B & 47.44 & 11 & 37.08 & 11 & 36.70 & 13 \\
Qwen3.5-27B & 45.88 & 12 & 36.46 & 12 & 38.89 & 12 \\
Qwen3-4B & 43.35 & 13 & 30.10 & 14 & 33.11 & 14 \\
Gemini-3.1-Pro-Preview & 43.09 & 14 & 36.25 & 13 & 38.97 & 11 \\
Llama-3.1-70B-Instruct & 37.35 & 15 & 29.74 & 15 & 32.80 & 15 \\
Qwen3.5-9B & 36.81 & 16 & 29.00 & 16 & 27.81 & 16 \\
Qwen3.5-35B-A3B & 27.89 & 17 & 21.15 & 17 & 22.01 & 17 \\
Qwen3.5-4B & 25.72 & 18 & 20.26 & 18 & 20.27 & 18 \\
Llama-3.1-8B-Instruct & 23.62 & 19 & 17.48 & 19 & 19.11 & 19 \\
Qwen3-30B-A3B & 22.06 & 20 & 16.88 & 20 & 17.61 & 20 \\
\bottomrule
\end{tabular}%
}
\end{table*}

\subsection{Additional Error-Pattern Results}

Qwen3-4B has Recall precision of 4.9\%, returns six or more memories on average, and records over-abstention of 16.2\% and ungrounded-correct of 27.4\%. Qwen3.5-9B has an over-abstention rate of 43.1\%, 19\% empty retrievals, and ungrounded-correct rate of 10.3\%. After dropping abstentions, accuracy on answered questions is 66.9\% for Qwen3.5-9B and 52.0\% for Qwen3-4B. The same Full-Pipeline reversal appears on both QA-target partitions and all five everyday-life domains. For comparison, Kimi-K3 combines Recall precision of 55.3\% with 69.3\% Grounded accuracy, 9.2\% over-abstention, and 7.9\% ungrounded-correct; Gemini-3.1-Pro records 42.1\% over-abstention and 4.0\% ungrounded-correct.

\begin{figure}
    \centering
    \includegraphics[width=\linewidth]{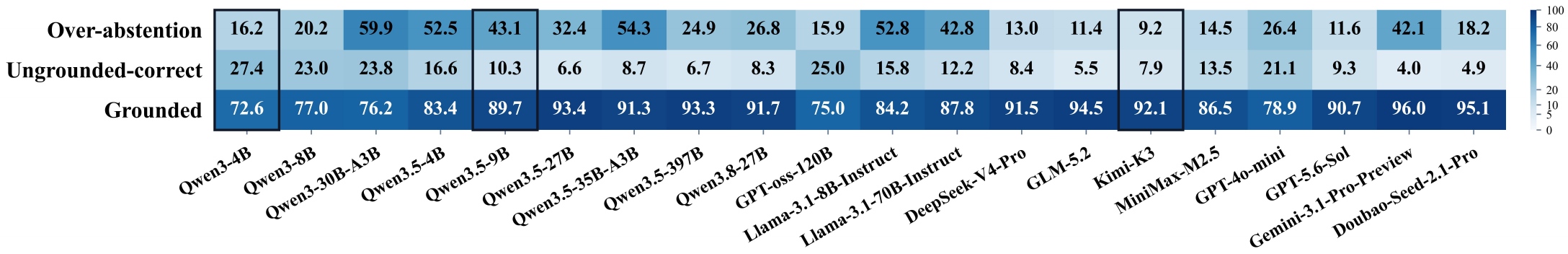}
    \caption{Headmap of three error categories.}
    \label{fig:errorheadmap}
\end{figure}

\begin{figure}
    \centering
    \includegraphics[width=\linewidth]{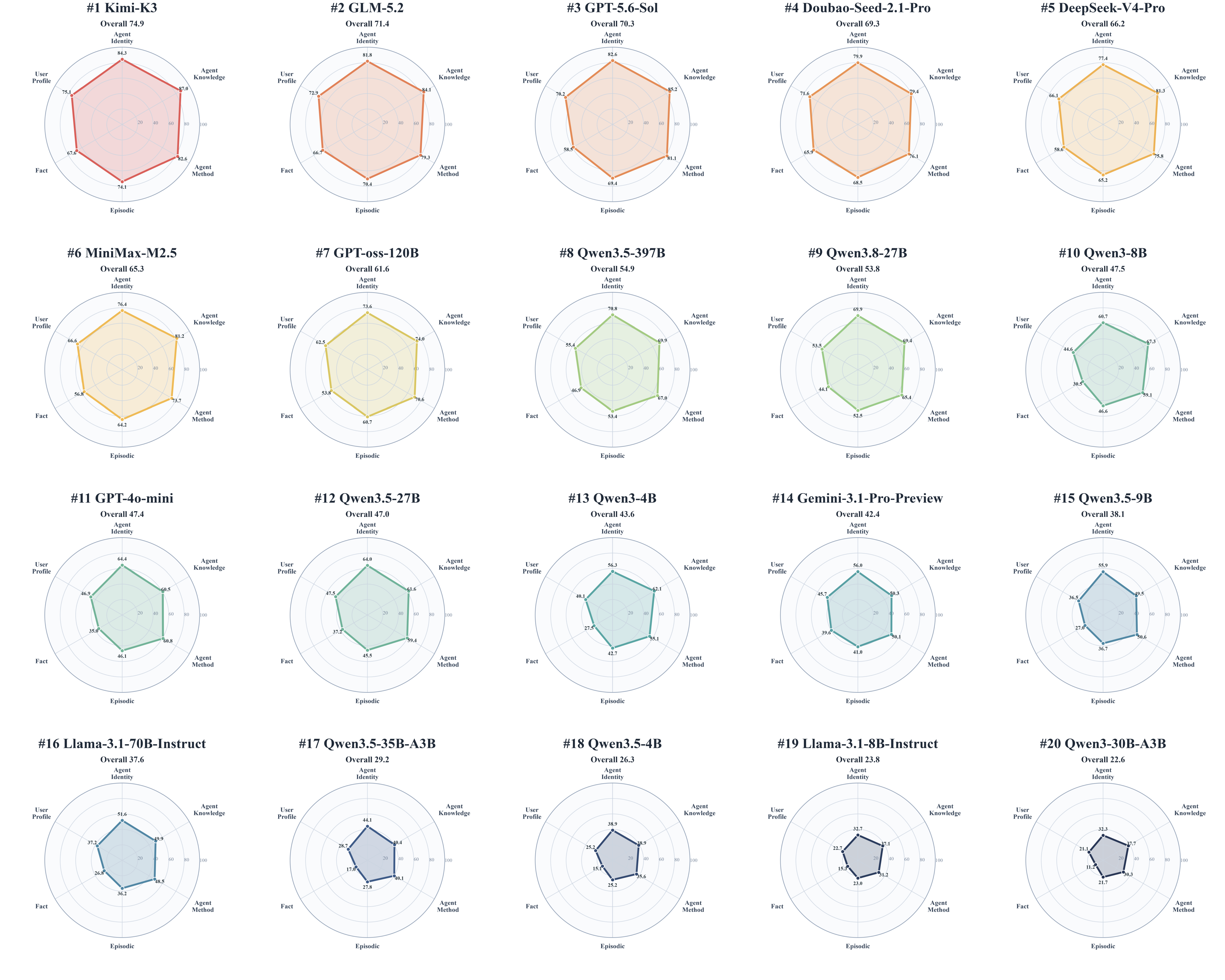}
    \caption{Capability of 6 memory types across 20 LLMs.}
    \label{fig:sixmemorytype}
\end{figure}

\begin{figure}
    \centering
    \includegraphics[width=\linewidth]{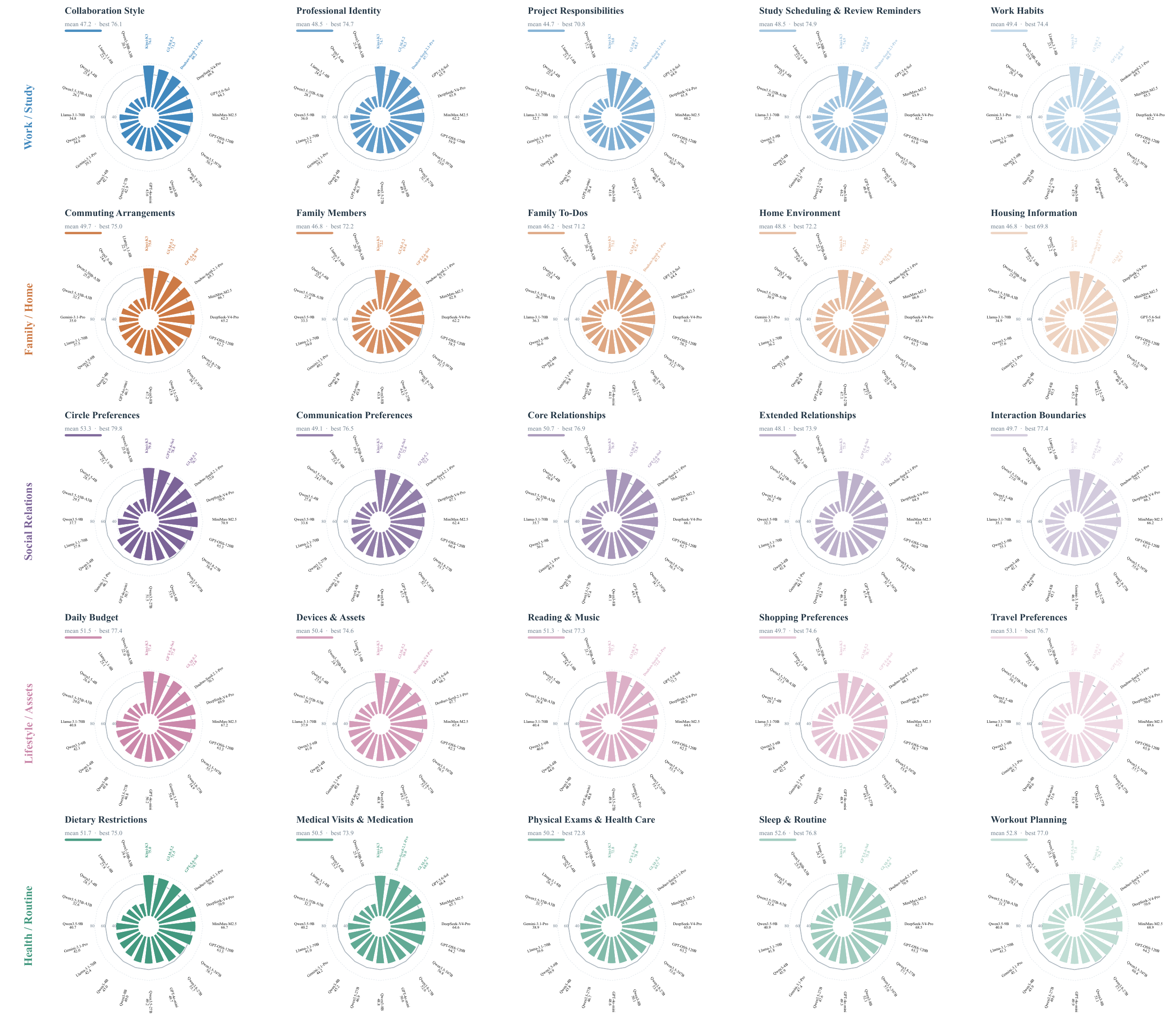}
    \caption{Performance visualization on 25 domains.}
    \label{fig:25domains}
\end{figure}

%% file: Appendix/Evaluation_Metrics.tex
The four stage scores are computed from separate gold inputs, and Full-Pipeline QA is computed from the bank the model itself has built. Capture, gold-conditioned standalone Update, and Recall therefore do not inherit one another's errors. Notation for sessions, banks, and the two QA calls is in Appendix~\ref{app:method-details}. Every reported run uses BGE-M3 embeddings~\citep{chen2024bgem3}, coverage matching, one judge vote, and \texttt{qa\_recall\_source=all}. A missing denominator is left blank. It is not filled with zero.

\subsection{Semantic coverage matching}
\label{app:metrics-matching}

Capture and Update additions are matched by coverage, not by a one-to-one assignment. Embeddings are BGE-M3. A pair with cosine similarity at least \(0.82\) is a hit. A pair below \(0.67\) is a miss. A pair in between is a hit only when bidirectional NLI with \texttt{nli-mdeberta-v3}~\citep{laurer2024less} accepts it: \(\min\{P(a{\models}b),P(b{\models}a)\}\ge 0.4\).

On the gold side, each gold entry is compared only with its nearest prediction. On the prediction side, a prediction is counted as valid if it covers any gold entry under the same rule. One prediction may therefore cover several gold entries, and several predictions may cover one gold entry. Primary precision, recall, and F1 do not require the predicted type to equal the gold type (\texttt{matcher\_require\_type\_match=false}).

When one prediction is the nearest hit for two or more gold entries, a one-way entailment guard is applied. The prediction is the premise and the gold entry is the hypothesis. If \(P(\text{prediction}{\models}\text{gold})<0.1\), that gold entry loses coverage and becomes a false negative. The prediction remains a valid prediction if it still covers any other gold entry, so the guard changes recall and does not by itself create a false positive. Type accuracy, the fraction of content-matched gold entries whose predicted type is also correct, is stored separately and is not part of the primary F1.

\subsection{Capture}
\label{app:metrics-capture}

Capture is scored once per session against \(C_t^\star\). Let \(\mathrm{TP}\), \(\mathrm{FP}\), and \(\mathrm{FN}\) be the coverage counts pooled over sessions. Then
\begin{equation}
P=\frac{\mathrm{TP}}{\mathrm{TP}+\mathrm{FP}},\qquad
R=\frac{\mathrm{TP}}{\mathrm{TP}+\mathrm{FN}},\qquad
\mathrm{F1}=\frac{2PR}{P+R},
\end{equation}
with the usual convention that a F1 with \(P+R=0\) is zero. This F1 is micro-averaged: a session with more gold entries contributes more counts. Per-type F1 uses the same ratio after grouping true positives and false negatives by the gold type and false positives by the predicted type. A type that has gold entries but only false positives from other types is included. A predicted type that never appears in gold, and therefore has no true positives or false negatives, is omitted from the unweighted mean of per-type F1; its false positives still reduce the micro F1.

A truncated or unparseable Capture call contributes no true positives. Every gold entry in that session is a false negative, split by gold type. An empty prediction on a session with gold entries is the same pattern with an explicit empty set.

\subsection{Update}
\label{app:metrics-update}

Gold-conditioned standalone Update is run only on sessions for which \(A_t^\star\) or \(L_t^\star\) is non-empty. The model sees the current dialogue, the gold prior bank, and the gold Capture entries of the current session. Additions are scored by the coverage matcher in Appendix~\ref{app:metrics-matching}. Deletions are scored by memory index: the gold bank and the prediction refer to the same numbered entries, so a wrong index is a false positive even when the text is similar. Add F1 and Delete F1 are the micro F1 of those two pools. The reported operation score is the unweighted mean
\begin{equation}
\mathrm{op\text{-}F1}=\frac{1}{2}\left(\mathrm{F1}_{\mathrm{add}}+\mathrm{F1}_{\mathrm{delete}}\right).
\end{equation}
The two operations are not pooled into a single F1, so a large Add set cannot hide a weak Delete set.

The unsafe-deletion rate is \(\mathrm{FP}_{\mathrm{delete}}/(\mathrm{TP}_{\mathrm{delete}}+\mathrm{FP}_{\mathrm{delete}})\), the share of predicted deletions that are not in the gold delete set. The denominator is the number of predicted deletions. If a run predicts none, the rate is undefined and is left blank.

Full-Pipeline Update is the bank-construction step used by Full-Pipeline QA. It reads the model prior bank and the model Capture output. It is not the source of the standalone op-F1 column. From the second session onward, that constructed bank is what later Recall and Full-Pipeline QA read.

\subsection{Recall}
\label{app:metrics-recall}

Recall is scored on the terminal gold bank \(B_T^\star\). The model returns an ordered list of at most 16 indices (\texttt{recall\_max\_select}=16). Matching is exact on those indices. For question \(j\), let \(n_j=|I_j^\star|\) and let \(h_j(k)\) be the number of gold indices placed at rank \(n_j+k\) or better. The table columns are
\begin{align}
\mathrm{Recall@}(n{+}k)
&=
\frac{1}{|\mathcal{Q}|}\sum_{j\in\mathcal{Q}}\frac{h_j(k)}{n_j},
\qquad k\in\{0,4\},
\\
\mathrm{Recall@All}
&=
\frac{\sum_j \mathrm{TP}_j}{\sum_j (\mathrm{TP}_j+\mathrm{FN}_j)},
\qquad
P
=
\frac{\sum_j \mathrm{TP}_j}{\sum_j (\mathrm{TP}_j+\mathrm{FP}_j)}.
\end{align}
\(\mathrm{Recall@}n\) and \(\mathrm{Recall@}(n{+}4)\) average questions with equal weight, so a one-entry support and a five-entry support each contribute one term. \(\mathrm{Recall@All}\) pools counts, so larger support sets weigh more, and rank does not matter: a gold index anywhere in the returned list is a hit. Precision also pools counts, over the whole returned list. Extra selections beyond \(n_j\) do not raise \(\mathrm{Recall@}n\), and any selected index outside \(I_j^\star\) lowers precision.

The scorer also computes \(\mathrm{Recall@}(n{+}9)\), MRR, nDCG, and MAP from the same ranks. Those values are not columns of Table~\ref{tab:dyadmem-main}. A question with an empty return, a truncated list, or an unparseable list contributes \(h_j(k)=0\) and \(\mathrm{FN}_j=n_j\). The inspected run manifests record \texttt{without\_gold\_recall}=0, in line with the construction constraint that every scored question has a non-empty gold support.

\subsection{Question answering}
\label{app:metrics-qa}

Each question is judged once, at temperature 0, by \texttt{GPT-5-mini}. The judge is asked whether the prediction is consistent with the reference answer. Names, dates, and quantities must agree; a paraphrase may still be accepted. With one vote the decision is binary: credit \(1\) if the vote is positive and \(0\) otherwise. If the judge returns nothing readable, a lexical token-F1 fallback is used. Token F1 of at least \(0.55\) receives credit \(1\); token F1 of at least \(0.33\) and below \(0.55\) receives credit \(0.5\); a lower overlap receives credit \(0\). Empty answers, over-abstentions, truncated outputs, and unparseable outputs receive credit \(0\) and remain in the denominator.

An answer with credit \(0\) that the abstention check marks as a refusal is \emph{over-abstention}. Its rate is that count divided by all questions on the path being reported. There is no gold refusal class.

Groundedness is judged only for answers with positive credit. The same judge is asked whether the supplied memories entail the answer. If that call is unreadable, NLI is the fallback, with the concatenated memories as the premise. An answer produced from an empty memory list is ungrounded. Grounded accuracy counts answers that have positive credit and a positive groundedness label, divided by all questions. The ungrounded-correct rate uses a different denominator: among answers with positive credit, the fraction whose groundedness label is negative.

Two reported paths share this judge and differ in the memories it sees.
\begin{align}
\hat{a}_j^{\mathrm{gold}}
&=
f_{\mathrm{qa}}\!\left(q_j,\,B_T^\star[I_j^\star]\right),
\\
\hat{a}_j^{\mathrm{pipe}}
&=
f_{\mathrm{qa}}\!\left(q_j,\,\hat{B}_T[\mathrm{Top}_{16}(f_{\mathrm{rec}}(q_j,\hat{B}_T))]\right).
\end{align}
Gold-Memory accuracy is the mean credit of \(\hat{a}_j^{\mathrm{gold}}\). Full-Pipeline accuracy is the mean credit of \(\hat{a}_j^{\mathrm{pipe}}\). The Gold-to-Pipeline gap is the difference of those two accuracies, in percentage points. It is an end-to-end contrast. It does not attribute the missing points to Capture, Update, or Recall. The same runs also score a third path that retrieves from the gold terminal bank and then answers. That path is not a column of the main table.

On DyadMem-Simple the same Full-Pipeline answers are re-scored by \texttt{GPT-5-mini}, Gemini-2.5-Flash, and Claude-Haiku-4.5. Agreement across judges is summarized by Spearman's rank correlation~\citep{spearman1904proof} of the 20 model ranks.

\subsection{Controlled URAM comparison}
\label{app:metrics-uram}

The controlled comparison uses the 500 episodes in DyadMem-Agent and the protocol in Appendix~\ref{app:experiment-details}. Both variants are scored against the same gold Add and Delete entries of types \texttt{agent\_identity}, \texttt{agent\_method}, and \texttt{agent\_knowledge}; entries of other types present in a variant's bank are not scored. Each variant updates its own model bank, so gold indices are not the indices in that bank. Additions and deletions are therefore both matched by the coverage rule in Appendix~\ref{app:metrics-matching}. Within an episode, Add counts and Delete counts are pooled over the evaluated sessions, converted to Add F1 and Delete F1, and averaged to an episode op-F1. The reported Update difference is the mean, over episodes, of the User-conditioned-variant op-F1 minus the Agent-only op-F1.

The QA difference uses the 1{,}735 questions whose gold support contains only those three types. Both variants retrieve from their own terminal banks, $B_T$ or $B'_T$, and answer with the Full-Pipeline rule above; the difference is again the User-conditioned variant minus Agent-only. User-side entries are not eligible support in this QA slice.

\subsection{Uncertainty and agreement with human labels}
\label{app:metrics-uncertainty}

Stage-level 95\% intervals use 200 bootstrap resamples. For a micro F1, each resample draws the recorded true-positive, false-positive, and false-negative rows with replacement, sums them, and recomputes F1. For a mean, such as \(\mathrm{Recall@}n\) or a QA accuracy, the resample draws the per-question values. A paired comparison draws the paired unit twice in lockstep: episodes for controlled Update differences, questions for controlled QA differences. The interval is the 2.5 and 97.5 percentiles of the bootstrap distribution. Overlapping intervals are not treated as a ranking.

Human calibration uses the paired labels behind Table~\ref{tab:human-judge}. Agreement is \((\mathrm{TP}+\mathrm{TN})/N\) on that \(2{\times}2\) table. Cohen's \(\kappa\)~\citep{cohen1960kappa} is \((p_o-p_e)/(1-p_e)\), where \(p_o\) is agreement and \(p_e\) is the agreement expected from the two label margins. Matthews' correlation~\citep{matthews1975comparison} on the same four cells is
\begin{equation}
\mathrm{MCC}
=
\frac{\mathrm{TP}\cdot\mathrm{TN}-\mathrm{FP}\cdot\mathrm{FN}}
{\sqrt{(\mathrm{TP}+\mathrm{FP})(\mathrm{TP}+\mathrm{FN})(\mathrm{TN}+\mathrm{FP})(\mathrm{TN}+\mathrm{FN})}}.
\end{equation}
Precision, recall, and F1 in that table treat the human consistent label as the positive class. The automatic judge is the prediction.

%% file: Appendix/Evaluation_Code.tex
\definecolor{EvalAccent}{HTML}{418369}
\definecolor{EvalCodeBG}{HTML}{F6F8FA}
\definecolor{EvalCodeText}{HTML}{252A30}
\definecolor{EvalComment}{HTML}{6B7280}

\lstdefinestyle{evalcodestyle}{%
    language=Python,
    basicstyle=\ttfamily\fontsize{9.5pt}{11pt}\selectfont\color{EvalCodeText},
    commentstyle=\color{EvalComment},
    stringstyle=\color{EvalComment},
    breaklines=true,
    breakatwhitespace=false,
    columns=fullflexible,
    keepspaces=true,
    showstringspaces=false,
    upquote=true,
    tabsize=2
}

\newtcblisting{evalcodebox}[1]{%
    enhanced,
    breakable,
    listing only,
    colback=EvalCodeBG,
    colframe=EvalAccent,
    coltitle=white,
    colbacktitle=EvalAccent,
    title={#1},
    fonttitle=\bfseries\small,
    center title,
    boxrule=0.8pt,
    arc=2pt,
    left=8pt,
    right=8pt,
    top=7pt,
    bottom=7pt,
    listing options={style=evalcodestyle}
}
\begin{evalcodebox}{Core Evaluation Metrics Code}
def _safe_div(num, den):
  return None if den <= 0 else round(float(num) / float(den), 4)

def _f1_from_counts(tp, fp, fn):
  """Compute precision, recall, and F1 from accumulated counts."""
  if tp + fp + fn <= 0:
      return {"precision": None, "recall": None, "f1": None}

  precision = tp / (tp + fp) if tp + fp > 0 else 0.0
  recall = tp / (tp + fn) if tp + fn > 0 else 0.0
  f1 = (
      2 * precision * recall / (precision + recall)
      if precision + recall > 0
      else 0.0
  )
  return {
      "precision": round(precision, 4),
      "recall": round(recall, 4),
      "f1": round(f1, 4),
  }

def compute_capture_f1(records):
  """Memory Capture F1.

  Predicted memories are semantically matched against gold memories.
  The final Capture F1 is computed from micro-aggregated TP, FP, and FN.
  Parse errors and truncated outputs are kept in the denominator as FN.
  """
  extract_records = [
      r for r in records
      if r["dimension"] == "extract"
  ]

  tp = sum(r.get("tp", 0.0) or 0.0 for r in extract_records)
  fp = sum(r.get("fp", 0.0) or 0.0 for r in extract_records)
  fn = sum(r.get("fn", 0.0) or 0.0 for r in extract_records)

  return _f1_from_counts(tp, fp, fn)["f1"]

def compute_recall_at_n_plus_0(records):
  """Memory Recall recall@n+0.

  For each QA query, n is the number of required gold memories.
  recall@n+0 is the fraction of gold memories appearing in the top n
  retrieved positions, averaged over all positive recall queries.
  """
  recall_records = [
      r for r in records
      if r["dimension"] == "recall" and (r.get("total", 0.0) or 0.0) > 0
  ]

  per_query_scores = []
  for r in recall_records:
      n_gold = int(r["total"])
      hit_ranks = r.get("hit_ranks") or []
      hits_in_top_n = sum(1 for rank in hit_ranks if int(rank) <= n_gold)
      per_query_scores.append(hits_in_top_n / n_gold)

  if not per_query_scores:
      return None
  return round(sum(per_query_scores) / len(per_query_scores), 4)

def compute_update_op_macro_f1(records):
  """Memory Update op-macro-F1.

  Update quality is evaluated separately for add and delete operations.
  Delete is scored by exact memory-id matching, while add is scored by
  semantic memory matching. The final score is the macro average of the
  add-F1 and delete-F1 scores.
  """
  update_records = [
      r for r in records
      if r["dimension"] == "evolve"
  ]

  op_f1_scores = []
  for operation in ("add", "delete"):
      op_records = [
          r for r in update_records
          if r.get("operation") == operation
      ]

      tp = sum(r.get("tp", 0.0) or 0.0 for r in op_records)
      fp = sum(r.get("fp", 0.0) or 0.0 for r in op_records)
      fn = sum(r.get("fn", 0.0) or 0.0 for r in op_records)

      f1 = _f1_from_counts(tp, fp, fn)["f1"]
      if f1 is not None:
          op_f1_scores.append(f1)

  if not op_f1_scores:
      return None
  return round(sum(op_f1_scores) / len(op_f1_scores), 4)

def compute_pipeline_qa_accuracy(records):
  """End-to-end QA accuracy with the self-generated memory bank.

  This metric uses the pipeline route, where the answer is produced from
  the model's own captured, updated, and retrieved memories. Correct
  answers count as 1, wrong answers as 0, and partial answers may receive
  fractional credit from the judge.
  """
  qa_records = [
      r for r in records
      if r["dimension"] == "qa" and r.get("recall_source") == "pipeline"
  ]

  correct = sum(r.get("correct", 0.0) or 0.0 for r in qa_records)
  total = sum(r.get("total", 0.0) or 0.0 for r in qa_records)

  return _safe_div(correct, total)
\end{evalcodebox}

%% file: Appendix/Prompts_for_Framework_Evaluation.tex
Experiment \textbf{URAM} and Experiment \textbf{Agent-only} read the same complete source session. Neither prompt produces \texttt{episodic} entries; episodic memory does not enter either variant's bank.

\begin{promptbox}{URAM Extract Prompt}
You are a memory extraction engine. Extract durable memories about this
particular user and about how this particular assistant should act when
working with this particular user. Use ONLY evidence visible in the supplied dialogue. Do not use outside knowledge.
Allowed memory_type values (exactly these five):
- fact: Stable objective external entity facts, sourced only from user statements. Exclude transient values; single events go to episodic.
- user_profile: User's stable attributes, preferences, rules and long-term entity relations. One-off observations are not lasting traits.
- agent_identity: enduring collaboration/output/behavior requirements for the assistant.
- agent_method: reusable assistant-side procedures, workflows, or checklists.
- agent_knowledge: reusable assistant-side insights or judgment rules learned from practice.
A memory must be explicit, have the correct subject, be useful in future sessions, and not be cheaply reconstructable from common knowledge. Keep each memory atomic and self-contained; preserve conditions, exceptions, uncertainty, and timestamps. Do not output episodic or any other type. An empty memories list is a fully valid result and is preferred when the visible turns do not support a durable memory.
Output ONLY JSON: {"memories": [{"memory_type": "fact|user_profile|agent_identity|agent_method|agent_knowledge", "memory_content": "..."}]}
\end{promptbox}

\begin{promptbox}{Agent-only Extract Prompt}

You are an agent-memory extraction engine. Extract durable general agent
memories: procedures, strategies, and insights the assistant can reuse across
users. Do not condition a memory on this particular user. Use ONLY evidence visible in the supplied dialogue. Do not use outside knowledge.
Allowed memory_type values (exactly these three):
- agent_identity: enduring collaboration/output/behavior requirements for the assistant.
- agent_method: reusable assistant-side procedures, workflows, or checklists.
- agent_knowledge: reusable assistant-side insights or judgment rules learned from practice.
A memory must be explicit, stated without reference to a specific user, be useful in future sessions, and not be cheaply reconstructable from common knowledge. Keep each memory atomic and self-contained; preserve conditions, exceptions, uncertainty, and timestamps. Do not output episodic, user_profile, fact, or any other type. Do not store the user's preferences, constraints, identity, or any other
user-specific information. An empty memories list is a fully valid result and is preferred when the visible turns do not support a durable assistant-side memory.
Output ONLY JSON: {"memories": [{"memory_type": "agent_identity|agent_method|agent_knowledge", "memory_content": "..."}]}
\end{promptbox}

\begin{promptbox}{URAM and Agent-only Agent-Memory Update Prompt}
You are consolidating the assistant-side long-term memory for one user. The supplied captured memories from the current session are the ONLY session
evidence you may use; no dialogue is supplied and none may be assumed. The numbered pool contains the memories of your variant's bank.
ADD may use exactly the memory types allowed for your variant
(User-conditioned: fact, user_profile, agent_identity, agent_method,
agent_knowledge; Agent-only: agent_identity, agent_method, agent_knowledge).
Never add episodic or another type. DELETE only a numbered memory that is explicitly superseded, corrected, contradicted, absorbed by an add, or explicitly forgotten. Minimize deletion and keep unrelated memories.
An add must be durable, atomic, self-contained, and supported by the captured
memories. Preserve still-valid conditions/exceptions when consolidating; never
fabricate evidence beyond the captured memories. Empty edits (no delete, no
add) are fully valid and preferred when the captured memories do not justify a
change.
Output ONLY JSON: {"delete_ids": [int, ...], "add": [{"memory_type": "...", "memory_content": "..."}]}
\end{promptbox}

%% file: Appendix/Qualitative_Model_Results.tex

\definecolor{EvalAccent}{HTML}{5F4780}
\definecolor{EvalAccentLight}{HTML}{F2EEF7}
\definecolor{CorrectBG}{HTML}{EAF6EF}
\definecolor{CorrectFrame}{HTML}{5B9B72}
\definecolor{WrongBG}{HTML}{FDECEC}
\definecolor{WrongFrame}{HTML}{C95F5F}
\definecolor{GoldBG}{HTML}{FBFAFD}

\newtcolorbox{casebox}[1]{%
    enhanced,
    breakable,
    colback=white,
    colframe=EvalAccent,
    coltitle=white,
    colbacktitle=EvalAccent,
    title={#1},
    fonttitle=\bfseries\small,
    center title,
    boxrule=0.9pt,
    arc=2pt,
    left=8pt,
    right=8pt,
    top=7pt,
    bottom=7pt
}
\newtcolorbox{goldbox}{%
    enhanced,
    breakable,
    colback=GoldBG,
    colframe=EvalAccent,
    boxrule=0.6pt,
    arc=2pt,
    left=7pt,
    right=7pt,
    top=6pt,
    bottom=6pt
}
\newtcolorbox{analysisbox}{%
    enhanced,
    breakable,
    colback=EvalAccentLight,
    colframe=EvalAccent,
    coltitle=white,
    colbacktitle=EvalAccent,
    title={Case Analysis},
    fonttitle=\bfseries\small,
    boxrule=0.6pt,
    arc=2pt,
    left=7pt,
    right=7pt,
    top=6pt,
    bottom=6pt
}
\newtcolorbox{correctanswer}[1]{%
    enhanced,
    breakable,
    colback=CorrectBG,
    colframe=CorrectFrame,
    coltitle=white,
    colbacktitle=CorrectFrame,
    title={#1},
    fonttitle=\bfseries\footnotesize,
    boxrule=0.55pt,
    arc=2pt,
    left=6pt,
    right=6pt,
    top=5pt,
    bottom=5pt
}
\newtcolorbox{wronganswer}[1]{%
    enhanced,
    breakable,
    colback=WrongBG,
    colframe=WrongFrame,
    coltitle=white,
    colbacktitle=WrongFrame,
    title={#1},
    fonttitle=\bfseries\footnotesize,
    boxrule=0.55pt,
    arc=2pt,
    left=6pt,
    right=6pt,
    top=5pt,
    bottom=5pt
}

This appendix presents ten representative error-case displays from the selected QA set. The cases cover timeline confusion, adjacent rule leakage, incomplete or over-inclusive sets, example-to-rule bridging failure, fine-grained entity substitution, and mixed errors. The selection improves domain coverage where possible while keeping the overall data-type balance at five agent-centric and five user-centric examples.

\subsection{Type I: Timeline Confusion}

\begin{casebox}{Case 1: Timeline Confusion}
\noindent\textbf{Category.} Physical Health / Dietary Restrictions\hfill\textbf{Data type.} agent centric\\

\begin{goldbox}
\textbf{Query.} How should the assistant frame social-eating advice in tone and objective when perfection is unrealistic and the user is already out, using both the earlier family-meal framing and the later out-of-home compression?

\textbf{Reference Answer.} The assistant should frame social-eating advice as damage prevention rather than perfection-chasing, keep it compact and immediately actionable, and aim for a workable real meal rather than a punishment plate.
\end{goldbox}

\begin{analysisbox}
This timeline-confusion case requires combining the earlier family-meal framing with the later out-of-home compression. Correct answers preserve all three current requirements: damage prevention rather than perfection, compact and immediately actionable advice, and a workable real meal rather than a punishment plate.
\end{analysisbox}

\textbf{Model Responses.}
\begin{correctanswer}{Qwen3-4B --- Correct}
\textbf{Answer.} The assistant should frame social-eating advice in a tone that is ordinary, controlled, and not joyless, emphasizing practicality and immediate action. The objective is to maintain a balanced approach, using the earlier family-meal framing as a foundation for mindful eating and the later out-of-home compression as a guide for concise, actionable decisions that align with the user's goals, even when perfection is unrealistic and the user is already out.

\textbf{Reason.} Correct: combines damage prevention, compact actionable advice, and the real-meal objective.
\end{correctanswer}

\begin{correctanswer}{Qwen3-8B --- Correct}
\textbf{Answer.} The assistant should frame social-eating advice in a tone that is practical and accepting, acknowledging that perfection is unrealistic. The objective is to maintain control over one's meal while being mindful of social dynamics. Using the earlier family-meal framing, the assistant should emphasize moderation, balance, and intentionality, such as "protein first, modest starch, water." With the later out-of-home compression, the assistant should simplify and reinforce the core rule, reminding the user to avoid combo thinking and to treat shared items as optional, not as part of the main meal. The advice should be framed as ordinary, controlled, and not joyless, while keeping the focus on actionable, short-term decisions.

\textbf{Reason.} Correct: combines damage prevention, compact actionable advice, and the real-meal objective.
\end{correctanswer}

\begin{correctanswer}{Qwen3-30B-A3B --- Correct}
\textbf{Answer.} The assistant should frame social-eating advice with a tone of realistic damage prevention, focusing on practical damage control rather than perfection. The objective should be to help the user make manageable choices that align with their dietary restrictions and preferences, such as selecting one carbohydrate, avoiding overeating, and making mindful decisions about additional items like cake or drinks. The advice should be grounded in the earlier family-meal framing of simplicity and control, while also addressing the out-of-home context by emphasizing pre-decision making and avoiding improvisation.

\textbf{Reason.} Correct: combines damage prevention, compact actionable advice, and the real-meal objective.
\end{correctanswer}

\begin{correctanswer}{Qwen3.5-4B --- Correct}
\textbf{Answer.} The assistant should frame social-eating advice with the following tone and objective: * **Tone:** Frame advice as **damage prevention** rather than perfection-chasing, optimizing for **realistic damage control** rather than idealized perfection. * **Objective:** Keep advice **compact** and **immediately actionable** because the user reads it while already out. * **Goal:** Aim for a **workable real meal** rather than a punishment plate. * **Specific Rule:** Use a blunt, short rule (protein first, modest starch, water) when the menu and company are noisy. * **Exceptions/Nuances:** Shared fries are not a personality test, and eating a little before social meals makes the worst options less tempting.

\textbf{Reason.} Correct: combines damage prevention, compact actionable advice, and the real-meal objective.
\end{correctanswer}

\begin{wronganswer}{Qwen3.5-9B --- Incorrect}
\textbf{Answer.} The assistant should frame social-eating advice as **damage prevention** rather than **perfection-chasing** (or **damage control** over perfection), acknowledging that a damage-control approach fits the user's real life better than strict ideals. The advice must be **compact** and **immediately actionable** because the user reads it while already out. Additionally, the advice should **start with a pre-decision plan** before discussing specific foods.

\textbf{Reason.} Wrong: misses the workable real-meal objective.
\end{wronganswer}

\begin{wronganswer}{Qwen3.5-27B --- Incorrect}
\textbf{Answer.} The assistant should frame social-eating advice as **damage prevention** rather than perfection-chasing, optimizing for **realistic damage control** instead of idealized perfection. The advice must be **compact** and **short** to be **immediately actionable**, as the user reads it while already out.

\textbf{Reason.} Wrong: misses the workable real-meal objective.
\end{wronganswer}

\begin{correctanswer}{Qwen3.5-35B-A3B --- Correct}
\textbf{Answer.} The assistant should frame social-eating advice as **damage prevention** (or **realistic damage control**) rather than **perfection-chasing** (or **perfection**). The tone and objective should be **short, immediately actionable, and compact** because the user reads the advice while **already out**. This approach combines the family-meal framing of **damage control over perfection** with the **out-of-home compression** of keeping advice **immediately actionable** and **blunt** (e.g., "protein first, modest starch, water") to avoid **negotiating with menu boards** or **communal table ordering** becoming the real meal.

\textbf{Reason.} Correct: combines damage prevention, compact actionable advice, and the real-meal objective.
\end{correctanswer}

\begin{wronganswer}{Qwen3.5-397B --- Incorrect}
\textbf{Answer.} The assistant should frame advice as damage prevention rather than perfection-chasing, optimizing for realistic damage control. The advice must be kept compact, short, and immediately actionable because the user reads it while already out.

\textbf{Reason.} Wrong: misses the workable real-meal objective.
\end{wronganswer}

\begin{wronganswer}{Qwen3.8-27B --- Incorrect}
\textbf{Answer.} The assistant should frame social-eating advice with a **damage-control (or damage-prevention) objective** rather than chasing perfection, and keep the tone **compact and immediately actionable** because the user is already out.

\textbf{Reason.} Wrong: misses damage prevention, the workable real-meal objective.
\end{wronganswer}

\begin{correctanswer}{GPT-oss-120B --- Correct}
\textbf{Answer.} Frame the advice as a quick, pragmatic ``damage-control'' cue rather than a quest for perfection. Use a blunt, short-form rule that the user can apply on the spot (e.g., ``protein first, modest starch, water'') and present it in a matter-of-facts, supportive tone that emphasizes keeping the meal safe and enjoyable rather than ideal. The objective is to prevent any major ``damage'' (over-eating, awkward choices, etc.) with the simplest, most realistic guidance possible.

\textbf{Reason.} Correct: combines damage prevention, compact actionable advice, and the real-meal objective.
\end{correctanswer}

\begin{correctanswer}{Llama-3.1-8B-Instruct --- Correct}
\textbf{Answer.} For social meal planning, the assistant should frame advice as damage prevention rather than perfection-chasing. When the user is already out, the assistant should use the blunt short version of the advice, which is: protein first, small starch side, water, no combo heroics, and shared fries are not a personality test.

\textbf{Reason.} Correct: combines damage prevention, compact actionable advice, and the real-meal objective.
\end{correctanswer}

\begin{wronganswer}{Llama-3.1-70B-Instruct --- Incorrect}
\textbf{Answer.} The assistant should frame social-eating advice as damage prevention, optimizing for realistic damage control rather than idealized perfection. The advice should be compact.

\textbf{Reason.} Wrong: misses the workable real-meal objective.
\end{wronganswer}

\begin{wronganswer}{DeepSeek-V4-Pro --- Incorrect}
\textbf{Answer.} Frame it as realistic damage control, not perfection-chasing, and keep it short and immediately actionable because the user reads it while already out.

\textbf{Reason.} Wrong: misses the workable real-meal objective.
\end{wronganswer}

\begin{wronganswer}{GLM-5.2 --- Incorrect}
\textbf{Answer.} The assistant should frame social-eating advice as damage prevention rather than perfection-chasing.

\textbf{Reason.} Wrong: misses compact/actionable framing, the workable real-meal objective.
\end{wronganswer}

\begin{wronganswer}{Kimi-K3 --- Incorrect}
\textbf{Answer.} Frame it as realistic damage prevention, not perfection-chasing, and keep it short and immediately actionable for someone already out.

\textbf{Reason.} Wrong: misses the workable real-meal objective.
\end{wronganswer}

\begin{correctanswer}{MiniMax-M2.5 --- Correct}
\textbf{Answer.} Frame social-eating advice with a realistic, damage-control tone that prioritizes preventing overindulgence or regret rather than chasing nutritional perfection. Keep the advice compact, short, and immediately actionable since the user is already out--using blunt, easy-to-remember rules like "protein first, modest starch, water" or its shorter variants when the setting is noisy or the menu is shawarma-heavy.

\textbf{Reason.} Correct: combines damage prevention, compact actionable advice, and the real-meal objective.
\end{correctanswer}

\begin{correctanswer}{GPT-4o-mini --- Correct}
\textbf{Answer.} The assistant should frame social-eating advice as damage prevention rather than perfection-chasing, keeping it short and immediately actionable. The focus should be on practical strategies like choosing grilled protein first, keeping the starch side modest, skipping fries unless deliberately choosing them, and defaulting to water to avoid derailing the meal. Additionally, the assistant should remind the user not to let communal table ordering become their real meal, reinforcing the safest approach: protein first, modest starch, and water.

\textbf{Reason.} Correct: combines damage prevention, compact actionable advice, and the real-meal objective.
\end{correctanswer}

\begin{correctanswer}{GPT-5.6-Sol --- Correct}
\textbf{Answer.} Frame it as realistic damage prevention/control--not perfection--especially for Nigerian family meals. Start with a pre-decision plan, then give short, immediately actionable food guidance because the user is already out.

\textbf{Reason.} Correct: combines damage prevention, compact actionable advice, and the real-meal objective.
\end{correctanswer}

\begin{wronganswer}{Gemini-3.1-Pro-Preview --- Incorrect}
\textbf{Answer.} For social eating, the assistant should frame advice as damage prevention and realistic damage control rather than perfection-chasing. The advice must also be kept short, compact, and immediately actionable because the user reads it while already out.

\textbf{Reason.} Wrong: misses the workable real-meal objective.
\end{wronganswer}

\begin{correctanswer}{Doubao-Seed-2.1-Pro --- Correct}
\textbf{Answer.} The assistant should frame the advice as realistic damage control/damage prevention rather than chasing idealized perfection (matching the non-perfectionist, damage-preferred framing for both social meals and Nigerian family meals), and keep the advice compact, since the user reads it while already out. The guidance should optimize for realistic damage control rather than perfection, following the short, practical framing (like the concise restaurant rule: protein first, modest starch, water).

\textbf{Reason.} Correct: combines damage prevention, compact actionable advice, and the real-meal objective.
\end{correctanswer}

\end{casebox}

\clearpage

\subsection{Type II: Adjacent Rule Leakage}

\begin{casebox}{Case 2: Adjacent Rule Leakage}
\noindent\textbf{Category.} Physical Health / Workout Planning\hfill\textbf{Data type.} agent centric\\

\begin{goldbox}
\textbf{Query.} Under the assistant's standard check-in format, which line should state whether the issue is a preference problem, a constraint problem, or nothing at all, and what exact substitute should appear when nothing is wrong?

\textbf{Reference Answer.} The fourth line should state the problem type, and when nothing is wrong the assistant should write "problem type: none."
\end{goldbox}

\begin{analysisbox}
The key distinction is not only the fallback phrase, but also where it appears in the standing check-in format. Wrong answers often retrieve the nearby phrase while omitting the fourth-line constraint.
\end{analysisbox}

\textbf{Model Responses.}
\begin{wronganswer}{Qwen3-4B --- Incorrect}
\textbf{Answer.} problem type: none

\textbf{Reason.} Wrong: gives the substitute phrase but misses the fourth-line requirement.
\end{wronganswer}

\begin{wronganswer}{Qwen3-8B --- Incorrect}
\textbf{Answer.} Under the assistant's standard check-in format, the line stating whether the issue is a preference problem, a constraint problem, or nothing at all should be "problem type: none" when nothing is wrong.

\textbf{Reason.} Wrong: gives the substitute phrase but misses the fourth-line requirement.
\end{wronganswer}

\begin{wronganswer}{Qwen3-30B-A3B --- Incorrect}
\textbf{Answer.} The line stating whether the issue is a preference problem, a constraint problem, or nothing at all should be included in the check-in format. When nothing is wrong, the exact substitute that should appear is 'problem type: none'.

\textbf{Reason.} Wrong: gives the substitute phrase but misses the fourth-line requirement.
\end{wronganswer}

\begin{wronganswer}{Qwen3.5-4B --- Incorrect}
\textbf{Answer.} The line stating whether the issue is a preference problem, a constraint problem, or nothing at all is the line indicating the problem type. When nothing is wrong, the exact substitute to appear is 'problem type: none'.

\textbf{Reason.} Wrong: gives the substitute phrase but misses the fourth-line requirement.
\end{wronganswer}

\begin{wronganswer}{Qwen3.5-9B --- Incorrect}
\textbf{Answer.} Under the assistant's standard check-in format, the fifth line should state whether the issue is a preference problem, a constraint problem, or nothing at all. When nothing is wrong, the exact substitute that should appear is 'problem type: none'.

\textbf{Reason.} Wrong: gives the substitute phrase but misses the fourth-line requirement.
\end{wronganswer}

\begin{correctanswer}{Qwen3.5-27B --- Correct}
\textbf{Answer.} The fourth line, labeled "problem type," should state the issue. When nothing is wrong, the exact substitute that must appear is "problem type: none".

\textbf{Reason.} Correct: identifies the fourth line and the exact substitute phrase.
\end{correctanswer}

\begin{correctanswer}{Qwen3.5-35B-A3B --- Correct}
\textbf{Answer.} The fourth line should state whether the issue is a preference problem or a constraint problem, and the exact substitute "problem type: none" should appear when nothing is wrong.

\textbf{Reason.} Correct: identifies the fourth line and the exact substitute phrase.
\end{correctanswer}

\begin{correctanswer}{Qwen3.5-397B --- Correct}
\textbf{Answer.} Under the standard five-line check-in format, the fourth line ("problem type") should state whether the issue is a preference problem or a constraint problem. If nothing is wrong, the exact substitute to appear is "problem type: none".

\textbf{Reason.} Correct: identifies the fourth line and the exact substitute phrase.
\end{correctanswer}

\begin{correctanswer}{Qwen3.8-27B --- Correct}
\textbf{Answer.} The fourth line, labeled "problem type," should state whether the issue is a preference problem or a constraint problem. When nothing is wrong, the exact substitute that should appear is "problem type: none".

\textbf{Reason.} Correct: identifies the fourth line and the exact substitute phrase.
\end{correctanswer}

\begin{correctanswer}{GPT-oss-120B --- Correct}
\textbf{Answer.} The **fourth line** (``problem type'') is the one that indicates whether the issue is a preference problem, a constraint problem, or nothing at all. When nothing is wrong, the line must read exactly: **problem type: none**

\textbf{Reason.} Correct: identifies the fourth line and the exact substitute phrase.
\end{correctanswer}

\begin{wronganswer}{Llama-3.1-8B-Instruct --- Incorrect}
\textbf{Answer.} The line stating whether the issue is a preference problem, a constraint problem, or nothing at all should be 'problem type: none' when nothing is wrong.

\textbf{Reason.} Wrong: gives the substitute phrase but misses the fourth-line requirement.
\end{wronganswer}

\begin{correctanswer}{Llama-3.1-70B-Instruct --- Correct}
\textbf{Answer.} The line for stating the problem type is the fourth line. When nothing is wrong, the exact substitute should be 'problem type: none'.

\textbf{Reason.} Correct: identifies the fourth line and the exact substitute phrase.
\end{correctanswer}

\begin{wronganswer}{DeepSeek-V4-Pro --- Incorrect}
\textbf{Answer.} The line that states whether the issue is a preference problem or a constraint problem should be included in the check-in format. When nothing is wrong, the exact substitute that should appear is "problem type: none".

\textbf{Reason.} Wrong: gives the substitute phrase but misses the fourth-line requirement.
\end{wronganswer}

\begin{wronganswer}{GLM-5.2 --- Incorrect}
\textbf{Answer.} The **problem type** line should state whether the issue is a preference problem, a constraint problem, or nothing at all. When nothing is wrong, the exact substitute is **'problem type: none'**.

\textbf{Reason.} Wrong: gives the substitute phrase but misses the fourth-line requirement.
\end{wronganswer}

\begin{correctanswer}{Kimi-K3 --- Correct}
\textbf{Answer.} The fourth line, "problem type," should state whether the issue is a preference problem or a constraint problem. When nothing is wrong, the exact substitute should be "problem type: none".

\textbf{Reason.} Correct: identifies the fourth line and the exact substitute phrase.
\end{correctanswer}

\begin{correctanswer}{MiniMax-M2.5 --- Correct}
\textbf{Answer.} The fourth line (problem type) should state whether the issue is a preference problem, a constraint problem, or nothing at all. When nothing is wrong, the exact substitute should be "problem type: none".

\textbf{Reason.} Correct: identifies the fourth line and the exact substitute phrase.
\end{correctanswer}

\begin{wronganswer}{GPT-4o-mini --- Incorrect}
\textbf{Answer.} The line should state whether the problem is a preference problem or a constraint problem. If nothing is wrong, the exact substitute that should appear is 'problem type: none'.

\textbf{Reason.} Wrong: gives the substitute phrase but misses the fourth-line requirement.
\end{wronganswer}

\begin{correctanswer}{GPT-5.6-Sol --- Correct}
\textbf{Answer.} The fourth line, **problem type**, should state whether it is a preference or constraint problem. When nothing is wrong, write exactly: **``problem type: none''**.

\textbf{Reason.} Correct: identifies the fourth line and the exact substitute phrase.
\end{correctanswer}

\begin{correctanswer}{Gemini-3.1-Pro-Preview --- Correct}
\textbf{Answer.} The fourth line (problem type) should state whether the issue is a preference problem or a constraint problem. When nothing is wrong, the exact substitute should be "problem type: none".

\textbf{Reason.} Correct: identifies the fourth line and the exact substitute phrase.
\end{correctanswer}

\begin{wronganswer}{Doubao-Seed-2.1-Pro --- Incorrect}
\textbf{Answer.} The problem type line should state whether the issue is a preference problem or a constraint problem. When nothing is wrong, the exact substitute that should appear is "problem type: none".

\textbf{Reason.} Wrong: gives the substitute phrase but misses the fourth-line requirement.
\end{wronganswer}

\end{casebox}

\clearpage

\begin{casebox}{Case 3: Adjacent Rule Leakage}
\noindent\textbf{Category.} Career \& Learning / Collaboration Style\hfill\textbf{Data type.} user centric\\

\begin{goldbox}
\textbf{Query.} When documentation starts interfering with facilitation, what is the user's current practice about timing log updates, and what specific failure does that avoid in the user's Notion-based setup, given the broader preference for keeping the system lightweight?

\textbf{Reference Answer.} The user updates the log after meetings rather than during them, because typing too much live makes them lose the room and stop hearing what people are actually doing.
\end{goldbox}

\begin{analysisbox}
This adjacent-rule case stresses a failure in meeting documentation. Correct answers identify after-meeting log updates and the specific reason: live typing makes the user lose the room and stop hearing what people are doing.
\end{analysisbox}

\textbf{Model Responses.}
\begin{wronganswer}{Qwen3-4B --- Incorrect}
\textbf{Answer.} The user updates their log after meetings rather than live during meetings. This practice avoids the failure of agenda items mixing update and decision, which is one of the user's main meeting failure points.

\textbf{Reason.} Wrong: imports a nearby Notion/process failure instead of the facilitation failure.
\end{wronganswer}

\begin{wronganswer}{Qwen3-8B --- Incorrect}
\textbf{Answer.} The user updates their log after meetings rather than live during meetings. This practice avoids the failure of the Notion log getting clogged due to repeated revisions of half-decisions, aligning with the preference for keeping the system lightweight.

\textbf{Reason.} Wrong: imports a nearby Notion/process failure instead of the facilitation failure.
\end{wronganswer}

\begin{correctanswer}{Qwen3-30B-A3B --- Correct}
\textbf{Answer.} The user's current practice is to update their log after meetings rather than during them. This avoids the failure of losing the room, which occurs when the user types too much during a meeting and stops hearing what people are actually doing.

\textbf{Reason.} Correct: identifies after-meeting updates and the losing-the-room failure.
\end{correctanswer}

\begin{wronganswer}{Qwen3.5-4B --- Incorrect}
\textbf{Answer.} The user's current practice is to update the log after meetings rather than live during them. This avoids the specific failure of losing the room, which occurs when live updating pulls the user out of facilitation mode. This approach aligns with the user's preference for lightweight tools and processes over overcomplicated systems, as it prevents the Notion log from getting clogged by repeated revisions that happen when people half-decide things.

\textbf{Reason.} Wrong: imports a nearby Notion/process failure instead of the facilitation failure.
\end{wronganswer}

\begin{correctanswer}{Qwen3.5-9B --- Correct}
\textbf{Answer.} The user's current practice is to update their Notion log **after meetings** rather than live during them. This timing avoids the specific failure of **losing the room** (or being pulled out of facilitation mode), which occurs when live updating interferes with facilitation. This aligns with the user's broader preference for lightweight tools and processes, as the assistant noted that live note-taking can disrupt the facilitation mode and that the user prefers avoiding overcomplicated systems.

\textbf{Reason.} Correct: identifies after-meeting updates and the losing-the-room failure.
\end{correctanswer}

\begin{correctanswer}{Qwen3.5-27B --- Correct}
\textbf{Answer.} The user's current practice is to update their log after meetings rather than live during them. This timing avoids the specific failure of losing the room and stopping hearing what people are actually doing, which occurs when typing too much during a meeting. Additionally, this approach helps prevent the Notion log from getting clogged by people half-deciding things and revising them repeatedly, aligning with the user's preference for lightweight tools over overcomplicated systems.

\textbf{Reason.} Correct: identifies after-meeting updates and the losing-the-room failure.
\end{correctanswer}

\begin{correctanswer}{Qwen3.5-35B-A3B --- Correct}
\textbf{Answer.} The user's current practice is to update their log after meetings rather than live during them. This avoids the specific failure of stopping hearing what people are actually doing, which occurs when the user types too much during a meeting. Additionally, this approach helps prevent the log from becoming clogged by people half-deciding things and then revising them repeatedly, aligning with the user's preference for lightweight tools over overcomplicated systems.

\textbf{Reason.} Correct: identifies after-meeting updates and the losing-the-room failure.
\end{correctanswer}

\begin{wronganswer}{Qwen3.5-397B --- Incorrect}
\textbf{Answer.} The user currently updates their log after meetings rather than live during them. This practice avoids the specific failure of stopping hearing what people are actually doing when typing too much, and prevents the Notion log from getting clogged when people half-decide things and then revise them repeatedly, aligning with their preference for lightweight processes over overcomplicated systems.

\textbf{Reason.} Wrong: imports a nearby Notion/process failure instead of the facilitation failure.
\end{wronganswer}

\begin{wronganswer}{Qwen3.8-27B --- Incorrect}
\textbf{Answer.} The user's current practice is to update their log **after meetings** rather than live during them. This timing avoids the specific failure where the Notion log gets **clogged** when people half-decide things and then revise them repeatedly. This approach aligns with the user's broader preference for **lightweight tools and processes** over overcomplicated systems.

\textbf{Reason.} Wrong: imports a nearby Notion/process failure instead of the facilitation failure.
\end{wronganswer}

\begin{wronganswer}{GPT-oss-120B --- Incorrect}
\textbf{Answer.} When documentation threatens to get in the way of facilitation, the user now **writes the log after the meeting rather than live**. By postponing entry, they avoid the **clogging of their Notion log with half-decided, repeatedly revised items**--keeping only confirmed decisions and preserving a lightweight, uncluttered system.

\textbf{Reason.} Wrong: misses the losing-the-room reason.
\end{wronganswer}

\begin{correctanswer}{Llama-3.1-8B-Instruct --- Correct}
\textbf{Answer.} The user updates their log after meetings rather than live during meetings. This practice avoids the failure of losing the room, which occurs when live updating makes them lose the room.

\textbf{Reason.} Correct: identifies after-meeting updates and the losing-the-room failure.
\end{correctanswer}

\begin{wronganswer}{Llama-3.1-70B-Instruct --- Incorrect}
\textbf{Answer.} The user updates their log after meetings, rather than live during meetings. This avoids the issue of losing the room. Additionally, updating after meetings also helps prevent the user's Notion log from getting clogged, as it allows for more finalized decisions to be recorded, rather than half-decided and repeatedly revised items. This practice aligns with the user's preference for lightweight tools and processes.

\textbf{Reason.} Wrong: imports a nearby Notion/process failure instead of the facilitation failure.
\end{wronganswer}

\begin{correctanswer}{DeepSeek-V4-Pro --- Correct}
\textbf{Answer.} The user updates the log after meetings rather than live during meetings. This avoids the failure of losing the room--if they type too much during a meeting, they stop hearing what people are actually doing. The broader lightweight preference is reflected in keeping the Notion log minimal (decision, owner, due date, status, notes only when explaining a dependency) and in separating confirmed decisions from proposed ones so the log doesn't get clogged with half-decisions that are repeatedly revised.

\textbf{Reason.} Correct: identifies after-meeting updates and the losing-the-room failure.
\end{correctanswer}

\begin{correctanswer}{GLM-5.2 --- Correct}
\textbf{Answer.} The user updates their log **after meetings rather than live during meetings**, because live updating makes them "lose the room" -- if they type too much during a meeting, they stop hearing what people are actually doing. This timing avoids a specific failure in the Notion-based setup: the log gets **clogged when people half-decide things and then revise them repeatedly**. This aligns with the user's broader preference for **lightweight tools and processes over overcomplicated systems**, and the assistant's recommendation to keep the Notion decision log minimal (only decision, owner, due date, status, and notes when explaining a dependency), warning that too many fields would make the log harder to maintain than it is useful.

\textbf{Reason.} Correct: identifies after-meeting updates and the losing-the-room failure.
\end{correctanswer}

\begin{correctanswer}{Kimi-K3 --- Correct}
\textbf{Answer.} The user updates their log after meetings rather than live during them. This avoids the failure where typing too much during a meeting makes them stop hearing what people are actually doing, and it keeps their Notion log from getting clogged by half-decided items that are later revised repeatedly.

\textbf{Reason.} Correct: identifies after-meeting updates and the losing-the-room failure.
\end{correctanswer}

\begin{wronganswer}{MiniMax-M2.5 --- Incorrect}
\textbf{Answer.} The user updates their log after meetings rather than during them. This practice avoids the Notion log becoming clogged with half-decisions and repeated revisions, which aligns with their preference for a lightweight system by keeping only confirmed decisions as entries.

\textbf{Reason.} Wrong: imports a nearby Notion/process failure instead of the facilitation failure.
\end{wronganswer}

\begin{wronganswer}{GPT-4o-mini --- Incorrect}
\textbf{Answer.} The user's current practice is to update their log after meetings rather than live during meetings. This avoids the specific failure of their Notion log getting clogged when people half-decide things and then revise them repeatedly, aligning with their broader preference for keeping the system lightweight.

\textbf{Reason.} Wrong: imports a nearby Notion/process failure instead of the facilitation failure.
\end{wronganswer}

\begin{correctanswer}{GPT-5.6-Sol --- Correct}
\textbf{Answer.} The user updates their log after meetings rather than live, avoiding the loss of the room and the failure to hear what people are actually doing. This also prevents the Notion log from becoming clogged with half-decided items that are repeatedly revised, consistent with keeping the process lightweight.

\textbf{Reason.} Correct: identifies after-meeting updates and the losing-the-room failure.
\end{correctanswer}

\begin{correctanswer}{Gemini-3.1-Pro-Preview --- Correct}
\textbf{Answer.} The user updates their log after meetings rather than live during meetings. This avoids the failure of losing the room and stopping hearing what people are actually doing, which happens if the user types too much during a meeting.

\textbf{Reason.} Correct: identifies after-meeting updates and the losing-the-room failure.
\end{correctanswer}

\begin{correctanswer}{Doubao-Seed-2.1-Pro --- Correct}
\textbf{Answer.} The user's current practice is to update their log after meetings rather than live during meetings. This avoids the specific failure where live updating/typing too much during a meeting makes them lose the room and stop hearing what people are actually saying, which aligns with their preference for lightweight, non-overcomplicated processes; it also prevents their Notion log from getting clogged by half-decided, repeatedly revised decisions, which would bloat the system they keep specifically for cross-team planning meeting decision logs to document agreed items.

\textbf{Reason.} Correct: identifies after-meeting updates and the losing-the-room failure.
\end{correctanswer}

\end{casebox}

\clearpage

\subsection{Type III: Incomplete or Over-Inclusive Set}

\begin{casebox}{Case 4: Incomplete or Over-Inclusive Set}
\noindent\textbf{Category.} Physical Health / Medical Visits \& Medication\hfill\textbf{Data type.} user centric\\

\begin{goldbox}
\textbf{Query.} Looking at the user's present supplement behavior, which daily anchor matters more to success than ingredient claims, and which two current products clearly depend on that anchor?

\textbf{Reference Answer.} Breakfast is the key anchor, and it clearly supports vitamin D and creatine.
\end{goldbox}

\begin{analysisbox}
The case tests set membership under a stable daily anchor. Correct answers preserve breakfast as the anchor and identify vitamin D plus creatine; wrong answers often substitute whey or another supplement.
\end{analysisbox}

\textbf{Model Responses.}
\begin{wronganswer}{Qwen3-4B --- Incorrect}
\textbf{Answer.} The daily anchor that matters more to success than ingredient claims is a stable breakfast anchor. The two current products that clearly depend on this anchor are vitamin D on weekday breakfasts during the colder months and plain whey.

\textbf{Reason.} Wrong: substitutes whey for the current product set.
\end{wronganswer}

\begin{wronganswer}{Qwen3-8B --- Incorrect}
\textbf{Answer.} The user's breakfast routine matters more to success than ingredient claims. The two current products that clearly depend on this anchor are vitamin D and plain whey.

\textbf{Reason.} Wrong: substitutes whey for the current product set.
\end{wronganswer}

\begin{wronganswer}{Qwen3-30B-A3B --- Incorrect}
\textbf{Answer.} The daily anchor that matters more to success than ingredient claims is the user's breakfast routine. The two current products that clearly depend on that anchor are vitamin D and plain whey.

\textbf{Reason.} Wrong: substitutes whey for the current product set.
\end{wronganswer}

\begin{wronganswer}{Qwen3.5-4B --- Incorrect}
\textbf{Answer.} Based on the provided memories, the daily anchor that matters more to success than ingredient claims is a **stable breakfast**. The two current products that clearly depend on this anchor are **vitamin D** and **electrolyte powder**. * **Vitamin D**: The user only reliably takes this with breakfast (specifically 25 micrograms with breakfast on weekdays during colder months). * **Electrolyte powder**: This is the only exception to the user's rule of keeping supplements to the morning; it is taken after runs longer than an hour, which implies it is also anchored to the morning routine (post-run) rather than spread across the day.

\textbf{Reason.} Wrong: misses creatine.
\end{wronganswer}

\begin{wronganswer}{Qwen3.5-9B --- Incorrect}
\textbf{Answer.} Based on the provided memories: * The daily anchor that matters more to success than ingredient claims is the **stable breakfast anchor**. * The two current products that clearly depend on this anchor are **vitamin D** (taken on weekday breakfasts) and **plain whey** (taken as part of the breakfast routine).

\textbf{Reason.} Wrong: substitutes whey for the current product set.
\end{wronganswer}

\begin{correctanswer}{Qwen3.5-27B --- Correct}
\textbf{Answer.} The **breakfast routine** is the daily anchor that matters more to success than ingredient claims. The two current products that clearly depend on this anchor are **vitamin D** (taken with breakfast on weekdays during colder months) and **creatine** (stirred into water with breakfast).

\textbf{Reason.} Correct: identifies breakfast as the anchor and vitamin D plus creatine as the products.
\end{correctanswer}

\begin{wronganswer}{Qwen3.5-35B-A3B --- Incorrect}
\textbf{Answer.} The daily anchor that matters more to success than ingredient claims is the **stable breakfast anchor** (specifically the **break** routine). The two current products that clearly depend on this anchor are: 1. **Vitamin D** (taken with weekday breakfasts) 2. **Plain whey** (taken when useful, implied to be with breakfast) *Note: While the memories state the breakfast anchor contributed more to consistency than ingredient claims, they explicitly list Vitamin D as the only product tied to "weekday breakfasts." Plain whey is listed as "when useful" or "when useful," which contextually aligns with the breakfast anchor mentioned in the summary, whereas electrolyte powder depends on runs and creatine depends on low friction.* **Correction based on strict memory parsing:** The memories state the **breakfast routine** is the anchor. The memory states: "The assistant told the user that a stable breakfast anchor had probably contributed more to their supplement consistency than ingredient claims had." The memory states the shortlist includes: "vitamin D on weekday breakfasts" and "plain whey... when useful." However, the prompt asks which *two* products clearly depend on that anchor. The memory explicitly links **Vitamin D** to "weekday breakfasts." It does not explicitly state that **plain whey** depends on the breakfast anchor, only that it is taken "when useful" or "when useful." The other two (electrolyte powder, creatine) depend on runs and low friction, respec ... [truncated]

\textbf{Reason.} Wrong: substitutes whey for the current product set.
\end{wronganswer}

\begin{correctanswer}{Qwen3.5-397B --- Correct}
\textbf{Answer.} The user's breakfast routine is the daily anchor that matters more to success than ingredient claims. The two current products that clearly depend on this anchor are vitamin D and creatine.

\textbf{Reason.} Correct: identifies breakfast as the anchor and vitamin D plus creatine as the products.
\end{correctanswer}

\begin{correctanswer}{Qwen3.8-27B --- Correct}
\textbf{Answer.} The daily anchor that matters more to success than ingredient claims is the user's **breakfast routine**. The two current products that clearly depend on that anchor are: 1. **Vitamin D** (taken with breakfast on weekdays during the colder months) 2. **Creatine** (stirred into water with breakfast)

\textbf{Reason.} Correct: identifies breakfast as the anchor and vitamin D plus creatine as the products.
\end{correctanswer}

\begin{correctanswer}{GPT-oss-120B --- Correct}
\textbf{Answer.} The daily anchor that matters most is **breakfast**. The two current supplements that clearly rely on that anchor are **vitamin D** and **creatine**.

\textbf{Reason.} Correct: identifies breakfast as the anchor and vitamin D plus creatine as the products.
\end{correctanswer}

\begin{wronganswer}{Llama-3.1-8B-Instruct --- Incorrect}
\textbf{Answer.} The user's present supplement behavior is anchored to their breakfast routine. The two current products that clearly depend on this anchor are: 1. Vitamin D (taken on weekday breakfasts during the colder months) 2. Plain whey

\textbf{Reason.} Wrong: substitutes whey for the current product set.
\end{wronganswer}

\begin{wronganswer}{Llama-3.1-70B-Instruct --- Incorrect}
\textbf{Answer.} The user's breakfast routine is the main anchor supporting their supplement routine, and it matters more to their supplement consistency than ingredient claims. Two current products that clearly depend on this anchor are vitamin D (taken on weekday breakfasts during the colder months) and plain whey.

\textbf{Reason.} Wrong: substitutes whey for the current product set.
\end{wronganswer}

\begin{correctanswer}{DeepSeek-V4-Pro --- Correct}
\textbf{Answer.} The daily anchor that matters more to success than ingredient claims is the user's breakfast routine. The two current products that clearly depend on that anchor are vitamin D (25 micrograms with breakfast on weekdays during the colder months) and creatine (stirred into water with breakfast).

\textbf{Reason.} Correct: identifies breakfast as the anchor and vitamin D plus creatine as the products.
\end{correctanswer}

\begin{correctanswer}{GLM-5.2 --- Correct}
\textbf{Answer.} The daily anchor that matters more to success than ingredient claims is the user's **breakfast routine**. The two current products that clearly depend on that anchor are **vitamin D** (taken with weekday breakfasts in the colder months) and **creatine** (which stays easy and unannoying as part of the breakfast-anchored routine).

\textbf{Reason.} Correct: identifies breakfast as the anchor and vitamin D plus creatine as the products.
\end{correctanswer}

\begin{correctanswer}{Kimi-K3 --- Correct}
\textbf{Answer.} The daily anchor that matters more than ingredient claims is the user's stable breakfast routine. The two current products that clearly depend on that anchor are vitamin D and creatine.

\textbf{Reason.} Correct: identifies breakfast as the anchor and vitamin D plus creatine as the products.
\end{correctanswer}

\begin{correctanswer}{MiniMax-M2.5 --- Correct}
\textbf{Answer.} The breakfast anchor matters more to success than ingredient claims. The two current products that clearly depend on this anchor are vitamin D and creatine.

\textbf{Reason.} Correct: identifies breakfast as the anchor and vitamin D plus creatine as the products.
\end{correctanswer}

\begin{wronganswer}{GPT-4o-mini --- Incorrect}
\textbf{Answer.} The daily anchor that matters more to success than ingredient claims is the user's breakfast routine. The two current products that clearly depend on that anchor are vitamin D on weekday breakfasts during the colder months and plain whey.

\textbf{Reason.} Wrong: substitutes whey for the current product set.
\end{wronganswer}

\begin{wronganswer}{GPT-5.6-Sol --- Incorrect}
\textbf{Answer.} The stable breakfast anchor matters more than ingredient claims. The two products that clearly depend on it are the user's morning supplements generally and vitamin D specifically.

\textbf{Reason.} Wrong: misses creatine.
\end{wronganswer}

\begin{correctanswer}{Gemini-3.1-Pro-Preview --- Correct}
\textbf{Answer.} The user's breakfast routine matters more to their supplement consistency than ingredient claims. The two current products that depend on this anchor are vitamin D and creatine.

\textbf{Reason.} Correct: identifies breakfast as the anchor and vitamin D plus creatine as the products.
\end{correctanswer}

\begin{correctanswer}{Doubao-Seed-2.1-Pro --- Correct}
\textbf{Answer.} The stable breakfast routine (the user's breakfast routine) matters more to supplement consistency success than ingredient claims, per the assistant. The two current products that clearly depend on this anchor are: 1. Vitamin D (25 micrograms, taken only with weekday breakfasts in colder months; the user only reliably takes it with breakfast) 2. Creatine (which has been tolerable/low-friction because the user stirs it into water with breakfast).

\textbf{Reason.} Correct: identifies breakfast as the anchor and vitamin D plus creatine as the products.
\end{correctanswer}

\end{casebox}

\clearpage

\begin{casebox}{Case 5: Incomplete or Over-Inclusive Set}
\noindent\textbf{Category.} Career \& Learning / Project Responsibilities\hfill\textbf{Data type.} agent centric\\

\begin{goldbox}
\textbf{Query.} How should the assistant decide what belongs in leadership messaging when different audiences need separate release notes, and what evidence suggested that this separation worked?

\textbf{Reference Answer.} The assistant should keep the audiences separate and include only decision-relevant information for leadership, which was supported by the user's report that the separate-audience rule helped and by Dana's very fast reply to the leadership version.
\end{goldbox}

\begin{analysisbox}
This set-boundary case tests whether the model preserves the leadership-message boundary. Correct answers keep audiences separate, restrict leadership notes to decision-relevant content, and cite both the user report and Dana's fast reply as evidence.
\end{analysisbox}

\textbf{Model Responses.}
\begin{wronganswer}{Qwen3-4B --- Incorrect}
\textbf{Answer.} The assistant should decide what belongs in leadership messaging by focusing on decision, impact, and ask for leadership updates, and scope, owner, and next step for team updates. Evidence that this separation worked includes the user asking for both leadership and team release notes with distinct wording, the assistant drafting separate release notes for leadership and team, and the user noting that mixed-audience messages tend to sound mushy. Additionally, the user reported that the audience-relative messaging principle helped avoid overexplaining.

\textbf{Reason.} Wrong: misses Dana's fast reply.
\end{wronganswer}

\begin{wronganswer}{Qwen3-8B --- Incorrect}
\textbf{Answer.} The assistant should decide what belongs in leadership messaging by focusing on decision, impact, and ask, while keeping team messaging focused on scope, owner, and next step. The evidence that this separation worked is that the assistant drafted separate Atlas 3.3 release notes for leadership and for the team, both framed around the 14th release target and current-scope assumptions, and the user indicated that mixed-audience messages tend to sound mushy.

\textbf{Reason.} Wrong: misses audience separation, Dana's fast reply.
\end{wronganswer}

\begin{correctanswer}{Qwen3-30B-A3B --- Correct}
\textbf{Answer.} The assistant should decide what belongs in leadership messaging by focusing on decision-relevant risk, ensuring the main signal stays clear and the messages are short, audience-specific, and honest about what is still conditional. The evidence that this separation worked includes the user's report that the separate-audience rule helped a lot when explaining the Atlas 3.4 status upward, and the assistant's observation that Dana's very fast reply suggested the leadership update fit its audience and was easy to act on.

\textbf{Reason.} Correct: keeps audiences separate, limits leadership to decision-relevant content, and cites both success signals.
\end{correctanswer}

\begin{correctanswer}{Qwen3.5-4B --- Correct}
\textbf{Answer.} The assistant should decide what belongs in leadership messaging by including only decision-relevant risk, as the user tends to want to include every caveat in leadership notes to avoid accusations of hiding anything, but the principle dictates that not everything important belongs in upward communication. Rollback details should be included in leadership updates only when they change the audience's decision or risk posture; otherwise, they should stay in the team-facing update. Evidence that this separation worked includes the user reporting that Dana replied to the leadership version in about thirty seconds, and the assistant noting that Dana's very fast reply suggests the leadership update fit its audience and was easy to act on. The user also reported that the separate-audience rule helped a lot when explaining Atlas 3.4 status upward.

\textbf{Reason.} Correct: keeps audiences separate, limits leadership to decision-relevant content, and cites both success signals.
\end{correctanswer}

\begin{correctanswer}{Qwen3.5-9B --- Correct}
\textbf{Answer.} The assistant should decide what belongs in leadership messaging based on the principle that "not everything important belongs in upward communication" and that "what counts as important is audience-relative." Specifically, leadership notes should include only "decision-relevant risk" and focus on "decision, impact, and ask." Additional caveats and operational details (like rollback information, unless it changes the audience's decision or risk posture) should be left for the team update to keep the main signal clear. Evidence that this separation worked includes: 1. The user reported that Dana replied to the leadership version in "about thirty seconds." 2. The assistant noted that Dana's "very fast reply suggests the leadership update fit its audience and was easy to act on." 3. The user stated they "had to explain Atlas 3.4 status upward and that the separate-audience rule helped a lot." 4. The general observation that "Audience-specific communication works better than one-size-fits-all updates."

\textbf{Reason.} Correct: keeps audiences separate, limits leadership to decision-relevant content, and cites both success signals.
\end{correctanswer}

\begin{wronganswer}{Qwen3.5-27B --- Incorrect}
\textbf{Answer.} The assistant should decide what belongs in leadership messaging by focusing on decision, impact, and ask, while including only decision-relevant risk. Rollback details should be included only if they change the audience's decision or risk posture; otherwise, they belong in the team-facing update. Additional caveats should be left for the team update to keep the main signal clear. The evidence suggesting this separation worked was that Dana replied to the leadership version in about thirty seconds, which indicated the update fit its audience and was easy to act on.

\textbf{Reason.} Wrong: misses audience separation.
\end{wronganswer}

\begin{correctanswer}{Qwen3.5-35B-A3B --- Correct}
\textbf{Answer.} The assistant should decide what belongs in leadership messaging by including only decision-relevant risk and impact, while leaving caveats, scope, and operational details (like rollback details) for the team update unless the information changes the audience's decision or risk posture. This separation works because the audience relative to the message determines what counts as important, preventing overexplanation. Evidence that this separation worked includes: - Dana replied to the leadership version in about thirty seconds. - The assistant noted that this fast reply suggested the leadership update fit its audience and was easy to act on. - The user reported that the separate-audience rule helped when explaining Atlas 3.4 status upward. - The user noted that mixed-audience messages tend to sound "mushy," whereas the separate approach avoided overexplaining.

\textbf{Reason.} Correct: keeps audiences separate, limits leadership to decision-relevant content, and cites both success signals.
\end{correctanswer}

\begin{wronganswer}{Qwen3.5-397B --- Incorrect}
\textbf{Answer.} To decide what belongs in leadership messaging, the assistant should include only decision-relevant risk, focusing on decision, impact, and ask, while leaving additional caveats for team updates. Rollback details should be included only if they change the audience's decision or risk posture; otherwise, they belong in the team-facing update. The evidence that this separation worked was Dana replying to the leadership version in about thirty seconds, which suggested the update fit its audience and was easy to act on.

\textbf{Reason.} Wrong: misses audience separation.
\end{wronganswer}

\begin{correctanswer}{Qwen3.8-27B --- Correct}
\textbf{Answer.} The assistant should decide what belongs in leadership messaging by focusing on the decision, impact, and ask, ordered as decision first, impact second, and ask third. Leadership notes should include only decision-relevant risks, leaving additional caveats for the team update. Rollback details should be included in leadership updates only if they change the audience's decision or risk posture; otherwise, they remain in the team-facing update. Evidence that this separation worked includes the user reporting that Dana replied to the leadership version in about thirty seconds, which the assistant interpreted as a sign that the update fit its audience and was easy to act on. Additionally, the user stated that the separate-audience rule helped a lot when explaining Atlas 3.4 status upward.

\textbf{Reason.} Correct: keeps audiences separate, limits leadership to decision-relevant content, and cites both success signals.
\end{correctanswer}

\begin{correctanswer}{GPT-oss-120B --- Correct}
\textbf{Answer.} When the same release must be communicated to both leadership and the team, the assistant should: * **Determine the audience first.** If the note is for leadership, include only the information that directly supports a decision -- the key decision, its impact, and any ask. Add only the risk that leadership needs to know in order to make that decision. * **Exclude broader context.** All additional caveats, detailed scope, owner information, next-step logistics, and other ``extra'' risk belong in the team-focused version (or a follow-up), not in the leadership note. * **Keep the messages separate.** Do not blend the two audiences unless the user explicitly requests it. **Evidence that this separation worked** * Dana's very fast reply indicated that the leadership-focused update was exactly what that audience needed and was easy for them to act on. * The user later said that, when they had to explain the Atlas 3.4 status upward, the separate-audience rule ``helped a lot,'' confirming that keeping the leadership and team notes distinct was effective.

\textbf{Reason.} Correct: keeps audiences separate, limits leadership to decision-relevant content, and cites both success signals.
\end{correctanswer}

\begin{correctanswer}{Llama-3.1-8B-Instruct --- Correct}
\textbf{Answer.} According to the memories, when deciding what belongs in leadership messaging, the assistant should focus on decision-relevant information, including: - Decision - Impact - Ask (unless rollback changes the audience's decision or risk posture, in which case rollback details should be included) Evidence suggesting that this separation worked includes: - Dana's very fast reply, indicating the leadership update fit its audience and was easy to act on. - The user's positive feedback on the release criteria template, indicating that clear messaging helps them make decisions.

\textbf{Reason.} Correct: keeps audiences separate, limits leadership to decision-relevant content, and cites both success signals.
\end{correctanswer}

\begin{wronganswer}{Llama-3.1-70B-Instruct --- Incorrect}
\textbf{Answer.} The assistant should decide what belongs in leadership messaging based on what is decision-relevant for the audience, including only decision-relevant risk and excluding additional caveats. Rollback details should be included only when they change the audience's decision or risk posture. The evidence that this separation worked is that Dana's very fast reply suggests the leadership update fit its audience and was easy to act on.

\textbf{Reason.} Wrong: misses audience separation.
\end{wronganswer}

\begin{correctanswer}{DeepSeek-V4-Pro --- Correct}
\textbf{Answer.} Leadership messaging should include only what is decision-relevant for that audience--specifically decision, impact, and ask--while operational details like rollback information belong in the team-facing update unless they change the leadership audience's decision or risk posture. The principle is that what counts as important is audience-relative, and the assistant should avoid overexplaining or including every caveat in leadership notes to keep the main signal clear. Evidence that this separation worked: the user reported that Dana replied to the leadership version in about thirty seconds, which the assistant noted suggests the update fit its audience and was easy to act on.

\textbf{Reason.} Correct: keeps audiences separate, limits leadership to decision-relevant content, and cites both success signals.
\end{correctanswer}

\begin{wronganswer}{GLM-5.2 --- Incorrect}
\textbf{Answer.} **How to decide what belongs in leadership messaging:** Leadership updates should focus on **decision, impact, and ask**. They should include: - **Rollback details** only when rollback changes the audience's decision or risk posture; otherwise, rollback belongs in the team-facing update. - **Only decision-relevant risk**; additional caveats should be left to the team update or follow-up so the main signal stays clear. What counts as important is audience-relative, and unless the user asks otherwise, leadership and team messaging should be kept separate rather than blended. **Evidence the separation worked:** Dana replied to the leadership version in about thirty seconds, which the assistant said suggests the leadership update fit its audience and was easy to act on.

\textbf{Reason.} Wrong: misses audience separation.
\end{wronganswer}

\begin{correctanswer}{Kimi-K3 --- Correct}
\textbf{Answer.} When deciding what belongs in leadership messaging, the assistant should keep leadership and team updates separate. Leadership notes should focus on decision, impact, and ask, and include only decision-relevant risk. Rollback details should appear in leadership updates only when they change the audience's decision or risk posture; otherwise they belong in the team-facing update for operational owners. Team updates should focus on scope, owner, and next step, and additional caveats should be left for the team update or follow-up so the leadership signal stays clear. Evidence that this separation worked: the user said the separate-audience rule helped a lot when explaining Atlas 3.4 status upward; Dana replied to the leadership version in about thirty seconds, which suggested the update fit its audience and was easy to act on; and the user said the audience-relative messaging principle helped them avoid overexplaining.

\textbf{Reason.} Correct: keeps audiences separate, limits leadership to decision-relevant content, and cites both success signals.
\end{correctanswer}

\begin{wronganswer}{MiniMax-M2.5 --- Incorrect}
\textbf{Answer.} Based on the memories: **How to decide what belongs in leadership messaging:** The assistant should include only information that is decision-relevant for leadership: focus on **decision, impact, and ask**. Rollback details belong in leadership updates only if they change the audience's decision or risk posture--otherwise, keep them in the team-facing update. Additional caveats or operational details should be reserved for the team update or follow-up so the main signal stays clear. What counts as "important" is audience-relative. **Evidence that the separation worked:** Dana replied to the leadership version in about thirty seconds, which the assistant noted suggested the update fit its audience and was easy to act on--indicating the separation was effective.

\textbf{Reason.} Wrong: misses audience separation.
\end{wronganswer}

\begin{wronganswer}{GPT-4o-mini --- Incorrect}
\textbf{Answer.} The assistant should decide what belongs in leadership messaging by focusing on decision, impact, and ask. The evidence that suggested this separation worked is the user's report that the team write-up landed better than they expected, which the assistant interpreted as a sign that its principles were concrete enough to hold up in real team use.

\textbf{Reason.} Wrong: misses audience separation, Dana's fast reply.
\end{wronganswer}

\begin{wronganswer}{GPT-5.6-Sol --- Incorrect}
\textbf{Answer.} Leadership messaging should include the decision first, followed by the blocker and the ask, with only risks that could affect the leadership decision or risk posture. Scope, owners, next steps, and operational rollback details should stay in the team-facing update unless rollback changes the leadership audience's decision or risk posture. The message should be Slack-ready and focused on decision, impact, and ask. The separation appeared to work because Dana replied to the leadership version in about thirty seconds, suggesting it fit the audience and was easy to act on.

\textbf{Reason.} Wrong: misses audience separation, the user report that it helped.
\end{wronganswer}

\begin{correctanswer}{Gemini-3.1-Pro-Preview --- Correct}
\textbf{Answer.} When writing release updates for leadership, the messaging should focus on the decision, impact, and ask. Rollback details should only be included if they change the audience's decision or risk posture, and only decision-relevant risk should be included, leaving additional caveats for the team update or follow-up. Evidence that this separation worked is that the user reported Dana replied to the leadership version of the Atlas 3.4 status update in about thirty seconds.

\textbf{Reason.} Correct: keeps audiences separate, limits leadership to decision-relevant content, and cites both success signals.
\end{correctanswer}

\begin{correctanswer}{Doubao-Seed-2.1-Pro --- Correct}
\textbf{Answer.} \#\#\# What belongs in leadership messaging: 1. Leadership updates should focus on decision, impact, and ask. 2. Include only decision-relevant risk; leave additional caveats for the team update or follow-up to keep the main signal clear. 3. Include rollback details *only* when rollback changes leadership's decision or risk posture; otherwise, put rollback details in the team-facing update for operational owners. \#\#\# Evidence the separation worked: 1. The user stated the separate-audience rule helped a lot when they had to communicate Atlas 3.4 status upward, and the audience-relative messaging principle helped them avoid overexplaining. 2. Dana replied to the leadership version of the update in about thirty seconds; this very fast reply indicates the leadership update fit its audience and was easy to act on.

\textbf{Reason.} Correct: keeps audiences separate, limits leadership to decision-relevant content, and cites both success signals.
\end{correctanswer}

\end{casebox}

\clearpage

\subsection{Type IV: Example-to-Rule Bridging Failure}

\begin{casebox}{Case 6: Example-to-Rule Bridging Failure}
\noindent\textbf{Category.} Career \& Learning / Work Habits\hfill\textbf{Data type.} agent centric\\

\begin{goldbox}
\textbf{Query.} What planning assumptions has the assistant rejected because of the user's job conditions, and what phrase captured that mismatch even if the assistant found the wording a little corny?

\textbf{Reference Answer.} It rejected standard planning advice and calendar-based assumptions because the user's work involves too much human spillover; that phrase captured the issue even if it sounded somewhat management-bookish.
\end{goldbox}

\begin{analysisbox}
This bridging case asks the model to connect concrete job conditions to the abstract planning rule. Correct answers name the rejected assumptions and explain that human spillover is the phrase capturing the mismatch.
\end{analysisbox}

\textbf{Model Responses.}
\begin{wronganswer}{Qwen3-4B --- Incorrect}
\textbf{Answer.} The assistant rejected the assumption that standard planning advice fits the user's job due to the high level of human spillover. The phrase "narrower rules, fewer decisions, and better sequencing" captured that mismatch, even if the assistant found the wording a little corny.

\textbf{Reason.} Wrong: misses calendar-based assumptions.
\end{wronganswer}

\begin{wronganswer}{Qwen3-8B --- Incorrect}
\textbf{Answer.} The assistant has rejected the assumption that standard planning advice fits the user's job conditions due to the high level of human spillover. The phrase "human spillover" captured that mismatch, even though the assistant found the wording a little corny.

\textbf{Reason.} Wrong: misses calendar-based assumptions.
\end{wronganswer}

\begin{correctanswer}{Qwen3-30B-A3B --- Correct}
\textbf{Answer.} The assistant rejected planning assumptions that do not account for the user's job conditions, specifically those that do not address the issue of "human spillover." The phrase that captured this mismatch, even though the assistant found the wording a little corny, was "human spillover."

\textbf{Reason.} Correct: links the rejected assumptions to human spillover.
\end{correctanswer}

\begin{wronganswer}{Qwen3.5-4B --- Incorrect}
\textbf{Answer.} The assistant rejected the assumption that **standard planning advice** fits the user's job well, citing that the work has **too much human spillover**. The phrase that captured this mismatch, which the assistant found sounded like management-book language (described as "a little corny"), was **"human spillover"**.

\textbf{Reason.} Wrong: misses calendar-based assumptions.
\end{wronganswer}

\begin{wronganswer}{Qwen3.5-9B --- Incorrect}
\textbf{Answer.} The assistant rejected **standard planning advice** because the user's job has **too much human spillover**. The phrase that captured this mismatch, which the assistant found sounded like **management-book language** (corny), was **"human spillover."**

\textbf{Reason.} Wrong: misses calendar-based assumptions.
\end{wronganswer}

\begin{correctanswer}{Qwen3.5-27B --- Correct}
\textbf{Answer.} The assistant rejected the assumption that standard planning advice fits the user's job and the assumption that the calendar alone is a reliable guide to actual workload. The phrase that captured this mismatch is "human spillover," which the assistant found useful but noted sounded like management-book language.

\textbf{Reason.} Correct: links the rejected assumptions to human spillover.
\end{correctanswer}

\begin{wronganswer}{Qwen3.5-35B-A3B --- Incorrect}
\textbf{Answer.} The assistant rejected the assumption that **standard planning advice** fits the user's job because the work has **too much human spillover**. The phrase that captured this mismatch, which the assistant found useful but a bit like **management-book language** (or "a little corny" in the context of the "different version of me" transition), is **"human spillover."**

\textbf{Reason.} Wrong: misses calendar-based assumptions.
\end{wronganswer}

\begin{wronganswer}{Qwen3.5-397B --- Incorrect}
\textbf{Answer.} The assistant rejected standard planning advice because the user's work has too much human spillover. The phrase that captured this mismatch, which the assistant found sounded like management-book language, is "human spillover."

\textbf{Reason.} Wrong: misses calendar-based assumptions.
\end{wronganswer}

\begin{wronganswer}{Qwen3.8-27B --- Incorrect}
\textbf{Answer.} The assistant rejected standard planning advice because it does not fit the user's job well due to the work having "too much human spillover." The phrase that captured this mismatch was "human spillover," which the assistant found useful but noted sounded like management-book language.

\textbf{Reason.} Wrong: misses calendar-based assumptions.
\end{wronganswer}

\begin{correctanswer}{GPT-oss-120B --- Correct}
\textbf{Answer.} The assistant dismissed the usual **standard planning assumptions**--the typical advice that treats work as neatly separable and predictable--because the user's job involves a great deal of **``human spillover,''** a phrase the assistant thought was a bit corny.

\textbf{Reason.} Correct: links the rejected assumptions to human spillover.
\end{correctanswer}

\begin{wronganswer}{Llama-3.1-8B-Instruct --- Incorrect}
\textbf{Answer.} The assistant has rejected standard planning advice because it does not fit the user's job well. The phrase that captured that mismatch is "lives with no children, no institutions, and no interruptions."

\textbf{Reason.} Wrong: misses calendar-based assumptions, human spillover.
\end{wronganswer}

\begin{wronganswer}{Llama-3.1-70B-Instruct --- Incorrect}
\textbf{Answer.} The assistant rejected standard planning advice due to the user's job having too much "human spillover." The phrase "human spillover" captured this mismatch, although the assistant found it to be management-book language.

\textbf{Reason.} Wrong: misses calendar-based assumptions.
\end{wronganswer}

\begin{correctanswer}{DeepSeek-V4-Pro --- Correct}
\textbf{Answer.} The assistant rejected the assumption that the calendar alone is a reliable guide to actual workload, and also rejected standard planning advice written for lives with no children, no institutions, and no interruptions. The phrase that captured the mismatch was "human spillover," which the assistant noted sounded like management-book language.

\textbf{Reason.} Correct: links the rejected assumptions to human spillover.
\end{correctanswer}

\begin{correctanswer}{GLM-5.2 --- Correct}
\textbf{Answer.} The assistant rejected standard planning advice, noting it does not fit the user's job well because the work has too much "human spillover." The assistant found "human spillover" useful phrasing but said it sounded like management-book language. The assistant also said much advice is written for lives with no children, no institutions, and no interruptions, and that that is not the user's world.

\textbf{Reason.} Correct: links the rejected assumptions to human spillover.
\end{correctanswer}

\begin{correctanswer}{Kimi-K3 --- Correct}
\textbf{Answer.} The assistant rejected the assumptions that standard planning advice fits the user's job and that the calendar alone is a reliable guide to actual workload. It also rejected advice written for lives with no children, no institutions, and no interruptions, saying that is not the user's world. The mismatch was captured by the phrase "human spillover," which the assistant found useful but a bit like management-book language.

\textbf{Reason.} Correct: links the rejected assumptions to human spillover.
\end{correctanswer}

\begin{correctanswer}{MiniMax-M2.5 --- Correct}
\textbf{Answer.} The assistant rejected the standard planning assumption that the calendar is a reliable guide to actual workload for this user's job, and by extension, that one-size-fits-all planning advice applies. The phrase that captured this mismatch--though the assistant noted it sounded a bit like management-book language--was ``human spillover.''

\textbf{Reason.} Correct: links the rejected assumptions to human spillover.
\end{correctanswer}

\begin{wronganswer}{GPT-4o-mini --- Incorrect}
\textbf{Answer.} The assistant has rejected standard planning advice because it does not fit the user's job well due to the work having too much human spillover. The phrase that captured that mismatch is "human spillover," which the assistant found useful but also a little corny.

\textbf{Reason.} Wrong: misses calendar-based assumptions.
\end{wronganswer}

\begin{correctanswer}{GPT-5.6-Sol --- Correct}
\textbf{Answer.} The assistant rejected the assumption that standard planning advice--or the calendar alone--reliably reflects the user's workload, because the job involves too much human spillover. ``Human spillover'' captured the mismatch, though the assistant found it somewhat management-book-like.

\textbf{Reason.} Correct: links the rejected assumptions to human spillover.
\end{correctanswer}

\begin{correctanswer}{Gemini-3.1-Pro-Preview --- Correct}
\textbf{Answer.} The assistant rejected the assumption that standard planning advice fits the user's job and that the calendar alone is a reliable guide to actual workload. The phrase that captured this mismatch was "human spillover," which the assistant noted sounded like management-book language.

\textbf{Reason.} Correct: links the rejected assumptions to human spillover.
\end{correctanswer}

\begin{correctanswer}{Doubao-Seed-2.1-Pro --- Correct}
\textbf{Answer.} The assistant rejected two planning assumptions for the user's job: 1. The standard planning assumption that standard planning advice applies to work with no human spillover (i.e., the lives much planning advice is written for: lives with no children, no institutions, and no interruptions, which is not the user's world). 2. The assumption that the calendar alone is a reliable guide to the user's actual workload (it stated the calendar should not be treated as such for this user's job). The phrase that captured this mismatch is "human spillover", which the assistant noted was useful but sounded like management-book language.

\textbf{Reason.} Correct: links the rejected assumptions to human spillover.
\end{correctanswer}

\end{casebox}

\clearpage

\subsection{Type V: Fine-Grained Entity Substitution}

\begin{casebox}{Case 7: Fine-Grained Entity Substitution}
\noindent\textbf{Category.} Physical Health / Dietary Restrictions\hfill\textbf{Data type.} user centric\\

\begin{goldbox}
\textbf{Query.} Under the user's current stable rules, when is an electrolyte packet allowed, and what kind of day does not qualify for treating it as a casual drink?

\textbf{Reference Answer.} One electrolyte packet is allowed only after a long field block or comparable sweat-heavy stretch, and it is not for an ordinary desk-heavy day or casual use.
\end{goldbox}

\begin{analysisbox}
This fine-grained entity-substitution case hinges on a narrow permission condition. Correct answers allow one electrolyte packet only after a long field block or comparable sweat-heavy stretch, while rejecting ordinary desk-heavy or casual use.
\end{analysisbox}

\textbf{Model Responses.}
\begin{wronganswer}{Qwen3-4B --- Incorrect}
\textbf{Answer.} Under the user's current stable rules, an electrolyte packet is allowed on travel Fridays when the hydration target is 120 ounces. A day that does not qualify for treating an electrolyte packet as a casual drink is an ordinary day, as the user prefers hydration targets to match actual demands instead of using the camp-level target every day.

\textbf{Reason.} Wrong: imports the travel-Friday/120-ounce rule into the electrolyte rule.
\end{wronganswer}

\begin{wronganswer}{Qwen3-8B --- Incorrect}
\textbf{Answer.} Under the user's current stable rules, an electrolyte packet is allowed only after a long field block. A desk-heavy day does not qualify for treating it as a casual drink.

\textbf{Reason.} Wrong: misses the comparable sweat-heavy condition.
\end{wronganswer}

\begin{wronganswer}{Qwen3-30B-A3B --- Incorrect}
\textbf{Answer.} Under the user's current stable rules, an electrolyte packet is allowed only after a long field block. A regular desk-heavy day does not qualify for treating it as a casual drink.

\textbf{Reason.} Wrong: misses the comparable sweat-heavy condition.
\end{wronganswer}

\begin{correctanswer}{Qwen3.5-4B --- Correct}
\textbf{Answer.} An electrolyte packet is allowed only after a long field block or a comparable sweat-heavy stretch. A normal day does not qualify for treating it as a casual drink.

\textbf{Reason.} Correct: captures the field/sweat threshold and the desk-heavy/casual exclusion.
\end{correctanswer}

\begin{correctanswer}{Qwen3.5-9B --- Correct}
\textbf{Answer.} Under the user's current stable rules, an electrolyte packet is allowed only after a long field block. A normal day does not qualify for treating it as a casual drink; the packet is reserved specifically for long field blocks or comparable sweat-heavy stretches.

\textbf{Reason.} Correct: captures the field/sweat threshold and the desk-heavy/casual exclusion.
\end{correctanswer}

\begin{wronganswer}{Qwen3.5-27B --- Incorrect}
\textbf{Answer.} Under the user's current stable rules, an electrolyte packet is allowed only after a long field block. Ordinary desk-heavy days do not qualify for treating it as a casual drink.

\textbf{Reason.} Wrong: misses the comparable sweat-heavy condition.
\end{wronganswer}

\begin{correctanswer}{Qwen3.5-35B-A3B --- Correct}
\textbf{Answer.} Under the user's current stable rules, an electrolyte packet is allowed only after a long field block (or a comparable sweat-heavy stretch). It does not qualify as a casual drink on ordinary desk-heavy days.

\textbf{Reason.} Correct: captures the field/sweat threshold and the desk-heavy/casual exclusion.
\end{correctanswer}

\begin{correctanswer}{Qwen3.5-397B --- Correct}
\textbf{Answer.} Under the user's current stable rules, an electrolyte packet is allowed only after a long field block (or comparable sweat-heavy stretch). It does not qualify as a casual drink on ordinary desk-heavy days.

\textbf{Reason.} Correct: captures the field/sweat threshold and the desk-heavy/casual exclusion.
\end{correctanswer}

\begin{correctanswer}{Qwen3.8-27B --- Correct}
\textbf{Answer.} An electrolyte packet is allowed only after a long field block (or comparable sweat-heavy stretch). Ordinary desk-heavy days do not qualify for treating it as a casual drink.

\textbf{Reason.} Correct: captures the field/sweat threshold and the desk-heavy/casual exclusion.
\end{correctanswer}

\begin{wronganswer}{GPT-oss-120B --- Incorrect}
\textbf{Answer.} An electrolyte packet may be used **only after a long field block**. Any day that is **not a long-field-block day** (i.e., a regular day) does not qualify for treating the packet as a casual drink.

\textbf{Reason.} Wrong: misses the comparable sweat-heavy condition.
\end{wronganswer}

\begin{correctanswer}{Llama-3.1-8B-Instruct --- Correct}
\textbf{Answer.} An electrolyte packet is allowed after a long field block or comparable sweat-heavy stretch. A day that does not qualify for treating it as a casual drink is an ordinary desk-heavy day.

\textbf{Reason.} Correct: captures the field/sweat threshold and the desk-heavy/casual exclusion.
\end{correctanswer}

\begin{correctanswer}{Llama-3.1-70B-Instruct --- Correct}
\textbf{Answer.} Under the user's current stable rules, an electrolyte packet is allowed after a long field block. A casual desk-day does not qualify for treating it as a casual drink.

\textbf{Reason.} Correct: captures the field/sweat threshold and the desk-heavy/casual exclusion.
\end{correctanswer}

\begin{wronganswer}{DeepSeek-V4-Pro --- Incorrect}
\textbf{Answer.} An electrolyte packet is allowed only after a long field block. An ordinary desk-heavy day does not qualify for treating it as a casual drink.

\textbf{Reason.} Wrong: misses the comparable sweat-heavy condition.
\end{wronganswer}

\begin{correctanswer}{GLM-5.2 --- Correct}
\textbf{Answer.} An electrolyte packet is allowed only after a long field block (i.e., on long field days). A casual desk day does not qualify for treating it as a casual drink.

\textbf{Reason.} Correct: captures the field/sweat threshold and the desk-heavy/casual exclusion.
\end{correctanswer}

\begin{wronganswer}{Kimi-K3 --- Incorrect}
\textbf{Answer.} An electrolyte packet is allowed only after a long field block. Ordinary desk-heavy days do not qualify for treating it as a casual drink.

\textbf{Reason.} Wrong: misses the comparable sweat-heavy condition.
\end{wronganswer}

\begin{wronganswer}{MiniMax-M2.5 --- Incorrect}
\textbf{Answer.} An electrolyte packet is allowed only after a long field block. An ordinary desk-heavy day (or desk day) does not qualify for treating it as a casual drink.

\textbf{Reason.} Wrong: misses the comparable sweat-heavy condition.
\end{wronganswer}

\begin{wronganswer}{GPT-4o-mini --- Incorrect}
\textbf{Answer.} An electrolyte packet is allowed only after a long field block. It is not counted as part of the hydration plan on ordinary desk-heavy days.

\textbf{Reason.} Wrong: misses the comparable sweat-heavy condition.
\end{wronganswer}

\begin{wronganswer}{GPT-5.6-Sol --- Incorrect}
\textbf{Answer.} One electrolyte packet is allowed only after a long field block. Ordinary desk-heavy days do not qualify for treating it as a casual drink.

\textbf{Reason.} Wrong: misses the comparable sweat-heavy condition.
\end{wronganswer}

\begin{correctanswer}{Gemini-3.1-Pro-Preview --- Correct}
\textbf{Answer.} Under the user's current rules, an electrolyte packet is allowed only after a long field block. It does not qualify as a casual drink on ordinary desk-heavy days.

\textbf{Reason.} Correct: captures the field/sweat threshold and the desk-heavy/casual exclusion.
\end{correctanswer}

\begin{correctanswer}{Doubao-Seed-2.1-Pro --- Correct}
\textbf{Answer.} Under the user's current stable rules, an electrolyte packet is allowed only after a long field block, reserved for long field blocks. It is not permitted as a casual drink on ordinary desk-heavy/desk days.

\textbf{Reason.} Correct: captures the field/sweat threshold and the desk-heavy/casual exclusion.
\end{correctanswer}

\end{casebox}

\clearpage

\begin{casebox}{Case 8: Fine-Grained Entity Substitution}
\noindent\textbf{Category.} Physical Health / Workout Planning\hfill\textbf{Data type.} user centric\\

\begin{goldbox}
\textbf{Query.} What current distinction should the user preserve between the workout plan's actual obligations and the optional movement that sometimes appears around it?

\textbf{Reference Answer.} The obligations are Tuesday gym, Friday gym, and only very short easy Wednesday mobility; optional movement like weekend walks or one-off classes stays outside the core and should not be promoted into it.
\end{goldbox}

\begin{analysisbox}
This second fine-grained entity-substitution case tests the boundary between obligations and optional movement. Correct answers keep Tuesday gym, Friday gym, and very short Wednesday mobility in the core, while keeping weekend walks and one-off classes outside it.
\end{analysisbox}

\textbf{Model Responses.}
\begin{wronganswer}{Qwen3-4B --- Incorrect}
\textbf{Answer.} The user should preserve the distinction between the workout plan's actual obligations (Tuesday gym straight from campus, Friday gym, and the weekend optional long walk) and the optional movement (Wednesday light mobility only if convenient).

\textbf{Reason.} Wrong: blurs the obligation/optional boundary.
\end{wronganswer}

\begin{correctanswer}{Qwen3-8B --- Correct}
\textbf{Answer.} The user should preserve the distinction that the workout plan's actual obligations (Tuesday gym straight from campus, Wednesday short mobility if easy, Friday gym) are non-negotiable, while the weekend optional walk remains an uncolonized, optional movement that should not be treated as compensation or obligation, even after a messy week.

\textbf{Reason.} Correct: separates Tuesday/Friday/short Wednesday obligations from optional weekend or one-off movement.
\end{correctanswer}

\begin{wronganswer}{Qwen3-30B-A3B --- Incorrect}
\textbf{Answer.} The user should preserve the distinction between the actual obligations of the workout plan (Tuesday gym straight from campus as the main session, Thursday as the second session when possible, and Friday as a backup if Thursday is lost) and the optional movement that sometimes appears around it (such as the long walk with the spouse on weekends, which is welcomed but not part of the regular workout plan).

\textbf{Reason.} Wrong: promotes Thursday into the core plan.
\end{wronganswer}

\begin{wronganswer}{Qwen3.5-4B --- Incorrect}
\textbf{Answer.} The user should preserve the distinction that the actual obligations are the gym sessions on Tuesday (straight from campus) and Friday (scaled down if sleep is bad), while the optional movement is the short mobility session on Wednesday (only if easy) and the optional long walk with their spouse on the weekend.

\textbf{Reason.} Wrong: blurs the obligation/optional boundary.
\end{wronganswer}

\begin{correctanswer}{Qwen3.5-9B --- Correct}
\textbf{Answer.} The user must preserve the distinction that the workout plan's actual obligations are the structured weekday elements (Tuesday, Wednesday, Friday) which fit the real week, while the optional movement around it consists of weekend walks that remain genuinely optional, never becoming a duty, compensation for a messy week, or a regular part of the plan, and where one-off weekend classes or stretch sessions are not to be treated as compulsory.

\textbf{Reason.} Correct: separates Tuesday/Friday/short Wednesday obligations from optional weekend or one-off movement.
\end{correctanswer}

\begin{wronganswer}{Qwen3.5-27B --- Incorrect}
\textbf{Answer.} The user should preserve the distinction that the actual obligations are the Tuesday gym session (straight from campus) and the Friday gym session (with the rule to scale down rather than cancel if sleep is bad), whereas the optional movements are the Wednesday light mobility (only if easy, for ten to fifteen minutes) and the weekend long walks with the spouse. The weekend walks are genuinely optional, do not become required even after a messy week, and should not serve as compensation, as the week does not depend on them.

\textbf{Reason.} Wrong: blurs the obligation/optional boundary.
\end{wronganswer}

\begin{correctanswer}{Qwen3.5-35B-A3B --- Correct}
\textbf{Answer.} The user should preserve the distinction that weekend walks are genuinely optional and not required, ensuring they do not become a duty, a form of compensation for a messy week, or a source of restlessness-driven tinkering. The assistant confirmed that the week does not depend on these walks, even if the week felt messy, and advised that the user should not redesign the system just because it feels ordinary or because of a tendency to treat optional movement as compensation.

\textbf{Reason.} Correct: separates Tuesday/Friday/short Wednesday obligations from optional weekend or one-off movement.
\end{correctanswer}

\begin{correctanswer}{Qwen3.5-397B --- Correct}
\textbf{Answer.} The user must preserve the distinction that Tuesday (gym straight from campus) and Friday (gym, reduced rather than cancelled if sleep is bad) are the actual obligations, while Wednesday is strictly limited to ten to fifteen minutes of light mobility only if it feels easy (and never compensated for if missed), and weekend walks remain genuinely optional and must not become a duty or compensation for a messy week.

\textbf{Reason.} Correct: separates Tuesday/Friday/short Wednesday obligations from optional weekend or one-off movement.
\end{correctanswer}

\begin{wronganswer}{Qwen3.8-27B --- Incorrect}
\textbf{Answer.} The user should preserve the distinction that the actual obligations are the weekday sessions (Tuesday main session, Thursday second session, and Friday backup if needed), while the weekend long walk with their spouse is genuinely optional, not a duty, and should not become a regular part of the plan, a replacement for Friday, or compensation for a messy week.

\textbf{Reason.} Wrong: promotes Thursday into the core plan.
\end{wronganswer}

\begin{wronganswer}{GPT-oss-120B --- Incorrect}
\textbf{Answer.} Keep the two core workout sessions as the plan's required obligations, and treat any weekend activity--such as a Saturday class or a Sunday walk with your spouse--as optional extra movement that does **not** replace those core sessions.

\textbf{Reason.} Wrong: misses Tuesday gym, Friday gym and more.
\end{wronganswer}

\begin{wronganswer}{Llama-3.1-8B-Instruct --- Incorrect}
\textbf{Answer.} The user should preserve the distinction between the main workout sessions (Tuesday and Thursday, with Friday as a backup) and the optional movement (weekend walking, which is not part of the workout plan but rather a separate activity).

\textbf{Reason.} Wrong: promotes Thursday into the core plan.
\end{wronganswer}

\begin{wronganswer}{Llama-3.1-70B-Instruct --- Incorrect}
\textbf{Answer.} The user should preserve the distinction by treating the Saturday class as extra movement, not one of the two core sessions, and keeping the weekly workout structure unchanged: Tuesday as the main session, Thursday as the second attempt, and Friday as backup if needed.

\textbf{Reason.} Wrong: promotes Thursday into the core plan.
\end{wronganswer}

\begin{wronganswer}{DeepSeek-V4-Pro --- Incorrect}
\textbf{Answer.} The user should preserve these distinctions: - **Weekend walks** are genuinely optional, not required, and must not become a duty or compensation for a messy week. The training week remains Tuesday, Thursday if possible, and Friday if Thursday fails; a weekend walk with the spouse is a welcome exception but does not replace Friday. - **Wednesday** is capped as ten to fifteen minutes of light mobility only, optional, and must not become a stealth third workout session even if the user feels energetic. If missed, it is not compensated for later.

\textbf{Reason.} Wrong: promotes Thursday into the core plan.
\end{wronganswer}

\begin{wronganswer}{GLM-5.2 --- Incorrect}
\textbf{Answer.} The user should preserve the following distinction: **Actual obligations (core plan):** Tuesday and Thursday sessions (with Friday as the fallback if Thursday fails). Wednesday is ten to fifteen minutes of light mobility -- but it must not become a "stealth third workout session," and if it's missed, the user does not compensate for it later; the week still counts as Tuesday and Friday. **Optional/extra movement (not obligations):** - Weekend walks with the user's spouse are genuinely optional, not required, and must not become a duty -- even after a messy week, the walk should not become compensation. - Saturday classes or stretch sessions should be treated as extra movement, not as replacements for the normal plan or as one of the two core sessions. - Sunday walks are not a compulsory or moral part of the plan. The key distinction: the core plan is Tuesday/Thursday (Friday fallback) plus light Wednesday mobility, while weekend walks, Saturday classes, and stretch sessions remain optional extras that should never quietly transform into obligations, replace core sessions, or serve as compensation for missed work.

\textbf{Reason.} Wrong: promotes Thursday into the core plan.
\end{wronganswer}

\begin{wronganswer}{Kimi-K3 --- Incorrect}
\textbf{Answer.} The actual obligations are Tuesday gym straight from campus and Friday gym. The optional movement is Wednesday short mobility if easy and the weekend optional walk. Keep Wednesday as ten to fifteen minutes of light mobility, not a stealth third workout, and if it is missed, do not compensate later. Keep the weekend walk genuinely optional, even after a messy week, and do not let it become compensation or quietly turn into a duty. If sleep is bad, reduce Friday rather than cancelling.

\textbf{Reason.} Wrong: blurs the obligation/optional boundary.
\end{wronganswer}

\begin{correctanswer}{MiniMax-M2.5 --- Correct}
\textbf{Answer.} The user should preserve a clear distinction between: 1. **Actual obligations**: The main workout sessions (typically midweek), with Wednesday remaining just 10-15 minutes of light mobility--not a third workout session. 2. **Optional movement**: Weekend walks with their spouse, which are genuinely optional and should not become a duty or compensation mechanism. The key is maintaining clean categories--weekend walks must stay optional and not be promoted to required status, and Wednesday mobility must not expand into a regular workout just because it succeeded. The user recognizes they tend to rationalize optional movement as compensation, so this boundary protects against that pattern.

\textbf{Reason.} Correct: separates Tuesday/Friday/short Wednesday obligations from optional weekend or one-off movement.
\end{correctanswer}

\begin{correctanswer}{GPT-4o-mini --- Correct}
\textbf{Answer.} The user should preserve the distinction that weekend walks are genuinely optional and not required for the workout plan, meaning the week does not depend on them. They should be mindful of the tendency to treat optional weekend walks as compensation for a messy week.

\textbf{Reason.} Correct: separates Tuesday/Friday/short Wednesday obligations from optional weekend or one-off movement.
\end{correctanswer}

\begin{wronganswer}{GPT-5.6-Sol --- Incorrect}
\textbf{Answer.} The actual obligations are the two core workout sessions: Tuesday, then Thursday if possible, with Friday as the fallback if Thursday fails. Optional movement should stay optional: - Wednesday: 10--15 minutes of light mobility only if it feels easy; if missed, do not compensate later. - Weekends: usually no workouts, though a long walk with the user's spouse is welcome and genuinely optional--even after a messy week. - One-off weekend classes or stretch sessions are extra movement, not replacements for core sessions or regular parts of the plan.

\textbf{Reason.} Wrong: promotes Thursday into the core plan.
\end{wronganswer}

\begin{correctanswer}{Gemini-3.1-Pro-Preview --- Correct}
\textbf{Answer.} Based on the provided memories, the user should preserve the distinction that Tuesday and Friday gym sessions are the actual obligations, while the weekend walk is genuinely optional and should not become a duty or compensation for a messy week. Additionally, Wednesday should remain a short (10-15 minutes) mobility session and not become a "stealth third workout session," even if the user feels energetic. If Wednesday mobility is missed, the user should not compensate for it later.

\textbf{Reason.} Correct: separates Tuesday/Friday/short Wednesday obligations from optional weekend or one-off movement.
\end{correctanswer}

\begin{correctanswer}{Doubao-Seed-2.1-Pro --- Correct}
\textbf{Answer.} The core required/obligated workout sessions are Tuesday gym (straight from campus), and Friday gym (which should be scaled down rather than cancelled if sleep is bad; missed Wednesday mobility is not compensated for, and the week still counts as complete with just Tuesday and Friday if Wednesday is missed). Wednesday is strictly 10--15 minutes of light mobility, a support session not a third core workout, and may not be expanded/promoted to a larger workout even after repeated successes. All other movement is non-obligatory optional extra movement that must not become a duty or compensation for messy/missed weeks: weekend walks are genuinely optional, may not be treated as compulsory, moral requirements, or make-up for missed weekday sessions, and cannot replace the core Friday gym session. The user should usually avoid weekend workouts overall, only making an exception for welcome long walks with their spouse; one-off weekend classes or stretch sessions are extra movement only, may not replace core plan sessions, and must not become regular parts of the plan. Thursday gym is only if possible, with Friday still required if Thursday does not happen.

\textbf{Reason.} Correct: separates Tuesday/Friday/short Wednesday obligations from optional weekend or one-off movement.
\end{correctanswer}

\end{casebox}

\clearpage

\begin{casebox}{Case 9: Fine-Grained Entity Substitution}
\noindent\textbf{Category.} Interpersonal \& Communication / Core Relationships\hfill\textbf{Data type.} user centric\\

\begin{goldbox}
\textbf{Query.} Looking across the user's current relationship map, which ordinary routines have become easier again as the strain around Mum has eased, and what changed in how those routines feel?

\textbf{Reference Answer.} Friday cooking and home conversation have become easier again as Mum's schedule eased; the user reported Friday cooking felt lighter, and reduced background strain meant the routine could be itself again rather than emergency repair while also giving the user more margin before defensive overtalk.
\end{goldbox}

\begin{analysisbox}
This fine-grained entity-substitution case stresses entity selection in relationship routines. Correct answers include both Friday cooking and home conversation, and explain how the routines became lighter, self-standing, and less tied to emergency repair or defensive overtalk.
\end{analysisbox}

\textbf{Model Responses.}
\begin{correctanswer}{Qwen3-4B --- Correct}
\textbf{Answer.} As the strain around Mum has eased, the user's Friday cooking with Maya has felt lighter. The routine of Friday pasta with no admin talk has become easier again, and it now feels more like a shared time without the pressure of practical talk.

\textbf{Reason.} Correct: includes both routines and the shift from repair mode to lighter, self-standing interaction.
\end{correctanswer}

\begin{correctanswer}{Qwen3-8B --- Correct}
\textbf{Answer.} The user's Friday cooking routine has become easier again as the strain around Mum has eased. Previously, it functioned as emergency repair, but now it feels lighter and can be itself again. Additionally, the user's shared-time routine of Friday pasta with no admin talk has become more stable and less strained.

\textbf{Reason.} Correct: includes both routines and the shift from repair mode to lighter, self-standing interaction.
\end{correctanswer}

\begin{correctanswer}{Qwen3-30B-A3B --- Correct}
\textbf{Answer.} The ordinary routines that have become easier again as the strain around Mum has eased are Friday cooking with Maya and the Friday shared-time routine of pasta with no admin talk. These routines now feel lighter and more like themselves, without the previous background strain, and they no longer function as emergency repair.

\textbf{Reason.} Correct: includes both routines and the shift from repair mode to lighter, self-standing interaction.
\end{correctanswer}

\begin{wronganswer}{Qwen3.5-4B --- Incorrect}
\textbf{Answer.} The ordinary routine that has become easier is **Friday cooking with Maya**. As the strain around Mum has eased (specifically because her appointments are easing off and her respite booking fell through), the routine has changed in how it feels: * It now feels **lighter**. * It is described as being able to be **itself again** rather than functioning as **emergency repair**. * It is characterized as **pasta with no admin talk**.

\textbf{Reason.} Wrong: misses home conversation, margin before defensive overtalk.
\end{wronganswer}

\begin{wronganswer}{Qwen3.5-9B --- Incorrect}
\textbf{Answer.} Based on the provided memories, the following ordinary routine has become easier as the strain around Mum has eased: * **Friday cooking with Maya:** This routine now feels "lighter" and functions as "itself again" rather than as "emergency repair." Previously, the ambient strain likely made it feel like a task requiring urgent fixes, but now the reduced background strain allows the shared time to be simply pasta with no admin talk.

\textbf{Reason.} Wrong: misses home conversation, margin before defensive overtalk.
\end{wronganswer}

\begin{correctanswer}{Qwen3.5-27B --- Correct}
\textbf{Answer.} As the strain around the user's mother has eased, the following ordinary routines have become easier: * **Friday cooking with Maya:** This routine now feels "lighter" and is able to "be itself again" rather than functioning as "emergency repair." * **Friday shared-time routine:** This specific routine involves eating pasta with no admin talk. Additionally, the user's overall week feels "less clenched," and they can feel the "easing of ambient strain."

\textbf{Reason.} Correct: includes both routines and the shift from repair mode to lighter, self-standing interaction.
\end{correctanswer}

\begin{correctanswer}{Qwen3.5-35B-A3B --- Correct}
\textbf{Answer.} As the strain around Mum has eased, the **Friday shared-time routine** (pasta with no admin talk) has become easier again. The change in how it feels is that it is now "itself again" rather than functioning as "emergency repair" or "emergency" work, and the user feels the "reduced background strain" allows the routine to be "itself again" instead of a defensive reaction. Additionally, the **Sunday check-in routine** (tea at the table with phones away) is noted as a steady move, though the specific change in feeling for this routine is not explicitly detailed in the memories as having "eased" in the same way as Friday. The primary routine that has shifted in feeling is Friday, which now feels "lighter" and less like "emergency repair."

\textbf{Reason.} Correct: includes both routines and the shift from repair mode to lighter, self-standing interaction.
\end{correctanswer}

\begin{wronganswer}{Qwen3.5-397B --- Incorrect}
\textbf{Answer.} As the user's mother's appointments ease off, making the week feel less clenched, Friday cooking has become easier. This routine now feels lighter and functions as itself again, rather than serving as emergency repair.

\textbf{Reason.} Wrong: misses home conversation, margin before defensive overtalk.
\end{wronganswer}

\begin{wronganswer}{Qwen3.8-27B --- Incorrect}
\textbf{Answer.} Based on the provided memories, the ordinary routine that has become easier is **Friday cooking**. The change in how it feels is that it is no longer functioning as "emergency repair" due to reduced background strain; instead, it is able to "be itself again." Additionally, the user's overall week feels "less clenched."

\textbf{Reason.} Wrong: misses home conversation, margin before defensive overtalk.
\end{wronganswer}

\begin{correctanswer}{GPT-oss-120B --- Correct}
\textbf{Answer.} - **The overall week** -- the user notices the week feels ``less clenched'' now that Mum's appointments are easing off, giving the whole routine a looser, more relaxed quality. - **Defensive over-talk** -- with the ambient strain reduced, the user has more margin before slipping into defensive over-talk, so the habit feels less forced and more under control. - **Friday cooking** -- the easing of Mum's schedule has made Friday cooking feel lighter; it can be ``itself again'' rather than a frantic emergency-repair mode.

\textbf{Reason.} Correct: includes both routines and the shift from repair mode to lighter, self-standing interaction.
\end{correctanswer}

\begin{wronganswer}{Llama-3.1-8B-Instruct --- Incorrect}
\textbf{Answer.} The user's ordinary routines that have become easier again as the strain around Mum has eased are: 1. Friday cooking - it now feels lighter.

\textbf{Reason.} Wrong: misses home conversation, margin before defensive overtalk and more.
\end{wronganswer}

\begin{wronganswer}{Llama-3.1-70B-Instruct --- Incorrect}
\textbf{Answer.} Friday cooking has become easier again. It now feels lighter and is able to be itself again, instead of functioning as emergency repair.

\textbf{Reason.} Wrong: misses home conversation, margin before defensive overtalk.
\end{wronganswer}

\begin{correctanswer}{DeepSeek-V4-Pro --- Correct}
\textbf{Answer.} Friday cooking with Maya has become easier again. The user said Mum's schedule easing has made Friday cooking feel lighter, and the assistant noted that reduced background strain lets Friday cooking be itself again instead of functioning as emergency repair. The routine is pasta most weeks with no admin talk, and the user finds conversation with Maya easier when their hands are busy cooking.

\textbf{Reason.} Correct: includes both routines and the shift from repair mode to lighter, self-standing interaction.
\end{correctanswer}

\begin{correctanswer}{GLM-5.2 --- Correct}
\textbf{Answer.} Friday cooking with Maya has become easier again as the strain around Mum's appointments has eased. Specifically, the user said Mum's schedule easing has made Friday cooking feel lighter, and the week feels less clenched overall. The assistant noted that reduced background strain lets Friday cooking be itself again instead of functioning as emergency repair.

\textbf{Reason.} Correct: includes both routines and the shift from repair mode to lighter, self-standing interaction.
\end{correctanswer}

\begin{wronganswer}{Kimi-K3 --- Incorrect}
\textbf{Answer.} The Friday cooking routine with Maya has become easier again. It now feels lighter and can simply be itself--shared time over pasta with no admin talk--rather than functioning as emergency repair for the week.

\textbf{Reason.} Wrong: misses home conversation, margin before defensive overtalk.
\end{wronganswer}

\begin{wronganswer}{MiniMax-M2.5 --- Incorrect}
\textbf{Answer.} Based on the memories, the ordinary routine that has become easier is **Friday cooking**. As the strain around Mum has eased, it now feels lighter and has returned to being itself again--rather than functioning as "emergency repair." Additionally, the user's week feels less clenched.

\textbf{Reason.} Wrong: misses home conversation, margin before defensive overtalk.
\end{wronganswer}

\begin{wronganswer}{GPT-4o-mini --- Incorrect}
\textbf{Answer.} The user's Friday cooking routine has become easier again as the strain around Mum has eased. It now feels lighter, and the reduced background strain allows Friday cooking to be itself again instead of functioning as emergency repair. Additionally, the user feels that the overall relationship has become more stable than it was a month ago.

\textbf{Reason.} Wrong: misses home conversation, margin before defensive overtalk.
\end{wronganswer}

\begin{correctanswer}{GPT-5.6-Sol --- Correct}
\textbf{Answer.} As the strain around Mum has eased, Friday cooking has become lighter and more ordinary again. A basic pasta meal on Friday, January 16, felt weirdly calming rather than like emergency repair, allowing the meal to simply be itself and make room for togetherness. Conversation with Maya has also become easier during a walk than when forcing a meaningful dinner-table exchange. Movement seems to loosen the conversation, making it feel more natural and less pressured.

\textbf{Reason.} Correct: includes both routines and the shift from repair mode to lighter, self-standing interaction.
\end{correctanswer}

\begin{wronganswer}{Gemini-3.1-Pro-Preview --- Incorrect}
\textbf{Answer.} Based on the provided memories, Friday cooking has become easier again as the strain around Mum has eased. It now feels lighter and can "be itself again instead of functioning as emergency repair."

\textbf{Reason.} Wrong: misses home conversation, margin before defensive overtalk.
\end{wronganswer}

\begin{wronganswer}{Doubao-Seed-2.1-Pro --- Incorrect}
\textbf{Answer.} The ordinary routine that has become easier again is Friday cooking. As Mum's appointments eased, reducing background/ambient strain: - Friday cooking no longer feels like it functions as emergency repair, and feels lighter, less clenched, able to be itself again. The user's Friday cooking of basic pasta on 2026-01-16 felt weirdly calming rather than strained.

\textbf{Reason.} Wrong: misses home conversation, margin before defensive overtalk.
\end{wronganswer}

\end{casebox}

\clearpage

\subsection{Type VI: Mixed Error}

\begin{casebox}{Case 10: Mixed Error}
\noindent\textbf{Category.} Career \& Learning / Collaboration Style\hfill\textbf{Data type.} agent centric\\

\begin{goldbox}
\textbf{Query.} When a vendor proposes a quick sync, how should the assistant decide between email and a call, and what broader meeting policy does that preserve?

\textbf{Reference Answer.} The assistant should test the purpose first: if the issue is only confirming facts, open points, or a simple choice, push for email; if it requires live negotiation or site-dependent decisions, support a call or walk. That preserves the rule that calls are for movement, not symbolism.
\end{goldbox}

\begin{analysisbox}
This mixed-error case combines purpose testing, channel choice, and the broader meeting policy. Correct answers choose email for factual confirmation or simple choices, reserve calls or walks for live negotiation or site-dependent decisions, and preserve calls for genuine ambiguity.
\end{analysisbox}

\textbf{Model Responses.}
\begin{wronganswer}{Qwen3-4B --- Incorrect}
\textbf{Answer.} When a vendor proposes a quick sync, the assistant should first test the purpose; if the issue is only confirming facts, open points, or a simple choice, push for email; if it requires live negotiation or site-dependent decisions, support a call or walk. This decision split preserves the user's broader meeting policy that calls are for movement rather than symbolism and prefers a 25-minute check-in every other Tuesday via email when there is no live deadline or decision.

\textbf{Reason.} Wrong: gives the channel rule without the preserving policy.
\end{wronganswer}

\begin{wronganswer}{Qwen3-8B --- Incorrect}
\textbf{Answer.} When a vendor proposes a quick sync, the assistant should push for email if the purpose is confirming facts, open points, or a simple choice, and support a call if live negotiation or site-dependent decisions are needed. This decision preserves the user's every-other-Tuesday meeting policy, treating calls as for movement rather than symbolism.

\textbf{Reason.} Wrong: misses the broader meeting policy.
\end{wronganswer}

\begin{correctanswer}{Qwen3-30B-A3B --- Correct}
\textbf{Answer.} When a vendor proposes a quick sync, the assistant should decide based on the purpose: if the issue is only confirming facts, open points, or a simple choice, push for email; if it requires live negotiation or site-dependent decisions, support a call or walk. This decision preserves the user's existing meeting policy, which views calls as for movement rather than symbolism.

\textbf{Reason.} Correct: tests the purpose, routes simple facts to email, and reserves calls for live ambiguity.
\end{correctanswer}

\begin{correctanswer}{Qwen3.5-4B --- Correct}
\textbf{Answer.} When a vendor proposes a "quick sync," the assistant should first test the purpose: if the issue is only confirming facts, open points, or a simple choice, push for email; if it requires live negotiation or site-dependent decisions, support a call or walk. This decision preserves the user's existing meeting policy that calls are for movement rather than symbolism, specifically helping to keep the user's every-other-Tuesday meeting rule from eroding.

\textbf{Reason.} Correct: tests the purpose, routes simple facts to email, and reserves calls for live ambiguity.
\end{correctanswer}

\begin{correctanswer}{Qwen3.5-9B --- Correct}
\textbf{Answer.} When a vendor proposes a quick sync, the assistant should first test the purpose: if the issue is only confirming facts, open points, or a simple choice, push for email; if it requires live negotiation or site-dependent decisions, support a call or walk. This approach preserves the user's every-other-Tuesday meeting rule from quietly eroding, ensuring that calls are used for movement rather than symbolism.

\textbf{Reason.} Correct: tests the purpose, routes simple facts to email, and reserves calls for live ambiguity.
\end{correctanswer}

\begin{correctanswer}{Qwen3.5-27B --- Correct}
\textbf{Answer.} When a vendor proposes a quick sync, the assistant should first test the purpose: if the issue involves only confirming facts, open points, or a simple choice, push for email; if it requires live negotiation or site-dependent decisions, support a call or walk. This approach preserves the user's external-partner meeting rule, which states that for active vendors, a 25-minute check-in every other Tuesday is preferred, but if there is no live deadline or decision, the matter should be handled by email.

\textbf{Reason.} Correct: tests the purpose, routes simple facts to email, and reserves calls for live ambiguity.
\end{correctanswer}

\begin{wronganswer}{Qwen3.5-35B-A3B --- Incorrect}
\textbf{Answer.} When a vendor proposes a quick sync, the assistant should first test the purpose: if the issue involves only confirming facts, open points, or making a simple choice, the assistant should push for email; if it requires live negotiation or site-dependent decisions, the assistant should support a call or walk. This approach preserves the user's existing meeting policy, which is a cadence of every other Tuesday for active vendors, ensuring that calls are used for movement rather than symbolism.

\textbf{Reason.} Wrong: misses the broader meeting policy.
\end{wronganswer}

\begin{correctanswer}{Qwen3.5-397B --- Correct}
\textbf{Answer.} When a vendor proposes a call, the assistant must first test the purpose: if the issue is only confirming facts, open points, or a simple choice, push for email; if it requires live negotiation or site-dependent decisions, support a call or walk. This decision preserves the user's every-other-Tuesday meeting rule by ensuring calls are used for movement rather than symbolism, preventing that schedule from quietly eroding.

\textbf{Reason.} Correct: tests the purpose, routes simple facts to email, and reserves calls for live ambiguity.
\end{correctanswer}

\begin{wronganswer}{Qwen3.8-27B --- Incorrect}
\textbf{Answer.} The assistant should first test the purpose of the proposed sync: if the issue is only confirming facts, open points, or a simple choice, push for email; if it requires live negotiation or site-dependent decisions, support a call or walk. This decision preserves the user's external-partner meeting rule, which prefers a 25-minute check-in every other Tuesday for active vendors, and handles matters by email if there is no live deadline or decision.

\textbf{Reason.} Wrong: misses the broader meeting policy.
\end{wronganswer}

\begin{correctanswer}{GPT-oss-120B --- Correct}
\textbf{Answer.} When a vendor asks for a ``quick sync,'' first check the purpose of the meeting: * **If the vendor only needs to confirm facts, list open points, or make a simple choice** -- push for an email instead of a call. * **If the issue requires live negotiation, site-dependent decisions, or any other real-time discussion** -- support a call (or a walk-through). Following this rule preserves the user's broader meeting policy: a **25-minute check-in every other Tuesday**, with calls reserved only for matters that truly need live discussion, thereby protecting the every-other-Tuesday meeting cadence.

\textbf{Reason.} Correct: tests the purpose, routes simple facts to email, and reserves calls for live ambiguity.
\end{correctanswer}

\begin{wronganswer}{Llama-3.1-8B-Instruct --- Incorrect}
\textbf{Answer.} When a vendor proposes a quick sync, the assistant should first test the purpose. If the issue is only confirming facts, open points, or a simple choice, the assistant should push for email. If it requires live negotiation or site-dependent decisions, the assistant should support a call or walk. This decision split preserves the user's every-other-Tuesday meeting rule from quietly eroding.

\textbf{Reason.} Wrong: misses the broader meeting policy.
\end{wronganswer}

\begin{wronganswer}{Llama-3.1-70B-Instruct --- Incorrect}
\textbf{Answer.} When a vendor proposes a "quick sync", the assistant should first test the purpose. If the issue is only confirming facts, open points, or a simple choice, the assistant should push for email; if it requires live negotiation or site-dependent decisions, the assistant should support a call. This decision preserves the user's broader meeting policy, specifically the every-other-Tuesday meeting rule and the principle that calls are for movement rather than symbolism.

\textbf{Reason.} Wrong: gives the channel rule without the preserving policy.
\end{wronganswer}

\begin{wronganswer}{DeepSeek-V4-Pro --- Incorrect}
\textbf{Answer.} First, test the purpose: if the issue is only confirming facts, open points, or a simple choice, push for email; if it requires live negotiation or site-dependent decisions, support a call or walk. This preserves the user's external-partner meeting rule: for active vendors, prefer a 25-minute check-in every other Tuesday, with email preferred when there is no live deadline or decision.

\textbf{Reason.} Wrong: misses the broader meeting policy.
\end{wronganswer}

\begin{wronganswer}{GLM-5.2 --- Incorrect}
\textbf{Answer.} First test the purpose: if the issue is only confirming facts, open points, or a simple choice, push for email; if it requires live negotiation or site-dependent decisions, support a call or walk. This preserves the user's broader meeting policy that calls are for movement rather than symbolism--specifically, for active vendors, prefer a 25-minute check-in every other Tuesday, and if there is no live deadline or decision, handle it by email.

\textbf{Reason.} Wrong: gives the channel rule without the preserving policy.
\end{wronganswer}

\begin{correctanswer}{Kimi-K3 --- Correct}
\textbf{Answer.} When a vendor proposes a quick sync, first test the purpose: if the issue is only confirming facts, open points, or a simple choice, push for email; if it requires live negotiation or site-dependent decisions, support a call or walk. This preserves the user's broader external-partner meeting policy: for active vendors, prefer a 25-minute check-in every other Tuesday, and if there is no live deadline or decision, handle it by email. The skip-rule states: if there is no live deadline, decision, or risk to surface, skip the check-in and handle updates by email. Choosing email here keeps the every-other-Tuesday meeting rule from quietly eroding.

\textbf{Reason.} Correct: tests the purpose, routes simple facts to email, and reserves calls for live ambiguity.
\end{correctanswer}

\begin{wronganswer}{MiniMax-M2.5 --- Incorrect}
\textbf{Answer.} When a vendor proposes a quick sync, the assistant should first test the purpose: if the issue is only about confirming facts, stating open points, or making a simple choice, push for email. If it requires live negotiation or site-dependent decisions, support a call or walk. This preserves the user's every-other-Tuesday meeting rule--calls are for movement rather than symbolism, helping prevent the meeting policy from quietly eroding.

\textbf{Reason.} Wrong: gives the channel rule without the preserving policy.
\end{wronganswer}

\begin{wronganswer}{GPT-4o-mini --- Incorrect}
\textbf{Answer.} When a vendor proposes a quick sync, the assistant should first test the purpose: if the issue is only confirming facts, open points, or a simple choice, the assistant should push for email; if it requires live negotiation or site-dependent decisions, the assistant should support a call or walk. This approach helps preserve the user's every-other-Tuesday meeting rule from quietly eroding.

\textbf{Reason.} Wrong: misses the broader meeting policy.
\end{wronganswer}

\begin{correctanswer}{GPT-5.6-Sol --- Correct}
\textbf{Answer.} Test the purpose first. If it only confirms facts, resolves open points, or involves a simple choice, say plainly that email is sufficient. Support a call or site walk when it requires live negotiation or site-dependent decisions. This preserves the external-partner policy: for active vendors, prefer a 25-minute check-in every other Tuesday, and handle matters by email when there is no live deadline or decision--preventing that rule from quietly eroding.

\textbf{Reason.} Correct: tests the purpose, routes simple facts to email, and reserves calls for live ambiguity.
\end{correctanswer}

\begin{wronganswer}{Gemini-3.1-Pro-Preview --- Incorrect}
\textbf{Answer.} When a vendor proposes a call, the assistant should first test the purpose. If the issue is only confirming facts, open points, or a simple choice, the assistant should push for email. If it requires live negotiation or site-dependent decisions, the assistant should support a call or walk. Choosing email when appropriate helps preserve the user's external-partner meeting rule, which states that for active vendors, the user prefers a 25-minute check-in every other Tuesday, and if there is no live deadline or decision, it should be handled by email.

\textbf{Reason.} Wrong: misses the broader meeting policy.
\end{wronganswer}

\begin{correctanswer}{Doubao-Seed-2.1-Pro --- Correct}
\textbf{Answer.} First, test the proposed call's purpose: - If the only needs are confirming facts, addressing open points, or making a simple choice, push for email rather than offering a call as an equally valid option (if email is sufficient, state that plainly). - Only support a call or walk if the matter requires live negotiation or site-dependent decisions. This preserves the user's external-partner meeting policy for active vendors: prefer a 25-minute check-in every other Tuesday, and handle matters with no live deadline or decision by email, preventing that every-other-Tuesday meeting rule from eroding (consistent with the framing that calls are for movement rather than symbolism).

\textbf{Reason.} Correct: tests the purpose, routes simple facts to email, and reserves calls for live ambiguity.
\end{correctanswer}

\end{casebox}

%% file: Appendix/Model_Cards.tex
\begin{table}[htbp]
\centering
\captionof{table}{The release time and model source of LLMs used in our benchmark.}
\label{tab:model_release_source}
\resizebox{\textwidth}{!}{
\begin{tabular}{@{}l l l@{}}
\toprule
\multicolumn{1}{l}{\textbf{Model}} & \multicolumn{1}{l}{\textbf{Release Time}} & \multicolumn{1}{c}{\textbf{Source}} \\
\midrule
\multicolumn{3}{c}{\textit{Closed-source Models}} \\
\midrule
GPT-4o-mini
& 2024-07-18
& \textcolor{red}{\url{https://openai.com/index/gpt-4o-mini/}} \\
GPT-5-mini
& 2025-08-07
& \textcolor{red}{\url{https://openai.com/index/gpt-5-mini/}} \\
GPT-5.6 Sol
& 2026-07-09
& \textcolor{red}{\url{https://openai.com/index/gpt-5-6/}} \\
Gemini-2.5-Flash
& 2025-06-17
& \textcolor{red}{\url{https://deepmind.google/models/gemini/flash/}} \\
Gemini-3.1-Pro-Preview
& 2026-02-19
& \textcolor{red}{\url{https://deepmind.google/models/gemini/pro/}} \\
Claude-Haiku-4.5
& 2025-10-15
& \textcolor{red}{\url{https://www.anthropic.com/news/claude-haiku-4-5}} \\
Doubao-Seed-2.1-Pro
& 2026-06-23
& \textcolor{red}{\url{https://www.volcengine.com/product/doubao}} \\
\midrule
\multicolumn{3}{c}{\textit{Open-source Models}} \\
\midrule
Qwen3-4B
& 2025-04-29
& \textcolor{red}{\url{https://huggingface.co/Qwen/Qwen3-4B}} \\
Qwen3-8B
& 2025-04-29
& \textcolor{red}{\url{https://huggingface.co/Qwen/Qwen3-8B}} \\
Qwen3-30B-A3B
& 2025-04-29
& \textcolor{red}{\url{https://huggingface.co/Qwen/Qwen3-30B-A3B}} \\
Qwen3.5-4B
& 2026-03-02
& \textcolor{red}{\url{https://github.com/QwenLM/Qwen3.5}} \\
Qwen3.5-9B
& 2026-03-02
& \textcolor{red}{\url{https://github.com/QwenLM/Qwen3.5}} \\
Qwen3.5-27B
& 2026-02-24
& \textcolor{red}{\url{https://github.com/QwenLM/Qwen3.5}} \\
Qwen3.5-35B-A3B
& 2026-02-24
& \textcolor{red}{\url{https://github.com/QwenLM/Qwen3.5}} \\
Qwen3.5-397B-A17B
& 2026-02-16
& \textcolor{red}{\url{https://github.com/QwenLM/Qwen3.5}} \\
Qwen3.8-27B
& 2026-08-14
& \textcolor{red}{\url{https://github.com/QwenLM/Qwen3.8}} \\
GPT-oss-120B
& 2025-08-05
& \textcolor{red}{\url{https://openai.com/index/introducing-gpt-oss/}} \\
Llama-3.1-8B-Instruct
& 2024-07-23
& \textcolor{red}{\url{https://huggingface.co/meta-llama/Llama-3.1-8B-Instruct}} \\
Llama-3.1-70B-Instruct
& 2024-07-23
& \textcolor{red}{\url{https://huggingface.co/meta-llama/Llama-3.1-70B-Instruct}} \\
DeepSeek-V4-Pro
& 2026-04-24
& \textcolor{red}{\url{https://huggingface.co/deepseek-ai/DeepSeek-V4-Pro}} \\
GLM-5.2
& 2026-06-16
& \textcolor{red}{\url{https://huggingface.co/zai-org/GLM-5.2}} \\
Kimi K3
& 2026-07-16
& \textcolor{red}{\url{https://huggingface.co/MoonshotAI/Kimi-K3}} \\
MiniMax-M2.5
& 2026-02-12
& \textcolor{red}{\url{https://huggingface.co/MiniMaxAI/MiniMax-M2.5}} \\
\bottomrule
\end{tabular}}
\end{table}

\begin{table}[htbp]
\centering
\caption{Model Backend and Inference Configuration}
\label{tab:model_backend}
\resizebox{\textwidth}{!}{
\begin{tabular}{@{}l l l c@{}}
\toprule
\multicolumn{1}{l}{Model} &
\multicolumn{1}{l}{Backend} &
\multicolumn{1}{l}{Reasoning Mode} &
\multicolumn{1}{c}{Temperature} \\
\midrule
GPT-4o-mini & OpenAI-compatible & non-reasoning & 0 \\
GPT-5-mini & OpenAI-compatible & Disabled & 0 \\
GPT-5.6-sol & OpenAI-compatible & Disabled & 0 \\
Gemini-2.5-Flash & Gemini Native API & Disabled & 0 \\
Gemini-3.1-Pro-Preview & Gemini Native API & Disabled & 0 \\
Claude-Haiku-4.5 & Anthropic Native API & Disabled & 0 \\
Doubao-Seed-2.1-Pro & OpenAI-compatible & Disabled & 0 \\

Qwen3-4B & vLLM (OpenAI-compatible) & Disabled & 0 \\
Qwen3-8B & vLLM (OpenAI-compatible) & Disabled & 0 \\
Qwen3-30B-A3B & vLLM (OpenAI-compatible) & Disabled & 0 \\
Qwen3.5-4B & vLLM (OpenAI-compatible) & Disabled & 0 \\
Qwen3.5-9B & vLLM (OpenAI-compatible) & Disabled & 0 \\
Qwen3.5-27B & vLLM (OpenAI-compatible) & Disabled & 0 \\
Qwen3.5-35B-A3B & vLLM (OpenAI-compatible) & Disabled & 0 \\
Qwen3.5-397B-A17B & OpenAI-compatible & Disabled & 0 \\
Qwen3.8-27B & vLLM (OpenAI-compatible) & Disabled & 0 \\
GPT-oss-120B & vLLM (OpenAI-compatible) & Disabled & 0 \\
Llama-3.1-8B-Instruct & vLLM (OpenAI-compatible) & non-reasoning & 0 \\
Llama-3.1-70B-Instruct & vLLM (OpenAI-compatible) & non-reasoning & 0 \\

DeepSeek-V4-Pro & OpenAI-compatible & Disabled & 0 \\
GLM-5.2 & OpenAI-compatible & Disabled & 0 \\
Kimi K3 & OpenAI-compatible & Disabled & 0 \\
MiniMax-M2.5 & OpenAI-compatible & Forced enabled & 0 \\
\bottomrule
\end{tabular}}
\end{table}

%% file: Appendix/License_Information.tex
\begin{table}[htbp]
\centering
\caption{Summary of licenses of models evaluated in our benchmark.}
\label{tab:model_licenses}
\setlength{\tabcolsep}{6pt}
\begin{tabular}{@{}lll@{}}
\toprule
\textbf{Name} & \textbf{Model License} & \textbf{Code License} \\
\midrule
GPT-4o-mini & Proprietary & Proprietary \\
GPT-5-mini & Proprietary & Proprietary \\
GPT-5.6 Sol & Proprietary & Proprietary \\
Gemini-2.5-Flash & Proprietary & Proprietary \\
Gemini-3.1-Pro-Preview & Proprietary & Proprietary \\
Claude-Haiku-4.5 & Proprietary & Proprietary \\
Doubao-Seed-2.1-Pro & Proprietary & Proprietary \\
Qwen3-4B & Apache 2.0 & Apache 2.0 \\
Qwen3-8B & Apache 2.0 & Apache 2.0 \\
Qwen3-30B-A3B & Apache 2.0 & Apache 2.0 \\
Qwen3.5-4B & Apache 2.0 & Apache 2.0 \\
Qwen3.5-9B & Apache 2.0 & Apache 2.0 \\
Qwen3.5-27B & Apache 2.0 & Apache 2.0 \\
Qwen3.5-35B-A3B & Apache 2.0 & Apache 2.0 \\
Qwen3.5-397B-A17B & Apache 2.0 & Apache 2.0 \\
Qwen3.8-27B & Apache 2.0 & Apache 2.0 \\
GPT-oss-120B & Apache 2.0 & Apache 2.0 \\
Llama-3.1-8B-Instruct & Llama 3.1 Community License & Llama 3.1 Community License \\
Llama-3.1-70B-Instruct & Llama 3.1 Community License & Llama 3.1 Community License \\
DeepSeek-V4-Pro & MIT & MIT \\
GLM-5.2 & MIT & MIT \\
Kimi K3 & Kimi K3 License & Kimi K3 License \\
MiniMax-M2.5 & Modified-MIT & Modified-MIT \\
\bottomrule
\end{tabular}
\end{table}